\documentclass{article}

\PassOptionsToPackage{numbers, compress}{natbib}

\usepackage[preprint]{corl_2021}

\usepackage[utf8]{inputenc} %
\usepackage[T1]{fontenc}    %
\usepackage{hyperref}       %
\usepackage{url}            %
\usepackage{booktabs}       %
\usepackage{amsfonts}       %
\usepackage{nicefrac}       %
\usepackage{microtype}      %
\usepackage{colortbl}
\usepackage{placeins}

\usepackage{tikz}
\usetikzlibrary{positioning}
\usetikzlibrary{calc}
\usepackage{wrapfig}

\usepackage{times}
\usepackage{epsfig}
\usepackage{graphicx}
\usepackage{amsmath}
\usepackage{amssymb}
\usepackage{subcaption}
\usepackage{listings}
\usepackage{siunitx}
\usepackage{enumitem}
\usepackage{multirow}
\usepackage{xspace}
\usepackage{float}

\usepackage{geometry}
\usepackage{pdflscape}

\usepackage{amssymb}%
\usepackage{pifont}%
\usepackage{fontawesome}
\usepackage{adjustbox}
\usepackage{array}
\newcolumntype{R}[2]{%
    >{\adjustbox{angle=#1,lap=\width-(#2)}\bgroup}%
    l%
    <{\egroup}%
}
\newcommand*\rot{\multicolumn{1}{R{30}{1em}}}

\usepackage{rotating}
\definecolor{forestgreen}{rgb}{0.13, 0.55, 0.13}
\definecolor{indiagreen}{rgb}{0.07, 0.53, 0.03}

\newcommand{\model}{BEHAVIOR\xspace} 
\newcommand{\bddlfull}{BEHAVIOR Domain Definition Language\xspace} 
\newcommand{\bddl}{BDDL\xspace} 
\newcommand{\ig}{iGibson 2.0\xspace}
\newcommand{\myitem}[1]{\vspace{-3pt}\item{#1}}

\newcommand\numActivities{100\xspace}
\newcommand\numDefinitions{200\xspace}
\newcommand\numInstances{300\xspace}
\newcommand\numDemos{500\xspace}
\newcommand\totalTime{758.5\xspace} %
\newcommand\numObjects{1217\xspace}
\newcommand\numCategories{391\xspace}

\begin{document}

\title{{BEHAVIOR}: {\Large\underline{B}enchmark for \underline{E}veryday \underline{H}ousehold \underline{A}ctivities in \underline{V}irtual, \underline{I}nteractive, and Ec\underline{o}logical Envi\underline{r}onments}}

\author{%
Sanjana Srivastava\thanks{indicates equal contribution\newline correspondence to \href{mailto:sanjana2@stanford.edu,chengshu@stanford.edu}{\{sanjana2,chengshu\}@stanford.edu}} \And Chengshu Li$^*$ \And Michael Lingelbach$^*$ \And Roberto Mart\'in-Mart\'in$^*$ \And Fei Xia \And Kent Vainio \And Zheng Lian \And Cem Gokmen \And Shyamal Buch \And C. Karen Liu \And Silvio Savarese \And Hyowon Gweon \And Jiajun Wu \And Li Fei-Fei  \\\\ 
}

\maketitle
\vspace{-2em}
\begin{center} 
Stanford University
\end{center}
\vspace{1em}

\begin{abstract}
We introduce \model, a benchmark for embodied AI with \numActivities activities in simulation, spanning a range of everyday household chores such as cleaning, maintenance, and food preparation. These activities are designed to be realistic, diverse and complex, aiming to reproduce the challenges that agents must face in the real world. Building such a benchmark poses three fundamental difficulties for each activity: definition (it can differ by time, place, or person), instantiation in a simulator, and evaluation. \model addresses these with three innovations.  First, we propose an object-centric, predicate logic-based description language for expressing an activity's initial and goal conditions, enabling generation of diverse instances for any activity. Second, we identify the simulator-agnostic features required by an underlying environment to support \model, and demonstrate its realization in one such simulator. Third, we introduce a set of metrics to measure task progress and efficiency, absolute and relative to human demonstrators. We include \numDemos human demonstrations in virtual reality (VR) to serve as the human ground truth. Our experiments demonstrate that even state-of-the-art embodied AI solutions struggle with the level of realism, diversity, and complexity imposed by the activities in our benchmark. We  make \model publicly available at \href{behavior.stanford.edu}{behavior.stanford.edu} to facilitate and calibrate the development of new embodied AI solutions.

\end{abstract}


\section{Introduction}
\label{s:intro}

Embodied AI refers to the study and development of artificial agents that can perceive, reason, and interact with the environment with the capabilities and limitations of a physical body. Recently, significant progress has been made in developing solutions to embodied AI problems such as (visual) navigation~\cite{zhu2017target,hirose2019deep,wijmans2019dd,gupta2017cognitive,bansal2020combining}, interactive Q\&A~\cite{das2018embodied,yu2019multi,das2018neural,mousavian2019visual,marino2019ok}, instruction following~\cite{shridhar2020alfred,fu2018language,mattersim,wang2019reinforced,fried2018speaker}, and manipulation~\cite{haarnoja2018soft,schulman2017proximal,watters2019cobra,wu2019learning,billard2019trends,andrychowicz2020learning,xia2020relmogen}. 
To calibrate the progress, several lines of pioneering efforts have been made towards benchmarking embodied AI in simulated environments, including Rearrangement~\cite{batra2020rearrangement,weihs2021visual}, TDW Transport Challenge~\cite{gan2021threedworld}, VirtualHome~\cite{puig2018virtualhome}, ALFRED~\cite{shridhar2020alfred}, Interactive Gibson Benchmark~\cite{xia2020interactive}, MetaWorld~\cite{yu2020meta}, and RLBench~\cite{james2019rlbench}, among others~\cite{zhu2020robosuite,tassa2018deepmind,brockman2016openai}). 
These efforts are inspiring, but their activities represent only a fraction of challenges that humans face in their daily lives.
To develop artificial agents that can eventually perform and assist with everyday activities with human-level robustness and flexibility, we need a comprehensive benchmark with activities that are more \textbf{realistic}, \textbf{diverse}, and \textbf{complex}. 

But this is easier said than done. There are three major challenges that have prevented existing benchmarks to accommodate more realistic, diverse, and complex activities:
\begin{itemize}
    \myitem Definition: Identifying and defining meaningful activities for benchmarking;
    \myitem Realization: Developing simulated environments that realistically support such activities;
    \myitem Evaluation: Defining success and objective metrics for evaluating performance.
\end{itemize} 

\begin{figure}[t!]
    \centering
    \includegraphics[width=0.99\linewidth]{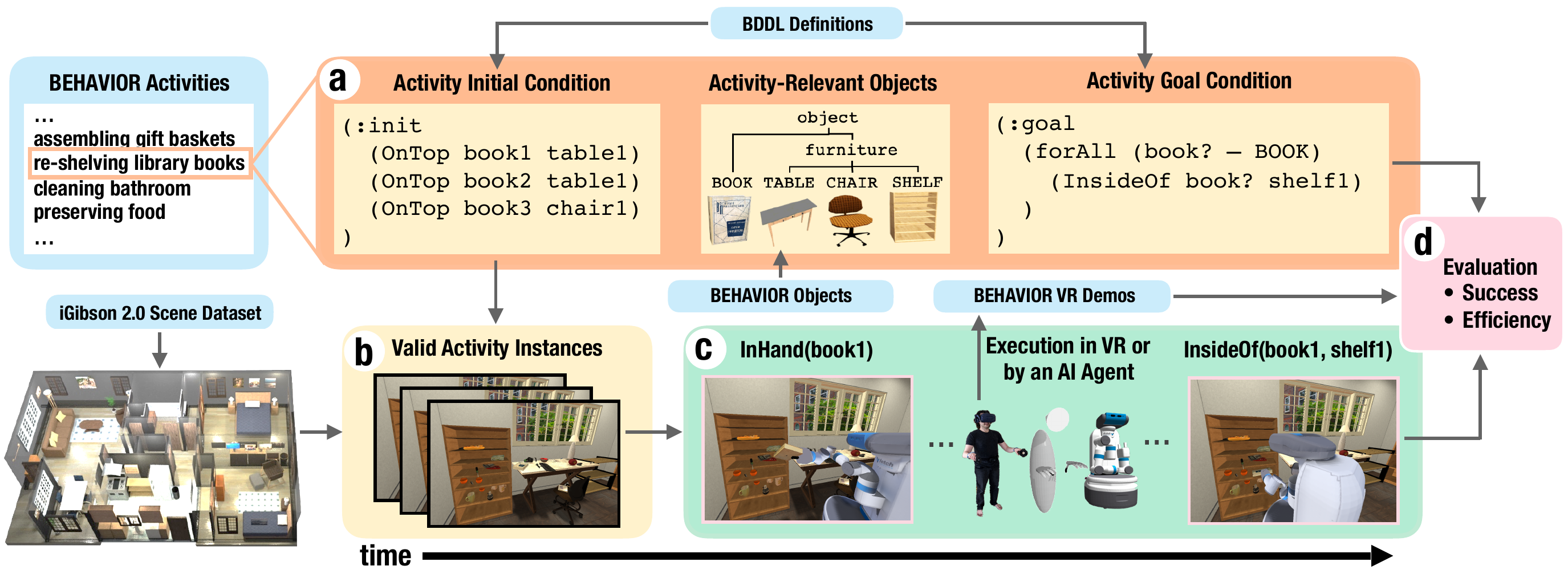}
    \vspace{-0.2cm}
    \caption{\textbf{Benchmarking Embodied AI with \model}: \raisebox{.5pt}{\textcircled{\raisebox{-.9pt} {a}}} We define \numActivities realistic household activities from the American Time Use Survey~\cite{atus} and define them with a set of relevant objects, organized with WordNet~\cite{miller1995wordnet}, and logic-symbolic initial and goal conditions in \bddl (Sec.~\ref{s:bddl}). \raisebox{.5pt}{\textcircled{\raisebox{-.9pt} {b}}} We provide an implementation of \model in \ig that generates potentially infinite diverse activity instances in realistic home scenes using the definition. \raisebox{.5pt}{\textcircled{\raisebox{-.9pt} {c}}} AI agents perform the activities in simulation through continuous physical interactions of an embodied avatar with the environment. Humans can perform the same activities in VR. \model includes a dataset of \numDemos successful VR demonstrations. \raisebox{.5pt}{\textcircled{\raisebox{-.9pt} {d}}} Changes in the scene are continuously mapped to their logic-symbolic equivalent representation in BDDL and checked against the goal condition; we provide intermediate success scores, metrics on agent's efficiency, and a human-centric metric relative to the demonstrations.}
    \label{fig:pullfig}
    \vspace{-0.5cm}
\end{figure}

We propose \textbf{BEHAVIOR} (Fig.~\ref{fig:pullfig})--\textbf{B}enchmark for \textbf{E}veryday \textbf{H}ousehold \textbf{A}ctivities in \textbf{V}irtual, \textbf{I}nteractive, and ec\textbf{O}logical envi\textbf{R}onments, addressing the three key challenges aforementioned with three technical innovations. First, we introduce \bddlfull (\bddl), a representation adapted from predicate logic that maps simulated states to semantic symbols. It allows us to define \numActivities activities as initial and goal conditions, and further enables generation of potentially infinite initial states and solutions for achieving the goal states.
Second, we facilitate its realization by listing environment-agnostic functional requirements for realistic simulation. With proper engineering, \model can be implemented in many existing environments; we provide a fully functional instantiation in \ig in this paper including the necessary object models (\numObjects models of \numCategories categories).
Third, we provide a comprehensive set of metrics to evaluate agent performance in terms of success and efficiency. To make evaluation comparable across diverse activities, scenes, and instances, we propose a set of metrics relative to demonstrated human performance on each activity, and provide a large-scale dataset of \numDemos human demonstrations (\totalTime min) in virtual reality, which serve as ground truth for evaluation and may also facilitate developing imitation learning solutions. 

\model activities are realistic, diverse, and complex. They comprise of \numActivities activities often performed by humans in their homes (e.g., cleaning, packing or preparing food) and require long-horizon solutions for changing not only the position of multiple objects but also their internal states or texture (e.g., temperature, wetness or cleanliness levels). As we demonstrate by experimentally evaluating the performance of two state-of-the-art reinforcement learning algorithms (Section 7), these properties make \model activities extremely challenging for existing solutions. By presenting well-defined challenges beyond the capabilities of current solutions, \model can serve as a unifying benchmark that guides the development of embodied AI.

\section{Related Work}
\label{sec:related-work}

Benchmarks and datasets have played a critical role in recent impressive advances in AI, particularly computer vision. 
Image~\cite{deng2009imagenet,lin2014microsoft,everingham2010pascal,krishna2017visual} and video datasets~\cite{geiger2012we,goyal2017something,sigurdsson2018actor,xiang2017posecnn,martin2021jrdb,caba2015activitynet} enable study and development of solutions for important research questions by providing both training data and fair comparison. These datasets, however, are passive observations, and therefore not well suited for development of embodied AI that must control and understand the consequences of their own actions. 

\paragraph{Benchmarks for Embodied AI:} Although real-world challenges~\cite{kitano1997robocup,wisspeintner2009robocup,iocchi2015robocup,buehler2009darpa,krotkov2017darpa,correll2016analysis,eppner2017lessons,roa2021mobile} provide the ultimate testbed for embodied AI agents, benchmarks in simulated environments serve as useful alternatives with several advantages; simulation enables faster, safer learning, and supports more reproducible, accessible, and fair evaluation.
However, in order to serve as a meaningful proxy for real-world performance, simulation benchmarks need to achieve high levels of 1) \textbf{realism} (in the activities, the models, the sensing and actuation of the agent), 2) \textbf{diversity} (of scenes, objects and activities benchmarked), and 3) \textbf{complexity} (length, number of objects, required skills and state changes). Below we review existing benchmarks based on these three criteria (see Table~\ref{tab:comparingbenchmarks} for a summary).

Benchmarks for \textit{visual navigation}~\cite{xia2018gibson,habitat19arxiv} provide high levels of visual realism and diversity of scenes, but they often lack interactivity or diversity of activities. 
The Interactive Gibson Benchmark~\cite{xia2020interactive} trades off some visual realism for physically realistic object manipulation in order to benchmark interactive visual navigation. %
While benchmarks for \textit{stationary manipulation}~\cite{lee2019ikea,james2019rlbench,yu2020meta,zhu2020robosuite,lin2020softgym,tassa2018deepmind,brockman2016openai} fare well on physical realism, they commonly fall short on diversity (of scenes, objects, tasks) and complexity (e.g., simple activities that take a few seconds). 
Benchmarks for \textit{instruction following}~\cite{shridhar2020alfred,puig2018virtualhome} provide diversity of scenes, objects and possible changes of the environment, but with low level of complexity; the horizon of the activities is shorter as the agents decide among a discrete set of predefined action primitives with full access to the state of the world.

Closer to \model, a recent group of benchmarks has focused on \textit{rearrangement tasks}~\cite{batra2020rearrangement,weihs2021visual,gan2021threedworld} in realistic simulation environments with diverse scenes.
The initial Rearrangement position paper~\cite{batra2020rearrangement} poses critical questions such as how to define embodied AI tasks and measure solution quality. Importantly, however, most household activities go far beyond the scope of rearrangement (see comparison in Fig.~\ref{fig:activitiesstatistics}). While such focus can inspire new solutions for solving rearrangement tasks, these solutions may not generalize to activities that require more than physical manipulation of object coordinates. Indeed, the majority of household activities involve other state changes (cooking, washing, etc. (Fig.~\ref{fig:activitiesstatistics},~\cite{atus}). \model therefore incorporates \numActivities activities that humans actually spend time on at home~\cite{atus} (Sec. \ref{s:pd}). To express such diverse activities in a common language, we present a novel logic-symbolic representation that defines activities in terms of initial and goal states, inspired by but distinct from the Planning Domain Definition Language~\cite{mcdermott1998pddl}. These yield in principle infinite instances per activity and accept any meaningful solution. We implement activity-independent metrics including a human-centric metric normalized to human performance; to facilitate comparison and development of new solutions, we also present a dataset of 500 successful VR demonstrations.

\section{\model: Benchmarking Realistic, Diverse, Complex Activities}
\label{s:pd}

\begin{table}[!t]
    \vspace{-1.6em}
    \definecolor{myGrayCell}{rgb}{.8,.8,.8}
    \centering
    \resizebox{\linewidth}{!}{
    \begin{tabular}{|c|c|c|ccccccccccccccccc|}
        \rot{} & 
        \rot{} & 
        \rot{\textbf{BEHAVIOR}} & 
        \rot{AI2THOR Vis. Room Rearr.} & 
        \rot{TDW Transport} & 
        \rot{Rearrangement T5 (Habitat)} &
        \rot{ManipulaTHOR ArmPointNav} & 
        \rot{Interactive Gibson Benchmark} & 
        \rot{VirtualHome} & 
        \rot{ALFRED} & 
        \rot{Rearrangement T2 (OCRTOC)} & 
        \rot{IKEA Furniture Assembly} & 
        \rot{RLBench} & 
        \rot{Metaworld} & 
        \rot{Robosuite} & 
        \rot{SoftGym} & 
        \rot{DeepMind Control Suite} & 
        \rot{OpenAIGym} & 
        \rot{Habitat 1.0} & 
        \rot{Gibson}
        \\\cline{3-20}
        \multicolumn{2}{c}{} & 
        \multicolumn{8}{|c}{Mobile manipulation} & 
        \multicolumn{8}{|c}{Static manipulation} & 
        \multicolumn{2}{|c|}{Navigation}
        \\\hline%
        & 
        \cellcolor[rgb]{.8,.8,.8} \begin{tabular}{@{}c@{}}Activity selections\\reflect human behavior\end{tabular} & 
        \cellcolor[rgb]{.8,.8,.8} \textcolor{indiagreen}\faCheck & 
        \cellcolor[rgb]{.8,.8,.8} \textcolor{red}\faTimes & 
        \cellcolor[rgb]{.8,.8,.8} \textcolor{red}\faTimes & 
        \cellcolor[rgb]{.8,.8,.8} \textcolor{red}\faTimes &
        \cellcolor[rgb]{.8,.8,.8} \textcolor{red}\faTimes &
        \cellcolor[rgb]{.8,.8,.8} \textcolor{red}\faTimes & 
        \cellcolor[rgb]{.8,.8,.8} \textcolor{red}\faTimes & 
        \cellcolor[rgb]{.8,.8,.8} \textcolor{red}\faTimes & 
        \cellcolor[rgb]{.8,.8,.8} \textcolor{red}\faTimes & 
        \cellcolor[rgb]{.8,.8,.8} \textcolor{red}\faTimes & 
        \cellcolor[rgb]{.8,.8,.8} \textcolor{red}\faTimes & 
        \cellcolor[rgb]{.8,.8,.8} \textcolor{red}\faTimes & 
        \cellcolor[rgb]{.8,.8,.8} \textcolor{red}\faTimes & 
        \cellcolor[rgb]{.8,.8,.8} \textcolor{red}\faTimes & 
        \cellcolor[rgb]{.8,.8,.8} \textcolor{red}\faTimes & 
        \cellcolor[rgb]{.8,.8,.8} \textcolor{red}\faTimes & 
        \cellcolor[rgb]{.8,.8,.8} \textcolor{red}\faTimes & 
        \cellcolor[rgb]{.8,.8,.8} \textcolor{red}\faTimes 
        \\
        & 
        \begin{tabular}{@{}c@{}}Kinematics, dynamics\end{tabular} & 
        \textcolor{indiagreen}\faCheck &
        \textcolor{indiagreen}\faCheck &
        \textcolor{indiagreen}\faCheck &
        \textcolor{indiagreen}\faCheck &
        \textcolor{indiagreen}\faCheck &
        \textcolor{indiagreen}\faCheck &
        \textcolor{red}\faTimes &
        \textcolor{indiagreen}\faCheck &
        \textcolor{indiagreen}\faCheck &
        \textcolor{indiagreen}\faCheck &
        \textcolor{indiagreen}\faCheck &
        \textcolor{indiagreen}\faCheck &
        \textcolor{indiagreen}\faCheck &
        \textcolor{indiagreen}\faCheck &
        \textcolor{indiagreen}\faCheck &
        \textcolor{indiagreen}\faCheck &
        \textcolor{indiagreen}\faCheck &
        \textcolor{indiagreen}\faCheck 
        \\ 
        \multirow{5}*{\rotatebox{90}{Realism}} & 
        \cellcolor[rgb]{.8,.8,.8} \begin{tabular}{@{}c@{}}Continuous extended\\states (e.g. temp., wetness)\end{tabular} & 
        \cellcolor[rgb]{.8,.8,.8} \textcolor{indiagreen}\faCheck & 
        \cellcolor[rgb]{.8,.8,.8} \textcolor{red}\faTimes & 
        \cellcolor[rgb]{.8,.8,.8} \textcolor{red}\faTimes & 
        \cellcolor[rgb]{.8,.8,.8} \textcolor{red}\faTimes & 
        \cellcolor[rgb]{.8,.8,.8} \textcolor{red}\faTimes & 
        \cellcolor[rgb]{.8,.8,.8} \textcolor{red}\faTimes & 
        \cellcolor[rgb]{.8,.8,.8} \textcolor{red}\faTimes & 
        \cellcolor[rgb]{.8,.8,.8} \textcolor{red}\faTimes & 
        \cellcolor[rgb]{.8,.8,.8} \textcolor{red}\faTimes & 
        \cellcolor[rgb]{.8,.8,.8} \textcolor{red}\faTimes & 
        \cellcolor[rgb]{.8,.8,.8} \textcolor{red}\faTimes & 
        \cellcolor[rgb]{.8,.8,.8} \textcolor{red}\faTimes & 
        \cellcolor[rgb]{.8,.8,.8} \textcolor{red}\faTimes & 
        \cellcolor[rgb]{.8,.8,.8} \textcolor{red}\faTimes & 
        \cellcolor[rgb]{.8,.8,.8} \textcolor{red}\faTimes & 
        \cellcolor[rgb]{.8,.8,.8} \textcolor{red}\faTimes & 
        \cellcolor[rgb]{.8,.8,.8} \textcolor{red}\faTimes & 
        \cellcolor[rgb]{.8,.8,.8} \textcolor{red}\faTimes 
        \\ 
        & 
        \begin{tabular}{@{}c@{}}Changing flexible materials\end{tabular} & 
        \textcolor{red}\faTimes & 
        \textcolor{red}\faTimes & 
        \textcolor{red}\faTimes & 
        \textcolor{red}\faTimes & 
        \textcolor{red}\faTimes & 
        \textcolor{red}\faTimes & 
        \textcolor{red}\faTimes & 
        \textcolor{red}\faTimes & 
        \textcolor{red}\faTimes & 
        \textcolor{red}\faTimes & 
        \textcolor{red}\faTimes & 
        \textcolor{red}\faTimes & 
        \textcolor{red}\faTimes & 
        \textcolor{indiagreen}\faCheck & 
        \textcolor{red}\faTimes & 
        \textcolor{red}\faTimes & 
        \textcolor{red}\faTimes & 
        \textcolor{red}\faTimes 
        \\ 
         &
        \cellcolor[rgb]{.8,.8,.8} \begin{tabular}{@{}c@{}}Realistic action execution\end{tabular} & 
        \cellcolor[rgb]{.8,.8,.8} \textcolor{indiagreen}\faCheck &
        \cellcolor[rgb]{.8,.8,.8} \textcolor{red}\faTimes &
        \cellcolor[rgb]{.8,.8,.8} \textcolor{indiagreen}\faCheck &
        \cellcolor[rgb]{.8,.8,.8} \textcolor{red}\faTimes &
        \cellcolor[rgb]{.8,.8,.8} \textcolor{indiagreen}\faCheck &
        \cellcolor[rgb]{.8,.8,.8} \textcolor{indiagreen}\faCheck &
        \cellcolor[rgb]{.8,.8,.8} \textcolor{red}\faTimes &
        \cellcolor[rgb]{.8,.8,.8} \textcolor{red}\faTimes &
        \cellcolor[rgb]{.8,.8,.8} \textcolor{indiagreen}\faCheck &
        \cellcolor[rgb]{.8,.8,.8} \textcolor{indiagreen}\faCheck &
        \cellcolor[rgb]{.8,.8,.8} \textcolor{indiagreen}\faCheck &
        \cellcolor[rgb]{.8,.8,.8} \textcolor{indiagreen}\faCheck &
        \cellcolor[rgb]{.8,.8,.8} \textcolor{indiagreen}\faCheck &
        \cellcolor[rgb]{.8,.8,.8} \textcolor{indiagreen}\faCheck &
        \cellcolor[rgb]{.8,.8,.8} \textcolor{indiagreen}\faCheck &
        \cellcolor[rgb]{.8,.8,.8} \textcolor{indiagreen}\faCheck &
        \cellcolor[rgb]{.8,.8,.8} \textcolor{indiagreen}\faCheck &
        \cellcolor[rgb]{.8,.8,.8} \textcolor{indiagreen}\faCheck 
        \\
        & 
        \begin{tabular}{@{}c@{}}Scenes reconstructed\\from real homes\end{tabular} & 
        \textcolor{indiagreen}\faCheck &
        \textcolor{red}\faTimes &
        \textcolor{red}\faTimes &
        \textcolor{indiagreen}\faCheck &
        \textcolor{red}\faTimes &
        \textcolor{indiagreen}\faCheck &
        \textcolor{red}\faTimes &
        \textcolor{red}\faTimes &
        \textcolor{red}\faTimes &
        \textcolor{red}\faTimes &
        \textcolor{red}\faTimes &
        \textcolor{red}\faTimes &
        \textcolor{red}\faTimes &
        \textcolor{red}\faTimes &
        \textcolor{red}\faTimes &
        \textcolor{red}\faTimes &
        \textcolor{indiagreen}\faCheck &
        \textcolor{indiagreen}\faCheck         
        \\\cline{1-2}
        & 
        \cellcolor[rgb]{.8,.8,.8} \begin{tabular}{@{}c@{}}\# Activities\end{tabular} & 
        \cellcolor[rgb]{.8,.8,.8} 100 & 
        \cellcolor[rgb]{.8,.8,.8} 1 & 
        \cellcolor[rgb]{.8,.8,.8} 1 & 
        \cellcolor[rgb]{.8,.8,.8} 1 &
        \cellcolor[rgb]{.8,.8,.8} 1 & 
        \cellcolor[rgb]{.8,.8,.8} 2 & 
        \cellcolor[rgb]{.8,.8,.8} \textbf{549} & 
        \cellcolor[rgb]{.8,.8,.8}  7 & 
        \cellcolor[rgb]{.8,.8,.8} 5 & 
        \cellcolor[rgb]{.8,.8,.8} 100 & 
        \cellcolor[rgb]{.8,.8,.8} 50 & 
        \cellcolor[rgb]{.8,.8,.8} 1 & 
        \cellcolor[rgb]{.8,.8,.8} 5 & 
        \cellcolor[rgb]{.8,.8,.8} 10 & 
        \cellcolor[rgb]{.8,.8,.8} 28 & 
        \cellcolor[rgb]{.8,.8,.8} 8 & 
        \cellcolor[rgb]{.8,.8,.8} 2 & 
        \cellcolor[rgb]{.8,.8,.8} 3
        \\
        \multirow{3}*{\rotatebox{90}{Diversity}} & 
        \begin{tabular}{@{}c@{}}Infinite scene-\\agnostic instantiation\end{tabular} & 
        \textcolor{indiagreen}\faCheck & 
        \textcolor{red}\faTimes & 
        \textcolor{red}\faTimes & 
        \textcolor{red}\faTimes & 
        \textcolor{red}\faTimes & 
        \textcolor{red}\faTimes & 
        \textcolor{red}\faTimes & 
        \textcolor{red}\faTimes & 
        \textcolor{red}\faTimes & 
        \textcolor{red}\faTimes & 
        \textcolor{red}\faTimes & 
        \textcolor{red}\faTimes & 
        \textcolor{red}\faTimes & 
        \textcolor{red}\faTimes & 
        \textcolor{red}\faTimes & 
        \textcolor{red}\faTimes & 
        \textcolor{red}\faTimes & 
        N/A
        \\ 
         & 
        \cellcolor[rgb]{.8,.8,.8} \begin{tabular}{@{}c@{}} Object categories\end{tabular} & 
        \cellcolor[rgb]{.8,.8,.8} 391 & 
        \cellcolor[rgb]{.8,.8,.8} 118 & 
        \cellcolor[rgb]{.8,.8,.8} ~50 & 
        \cellcolor[rgb]{.8,.8,.8} YCB & 
        \cellcolor[rgb]{.8,.8,.8} 150 & 
        \cellcolor[rgb]{.8,.8,.8} 5 & 
        \cellcolor[rgb]{.8,.8,.8} \textbf{509} & 
        \cellcolor[rgb]{.8,.8,.8} 84 & 
        \cellcolor[rgb]{.8,.8,.8} 12 & 
        \cellcolor[rgb]{.8,.8,.8} 73+ & 
        \cellcolor[rgb]{.8,.8,.8} ~28 & 
        \cellcolor[rgb]{.8,.8,.8} 7 & 
        \cellcolor[rgb]{.8,.8,.8} 10 & 
        \cellcolor[rgb]{.8,.8,.8} 4 & 
        \cellcolor[rgb]{.8,.8,.8} 4 & 
        \cellcolor[rgb]{.8,.8,.8} 4 & 
        \cellcolor[rgb]{.8,.8,.8} Matterport & 
        \cellcolor[rgb]{.8,.8,.8} N/A
        \\
        & 
        \begin{tabular}{@{}c@{}} Object models\end{tabular} & 
        \textbf{1217} & 
        118 &
        112 & 
        YCB &
        150 & 
        152 & 
        &
        84 &
        101 + YCB & 
        73+ & 
        28 & 
        80 & 
        10 & 
        4 & 
        4 & 
        4 & 
        N/A & 
        N/A
        \\
        & 
        \cellcolor[rgb]{.8,.8,.8}Scenes / Rooms & 
        \cellcolor[rgb]{.8,.8,.8}\begin{tabular}{@{}c@{}}\textbf{15 / }\\\textbf{100}\end{tabular} & 
        \cellcolor[rgb]{.8,.8,.8}\begin{tabular}{@{}c@{}}\textbf{- /} \\\textbf{120}\end{tabular} & 
        \cellcolor[rgb]{.8,.8,.8}\begin{tabular}{@{}c@{}}\textbf{15 /} \\\textbf{90-120}\end{tabular} & 
        \cellcolor[rgb]{.8,.8,.8}\begin{tabular}{@{}c@{}}55 static / \\-\end{tabular} & 
        \cellcolor[rgb]{.8,.8,.8}\begin{tabular}{@{}c@{}}- /\\30\end{tabular} & 
        \cellcolor[rgb]{.8,.8,.8}\begin{tabular}{@{}c@{}} 10 /\\-\end{tabular} & 
        \cellcolor[rgb]{.8,.8,.8}\begin{tabular}{@{}c@{}}7 / \\-\end{tabular} &
        \cellcolor[rgb]{.8,.8,.8}\begin{tabular}{@{}c@{}}- / \\120\end{tabular} & 
        \cellcolor[rgb]{.8,.8,.8}\begin{tabular}{@{}c@{}}1 / \\-\end{tabular} & 
        \cellcolor[rgb]{.8,.8,.8}\begin{tabular}{@{}c@{}}1 / \\-\end{tabular} & 
        \cellcolor[rgb]{.8,.8,.8}\begin{tabular}{@{}c@{}}1 / \\-\end{tabular} & 
        \cellcolor[rgb]{.8,.8,.8}\begin{tabular}{@{}c@{}}1 / \\-\end{tabular} & 
        \cellcolor[rgb]{.8,.8,.8}\begin{tabular}{@{}c@{}}1 / \\-\end{tabular} & 
        \cellcolor[rgb]{.8,.8,.8}\begin{tabular}{@{}c@{}}1 / \\-\end{tabular} & 
        \cellcolor[rgb]{.8,.8,.8}\begin{tabular}{@{}c@{}}1 / \\-\end{tabular} & 
        \cellcolor[rgb]{.8,.8,.8}\begin{tabular}{@{}c@{}}1 / \\-\end{tabular} & 
        \cellcolor[rgb]{.8,.8,.8}\begin{tabular}{@{}c@{}}Matterport\\+ Gibson\end{tabular} &
        \cellcolor[rgb]{.8,.8,.8} \begin{tabular}{@{}c@{}}572\\static\end{tabular} 

        \\ \cline{1-2}
        \multirow{6}*{\rotatebox{90}{Complexity}}  & 
         \begin{tabular}{@{}c@{}}Activity length\footnotemark (steps)\end{tabular} &
         \begin{tabular}{@{}c@{}}\textbf{300-}\\\textbf{20000}\end{tabular} & 
         <100 & 
         100-1000 &
         100-1000 &
         <100 & 
         100-1000 & 
         <100 &
         <100 & 
         100-1000 & 
         <1000 &
         <100 &
         <100 &
         <100 &
         <100 &
         <100 &          
         <100 &
         <100 & 
         100-1000
        \\
        &
        \cellcolor[rgb]{.8,.8,.8}\begin{tabular}{@{}c@{}} Objs. per activity\end{tabular} & 
        \cellcolor[rgb]{.8,.8,.8}\textbf{3-34} & 
        \cellcolor[rgb]{.8,.8,.8}5 &
        \cellcolor[rgb]{.8,.8,.8}7-9 &
        \cellcolor[rgb]{.8,.8,.8}2-5 &
        \cellcolor[rgb]{.8,.8,.8}2-3 & 
        \cellcolor[rgb]{.8,.8,.8}~10 & 
        \cellcolor[rgb]{.8,.8,.8}1-24 &
        \cellcolor[rgb]{.8,.8,.8}2 &
        \cellcolor[rgb]{.8,.8,.8}5-10 &
        \cellcolor[rgb]{.8,.8,.8}1-2 &
        \cellcolor[rgb]{.8,.8,.8}1-2 &
        \cellcolor[rgb]{.8,.8,.8}1 &
        \cellcolor[rgb]{.8,.8,.8}1-3 &
        \cellcolor[rgb]{.8,.8,.8}1-3 &
        \cellcolor[rgb]{.8,.8,.8}1-3 &
        \cellcolor[rgb]{.8,.8,.8}1 &
        \cellcolor[rgb]{.8,.8,.8}0-1 &
        \cellcolor[rgb]{.8,.8,.8}N/A
        \\
        & 
        \cellcolor[rgb]{1,1,1}\begin{tabular}{@{}c@{}}Benchmark focus: Task-\\Planning and/or Control\end{tabular} &
        \cellcolor[rgb]{1,1,1} TP+C & 
        \cellcolor[rgb]{1,1,1} TP & 
        \cellcolor[rgb]{1,1,1} TP+C & 
        \cellcolor[rgb]{1,1,1} TP+C & 
        \cellcolor[rgb]{1,1,1} TP+C & 
        \cellcolor[rgb]{1,1,1} C & 
        \cellcolor[rgb]{1,1,1} TP &         
        \cellcolor[rgb]{1,1,1} TP & 
        \cellcolor[rgb]{1,1,1} TP+C & 
        \cellcolor[rgb]{1,1,1} C & 
        \cellcolor[rgb]{1,1,1} TP+C & 
        \cellcolor[rgb]{1,1,1} C & 
        \cellcolor[rgb]{1,1,1} C & 
        \cellcolor[rgb]{1,1,1} C & 
        \cellcolor[rgb]{1,1,1} C & 
        \cellcolor[rgb]{1,1,1} C & 
        \cellcolor[rgb]{1,1,1} C & 
        \cellcolor[rgb]{1,1,1} C 
        \\
        & 
        \cellcolor[rgb]{.8,.8,.8}\begin{tabular}{@{}c@{}}Diff. state changes required\\per activity (see \ref{fig:activitiesstatistics})\end{tabular} & 
        \cellcolor[rgb]{.8,.8,.8}\textbf{2-8} & 
        \cellcolor[rgb]{.8,.8,.8}4 & 
        \cellcolor[rgb]{.8,.8,.8}4 &
        \cellcolor[rgb]{.8,.8,.8}4 &
        \cellcolor[rgb]{.8,.8,.8}2 &
        \cellcolor[rgb]{.8,.8,.8}1-3 & 
        \cellcolor[rgb]{.8,.8,.8}1-7 & 
        \cellcolor[rgb]{.8,.8,.8}2-3 & 
        \cellcolor[rgb]{.8,.8,.8}1 & 
        \cellcolor[rgb]{.8,.8,.8}1-3 & 
        \cellcolor[rgb]{.8,.8,.8}1-4 & 
        \cellcolor[rgb]{.8,.8,.8}4 & 
        \cellcolor[rgb]{.8,.8,.8}1 & 
        \cellcolor[rgb]{.8,.8,.8}1-3 & 
        \cellcolor[rgb]{.8,.8,.8}1-2 & 
        \cellcolor[rgb]{.8,.8,.8}1-2 & 
        \cellcolor[rgb]{.8,.8,.8}1 & 
        \cellcolor[rgb]{.8,.8,.8}1
        \\\cline{1-2}
        \rot{} & 
        \multicolumn{1}{|c|}{\begin{tabular}{@{}c@{}}\# Human VR demos\end{tabular}} & 
        \cellcolor[rgb]{1,1,1} \textbf{\numDemos} & 
        \cellcolor[rgb]{1,1,1} 0 & 
        \cellcolor[rgb]{1,1,1} 0 & 
        \cellcolor[rgb]{1,1,1} 0 & 
        \cellcolor[rgb]{1,1,1} 0 & 
        \cellcolor[rgb]{1,1,1} 0 & 
        \cellcolor[rgb]{1,1,1} 0 & 
        \cellcolor[rgb]{1,1,1} 0 & 
        \cellcolor[rgb]{1,1,1} 0 & 
        \cellcolor[rgb]{1,1,1} 0 & 
        \cellcolor[rgb]{1,1,1} 0 & 
        \cellcolor[rgb]{1,1,1} 0 & 
        \cellcolor[rgb]{1,1,1} 0 & 
        \cellcolor[rgb]{1,1,1} 0 & 
        \cellcolor[rgb]{1,1,1} 0 & 
        \cellcolor[rgb]{1,1,1} 0 & 
        \cellcolor[rgb]{1,1,1} 0 & 
        \cellcolor[rgb]{1,1,1} 0 
        \\\cline{2-20}
    \end{tabular}
  }
\begin{flushleft}\tiny\textsuperscript{1}Estimate of a near-optimal, e.g. human, execution of the activity given the platform's action space \end{flushleft} 
  \vspace{0.2em}
  \caption{\textbf{Comparison of Embodied AI Benchmarks:} 
  \model activities are exceptionally realistic due to their grounding in human population time use~\cite{atus} and realistic simulation (sensing, actuation, changes in environment) in \ig. The activity set is diverse in topic, objects used, scenes done in, and state changes required. The diversity is reinforced by the ability to generate infinite new instances scene-agnostically. \model activities are complex enough to reflect real-world housework: many decision steps and objects in each activity. This makes \model uniquely well-suited to benchmark task-planning and control, and it is the only one to include human VR demonstrations (see Table~\ref{tab:comparingbenchmarksfull} for more detail).
  }  
  \label{tab:comparingbenchmarks}
  \vspace{-1.5em}
\end{table}

Building on the advances led by existing benchmarks, \model aims to reach new levels of realism, diversity, and complexity by using household activities as a domain for benchmarking AI. See Table~\ref{tab:comparingbenchmarks} for comparisons between \model and existing benchmarks.

\paragraph{Realism in \model Activities:} To effectively benchmark embodied AI agents in simulation, we need realistic activities that pose similar challenges to those in the real world. \model achieves this by using a data-driven approach to identify activities that approximate the true distribution of real household activities. To this end, we use the American Time Use Survey (ATUS, ~\cite{atus}): A survey from the U.S. Bureau of Labor Statistics on how Americans spend their time. \model activities come from, and are distributed similarly to, the full space of simulatable activities in ATUS (see Fig.~\ref{fig:activitiesstatistics}). The use of an independently curated source of real-world activities is a unique strength of \model as a benchmark that reflects natural behaviors of a large population. %

\model also achieves realism by simulating these activities in reconstructions of real-world homes. We use \ig, a simulation environment with realistic physics simulation from the Bullet~\cite{coumans2016pybullet} physics engine and high-quality virtual sensor signals (see Fig.~\ref{fig:igvisuals}), which includes 15 ecological, fully interactive 3D models of real-world homes with furniture layouts that approximate their real counterparts. These scenes are further populated with object models created by professional artists from the new \model Object dataset, which includes 1217 models of 391 categories grounded in the WordNet~\cite{miller1995wordnet} taxonomy. The dataset covers a data-driven selection of activity-related objects (see Fig.~\ref{fig:objects_from_wiki}). Figs.~\ref{fig:supp_object_collage} and~\ref{fig:object_taxonomy} illustrate examples of objects and taxonomic arrangement. The 100 BEHAVIOR activities, visualized in Fig.~\ref{fig:behavior_acts}, go beyond comparable benchmarks that evaluate a few hand-picked activities in less realistic setups (see Table~\ref{tab:comparingbenchmarks} Realism). 

\paragraph{Diversity in \model Activities:} %
Benchmarks with diverse activities demand generalizable solutions. In real-world homes, agents encounter a range of activities that differ in 1) the capabilities required for achieving them, 2) the environments in which they occur (e.g., scenes, objects), and 3) the initial states of a particular scene.
\model presents extensive diversity in all these dimensions. 
We include \numActivities activities that require a wide variety of state changes (e.g., moving objects, soaking materials, cleaning surfaces, heating/freezing food) demanding a broad set of agent capabilities (see Fig~\ref{fig:activitiesstatistics}). To reflect the diversity in the ways humans encounter, understand, and accomplish these activities, we provide two example definitions per activity.
\bddl, our novel representation for activity definition, allows new valid instances to be sampled from each definition, providing potentially infinite number of instances per activity. The resulting instances vary over scene, object models, and configuration, supported by implementation in \ig and \model Object dataset. Related benchmarks focus on fewer tasks, mostly limited to kinematic state changes and with scene- or position-constant instantiation (see Table~\ref{tab:comparingbenchmarks} Diversity).

\paragraph{Complexity in \model Activities:} Beyond diversity across activities, \model also raises the complexity of the activities themselves by benchmarking full household activities that parallel the length (number of steps an agent needs), the number of objects involved, and the number of required capabilities of real-world chores (see Fig.~\ref{fig:act_volumes}, comparison in Table~\ref{tab:comparingbenchmarks} Complexity). 
Compared to activities in existing benchmarks, these activities are very long-horizon with some requiring several thousand steps (even for humans in VR; see Fig.~\ref{fig:vrstatistics}), involve more objects (avg. 10.5), and require a heterogeneous set of capabilities (range: 2 - 8) to change various environment states.

\section{Defining Realistic, Diverse, and Complex Household Activities with BDDL}
\label{s:bddl}

\model challenges embodied AI agents to achieve a diverse set of complex long-horizon household activities through physical interactions in a realistically simulated home environment.
Adopting the common formalism of partially-observable Markov decision processes (POMDP), each activity is represented by the tuple $\mathcal{M} = (\mathcal{S},\mathcal{A},\mathcal{O}, \mathcal{T},\mathcal{R},\gamma)$. Here, $\mathcal{S}$ is the state space; $\mathcal{A}$ is the action space; $\mathcal{O}$ is the observation space; $\mathcal{T}(s'|s,a), s\in\mathcal{S}, a\in\mathcal{A}$, is the state transition model; $\mathcal{R}(s, a) \in \mathbb{R}$ is the reward function; $\gamma$ is the discount factor.
Based on a full representation of the physical state, $\mathcal{S}$, the simulation environment generates realistic transitions to embodied AI agents' actions, $a\in\mathcal{A}$, i.e., physical interactions, and close-to-real observations, $o\in\mathcal{O}$, e.g., virtual images. 

We define an \textit{activity} $\tau$ as two sets of states, $\tau = \{S_{\tau,0}, S_{\tau,g}\}$, where $S_{\tau,0}$ is a set of possible initial states and $S_{\tau,g}$ is a set of acceptable goal states. In an \textit{activity instance}, the agent must change the world state from some concrete $s_0 \in S_{\tau,0}$ to any $s_g \in S_{\tau,g}$. However, describing activities in the physical state space generates scene- or pose-specific definitions (e.g., ~\cite{batra2020rearrangement, zhu2020robosuite, james2019rlbench}) that are far more specific than how humans represent these activities, limiting the diversity and complexity of existing embodied AI benchmarks.
To overcome this, we introduce \textit{\model Domain Definition Language} (\bddl), a predicate logic-based language that establishes a symbolic state representation built on predefined, meaningful predicates grounded in simulated physical states; its variables and constants represent object categories from the \model object dataset.
Each activity is defined in \bddl as an initial and goal condition parametrizing sets of possible initial states and satisfactory goal states $\bar{{S}}_{\tau,0}$ and $\bar{{S}}_{\tau,g}$. \bddl predicates create symbolic counterparts of the physical state, $\bar{{S}}$ (see Fig.~\ref{fig:tasknetpreds}).

\newcommand\hspacehere{2}
\newcommand\widthhere{0.190}

\newcommand{\tikzpicpred}[3]{
\begin{scope}[xshift=#1]
\node [below right, outer sep=0pt, inner sep=0] (image) at (0,0){\includegraphics[width=0.19\textwidth]{#2}};%
\node [font=\fontsize{5.5pt}{5}\selectfont, below right, fill=white,rounded corners=0.5pt,inner sep=1pt] at (0.3pt,-0.3pt) {#3};%
\end{scope}
}%
\newcommand{\myheightbtwpredtikz}{2pt}
\begin{figure}%{r}{0.52\linewidth}
% \vspace{-1em}
\begin{subfigure}[b]{0.99\linewidth}
\centering
\begin{tikzpicture}
\tikzpicpred{0}{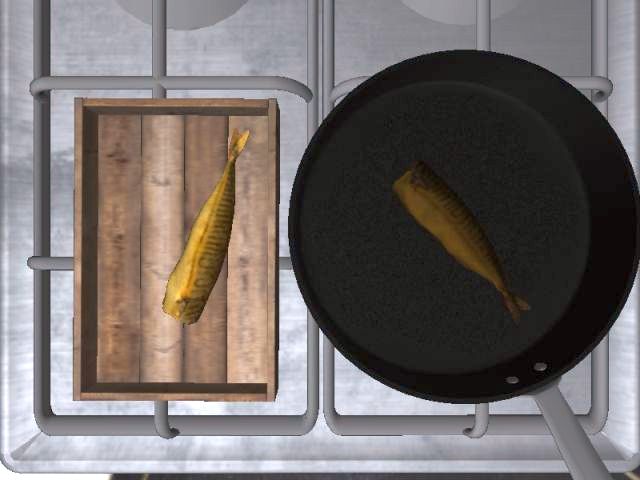}{Burnt(fish)}%
\tikzpicpred{0.2\textwidth}{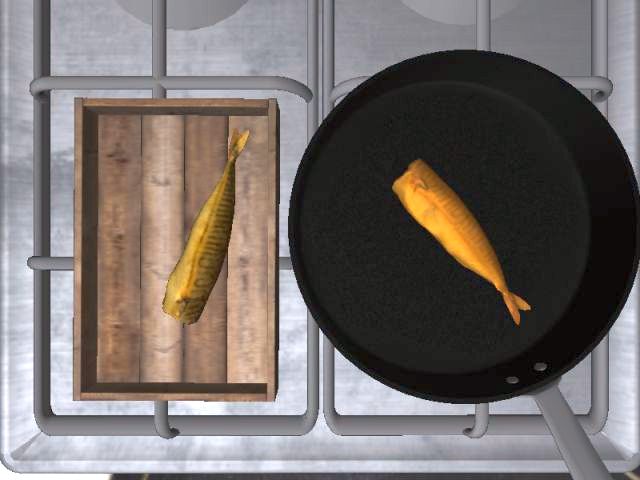}{Cooked(fish)}%
\tikzpicpred{0.4\textwidth}{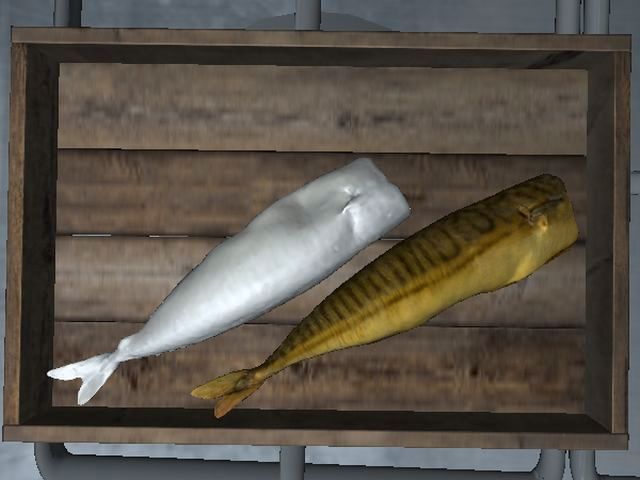}{Frozen(fish)}%
\tikzpicpred{0.6\textwidth}{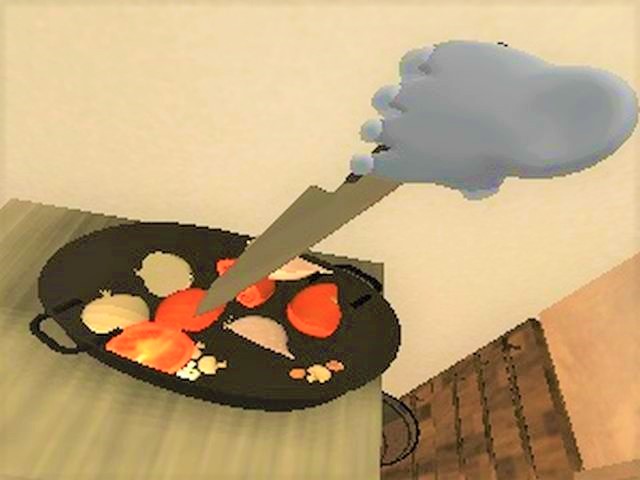}{Sliced(tomato)}%
\tikzpicpred{0.8\textwidth}{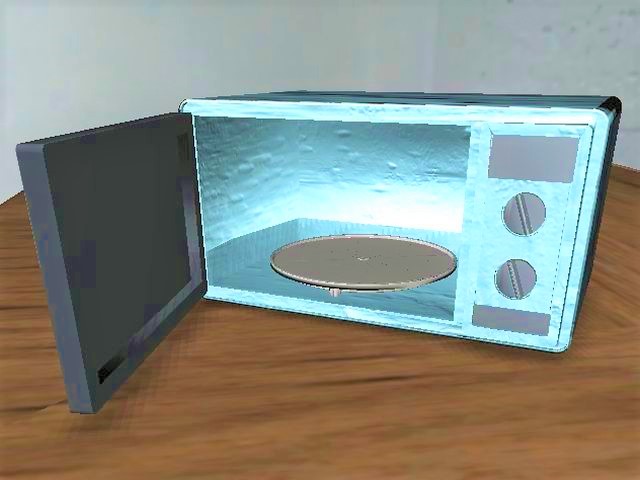}{Open(microwave)}%
\end{tikzpicture}%
\vspace{\myheightbtwpredtikz}
\begin{tikzpicture}
\tikzpicpred{0}{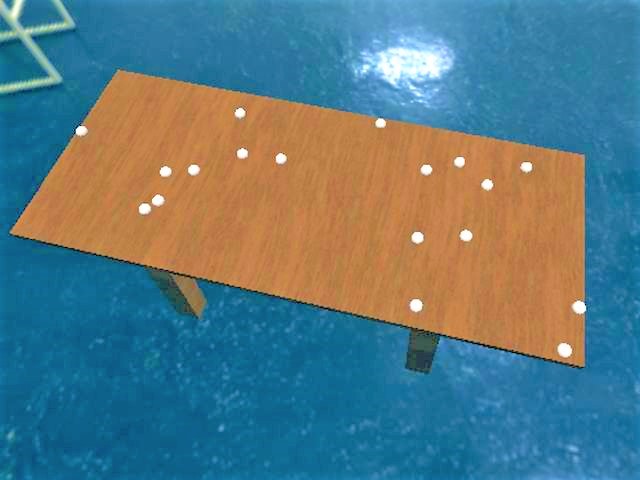}{Dusty(table)}%
\tikzpicpred{0.2\textwidth}{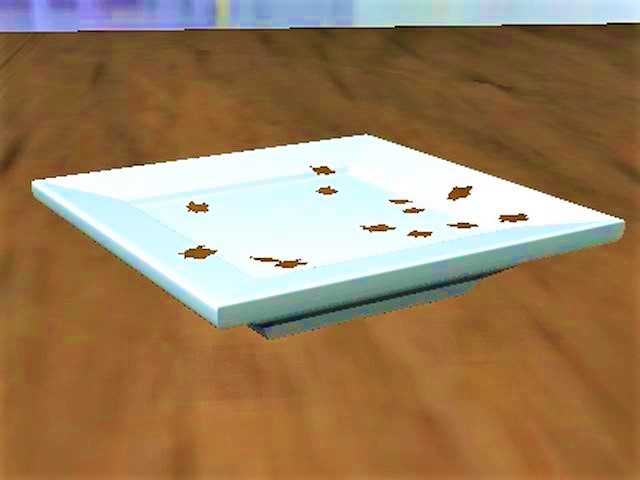}{Stained(plate)}%
\tikzpicpred{0.4\textwidth}{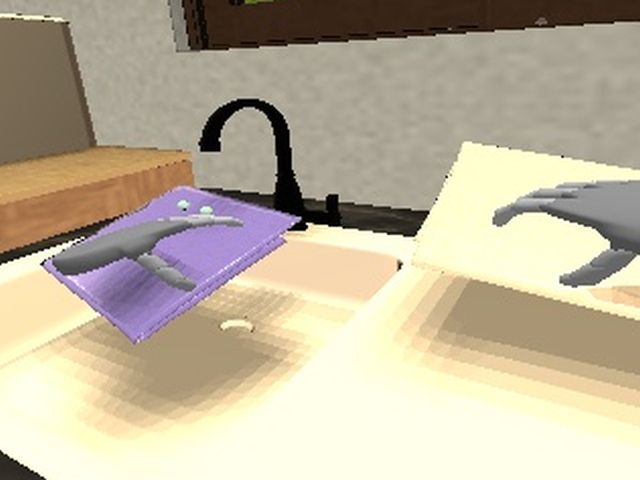}{Soaked(rag)}%
\tikzpicpred{0.6\textwidth}{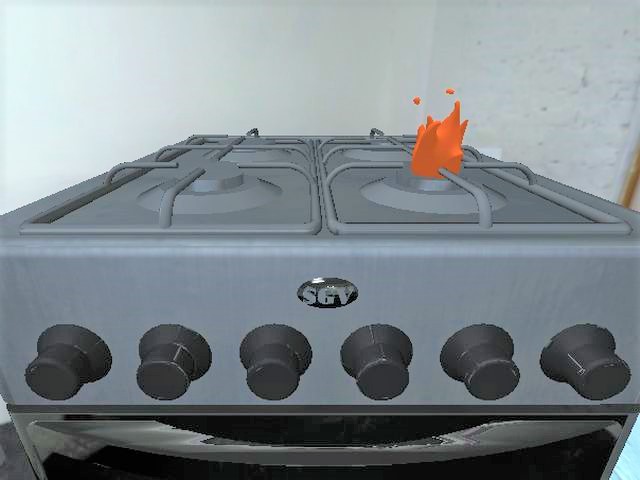}{ToggledOn(stove)}%
\tikzpicpred{0.8\textwidth}{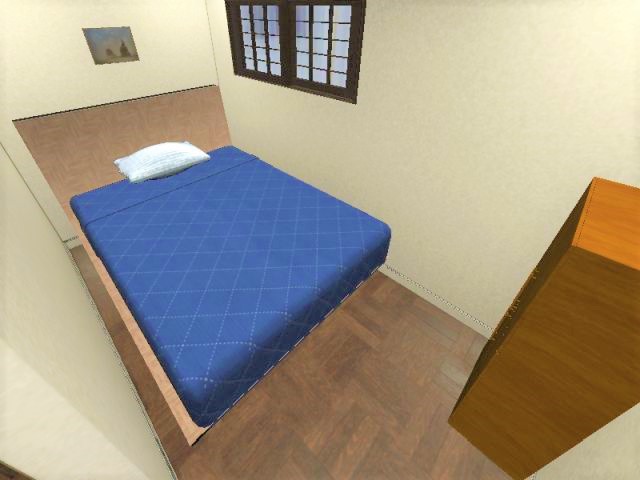}{InRoom(bed, bedroom)}%
\end{tikzpicture}%
\vspace{\myheightbtwpredtikz}
\begin{tikzpicture}
\tikzpicpred{0}{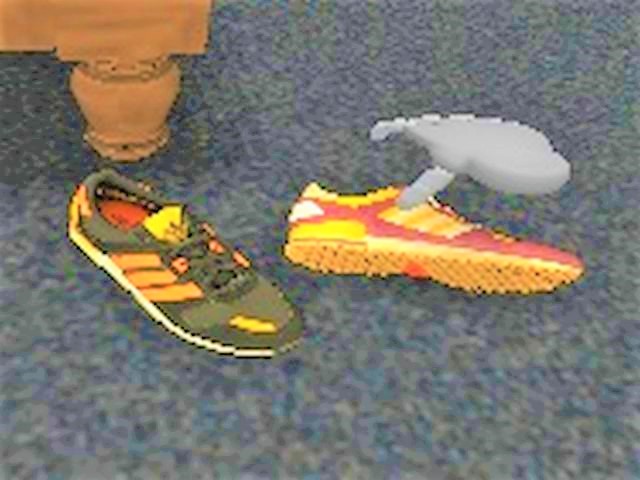}{OnFloor(shoe)}%
\tikzpicpred{0.2\textwidth}{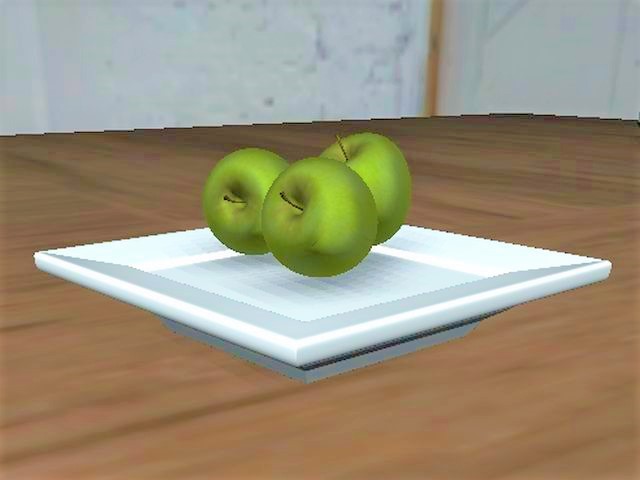}{OnTopOf(apple, plate)}%
\tikzpicpred{0.4\textwidth}{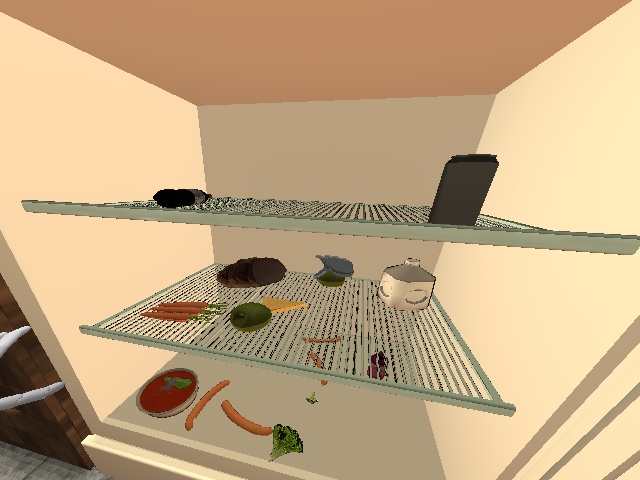}{InsideOf(food, fridge)}%
\tikzpicpred{0.6\textwidth}{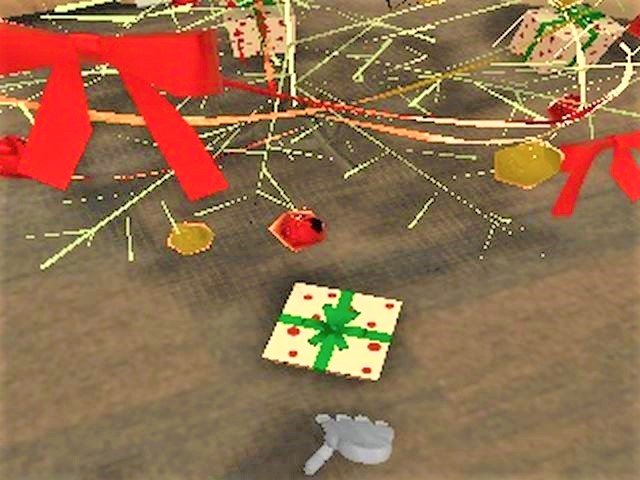}{Under(present, tree)}%
\tikzpicpred{0.8\textwidth}{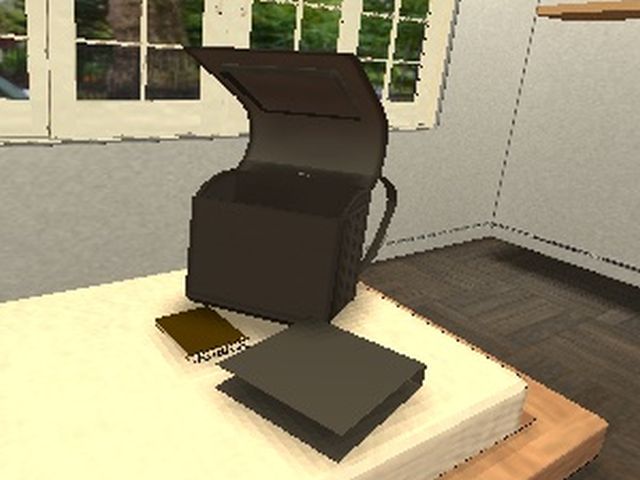}{NextTo(book, bag)}%
\end{tikzpicture}

    \end{subfigure}
        \caption{\textbf{Unary and Binary Predicates in BDDL:} We represent object states and relationships to other objects based on their kinematics, temperature, wetness level and other physical and functional properties, enabling a diverse and complex set of realistic activities}
    \label{fig:tasknetpreds}
    \vspace{-1em}
\end{figure}

BDDL overcomes limitations that hinder diversity through two mechanisms: first, an initial condition maps to infinite physical states in diverse scenes. Second, a goal condition detects all semantically satisfactory solutions, rather than limiting to a few or only those that obey semantically uninteresting geometric constraints (see Fig.~\ref{fig:multiple_init_goal} for examples). This state-based definition is also entirely declarative, providing a true benchmark of planning ability. By comparison, other benchmarks are limited to scene- or pose-specific instantiation and solution acceptance, and/or have imperative plans. \model includes a systematic generation pipeline (see~\ref{ss:crowd}) allowing unlimited definitions per activity and formalizing the inherent subjectivity and situationality of household activities. We include \numDefinitions definitions and \numInstances activity instances in simulation (see Sec.~\ref{s:instan}). \model is thus the only benchmark equipped to formalize unlimited human-defined versions of an activity and create practically infinite unique instantiations in any scene.

\section{Instantiating \model in a Realistic Physics Simulator}
\label{s:instan}

While \model is not bounded to any specific simulation environment, there are a set of functional requirements that are necessary to simulate \model activities: 1) maintain an object-centric representation (object identities enriched with properties and states), 2) simulate physical forces and motion, and generate virtual sensor signals (images), 3) simulate additional, non-kinematic properties per object (e.g. temperature, wetness level, cleanliness level), 4) implement functionality to \textbf{generate} valid instances based on the literals defining an activity's initial condition, e.g., instantiating an object \texttt{insideOf} another, and 5) implement functionality to \textbf{evaluate} the atomic formulae relevant to the goal condition, e.g. checking whether an object is \texttt{cooked} or \texttt{onTopOf} another. 

Additionally, the simulator must provide an interface of the action space $\mathcal{A}$ and the observation space $\mathcal{O}$ of the underlying POMDP to embodied AI agents (Sec.~\ref{s:bddl}). 
While \model activities are not tailored to a specific embodiment, we propose two concrete bodies to fulfill the activities (see Fig.~\ref{fig:pullfig}): a \textit{bimanual humanoid} avatar (24 degrees of freedom, DoF), and a \textit{Fetch robot} (12/13 DoF), both capable of navigating, grasping and interacting with the hand(s).
Humans in VR embody the bimanual humanoid.
Agents trained with the Fetch embodiment could be directly tested with a real-world version of the hardware (see discussion on sim2real in Sec.~\ref{ss:sim2real}). Both embodiments receive sensor signals from the on-board virtual sensors, and perform actions at \SI{30}{\hertz}. 

We provide a fully functional implementation of \model using \ig, a new version of the open-source simulation environment iGibson that fulfills the requirements above. \ig provides an object-centric representation with additional properties, support for sources of heat and water, dust and stain particles, and changes in object appearance based on extended states. We implement the two embodiments in \ig: the agent receives proprioceptive information and has access to \ig's generated realistic signals: RGB, depth images, LiDAR, normals, flow (optical, spatial), and semantic and instance segmentation. 
While this control and sensing setup is standard in \model, we additionally implement a set of action primitives inspired by~\cite{gan2021threedworld, habitat19arxiv, kolve2017ai2, weihs2021visual} to facilitate solution prototyping and task-planning research. 
The primitives execute sequences of low-level actions resulting from a motion planning process (bilateral RRT$^*$~\cite{Jordan.Perez.ea:CSAIL13}) to {\small\texttt{navigateTo}}, {\small\texttt{grasp}}, {\small\texttt{placeOnTop}}, {\small\texttt{placeInside}}, {\small\texttt{open}}, and {\small\texttt{close}} the target object provided as arguments. Even though the agent only relies on sensory observations to decide on action primitive, the primitives themselves internally assume access to privileged information (e.g. object identities, poses, and geometric shapes for planning). 
Further details can be found in Sec.~\ref{ss:igv2} and in the cross-submission included in appendix. 
Our implementation of \model in \ig goes beyond the capabilities of existing benchmarks and amplifies realism, diversity, and complexity.

\section{Evaluation Metrics: Success, Efficiency and Human-Centric Metric}
\label{s:metrics}

\model provides evaluation metrics to quantify the performance of an embodied AI solution.
Extending prior metrics suggested for Rearrangement~\cite{batra2020rearrangement}, we propose a primary metric based on success and several secondary metrics for characterizing efficiency.

\paragraph{Primary Metric -- Success Score Q:}
The main goal of an embodied AI agent in \model is to perform an activity successfully (i.e., all logical expressions in the goal condition are met). A binary definition of success, however, only signals the end of a successful execution and cannot assess interim progress. To provide more guidance to agents and enable comparisons of partial solutions, we propose \textbf{success score} as the primary metric, defined as the \textbf{maximum fraction of satisfied goal literals in a ground solution to the goal condition} at each step. More formally: %

Given an activity $\tau$ with goal state set $\bar{S}_{\tau,g}$, its goal condition can be flattened to a set $C$ of conjunctions $C_i$ of ground literals $l_{j_i}$. 
For any $C_i \in C$, if all $l_{j_i} \in C_i$ are true then the goal condition is satisfied (see~\ref{sss:bddl} for definitions and technical details on flattening), i.e. for some current environment state $s$, we have {\small $\bigvee\limits_{C_i}\bigwedge\limits_{l_{j_i}}l_{j_i} = \texttt{True}\implies s \in \bar{S}_{\tau,g}$} .
We compute the fraction of literals $l_{j_i}$ that are \texttt{True} for each $C_i$, and 
define the overall success score by taking the maximum: {\small$Q = \max\limits_{C}\frac{|\{l_{j_i}|l_{j_i} = \texttt{True}\}|}{|C_i|}$}, where $|\cdot|$ is set cardinality.

\begin{figure}
        \centering
        \includegraphics[width=0.49\linewidth]{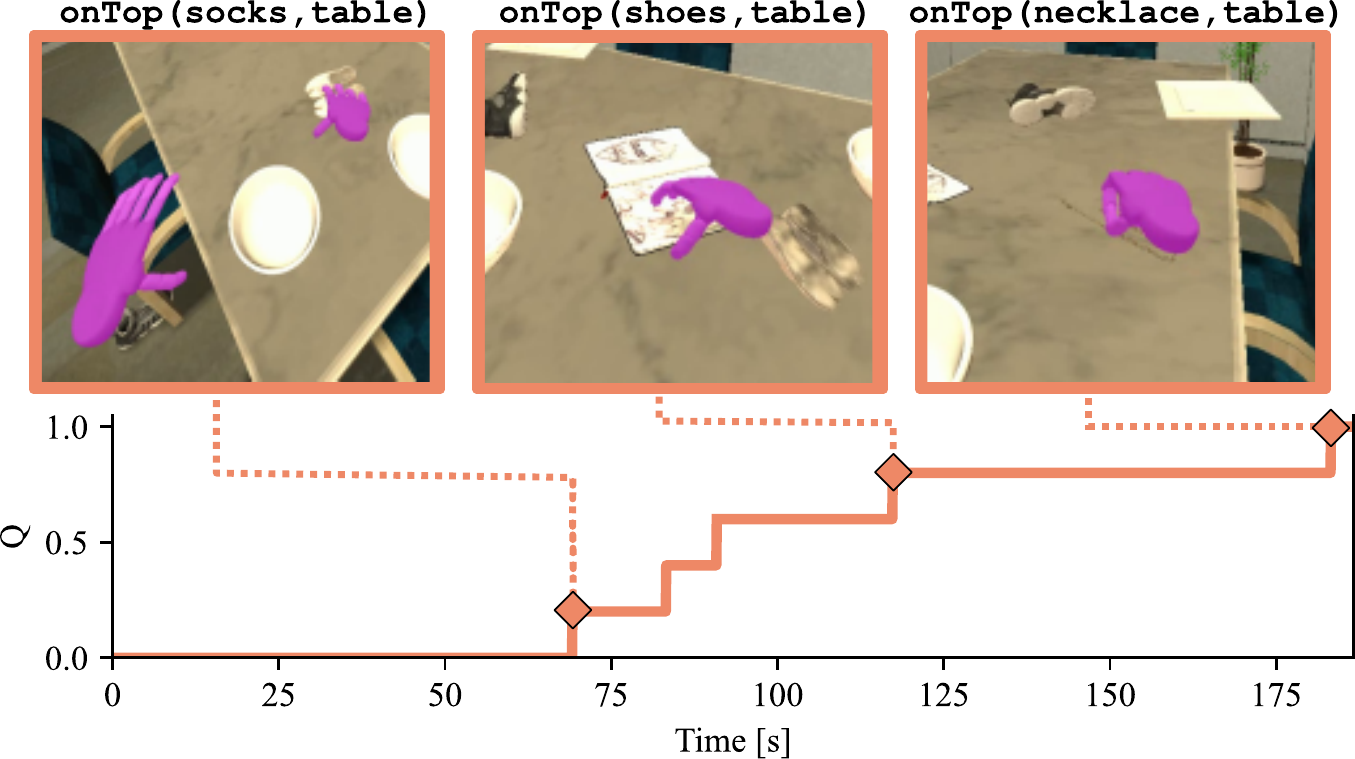}
        \includegraphics[width=0.49\linewidth]{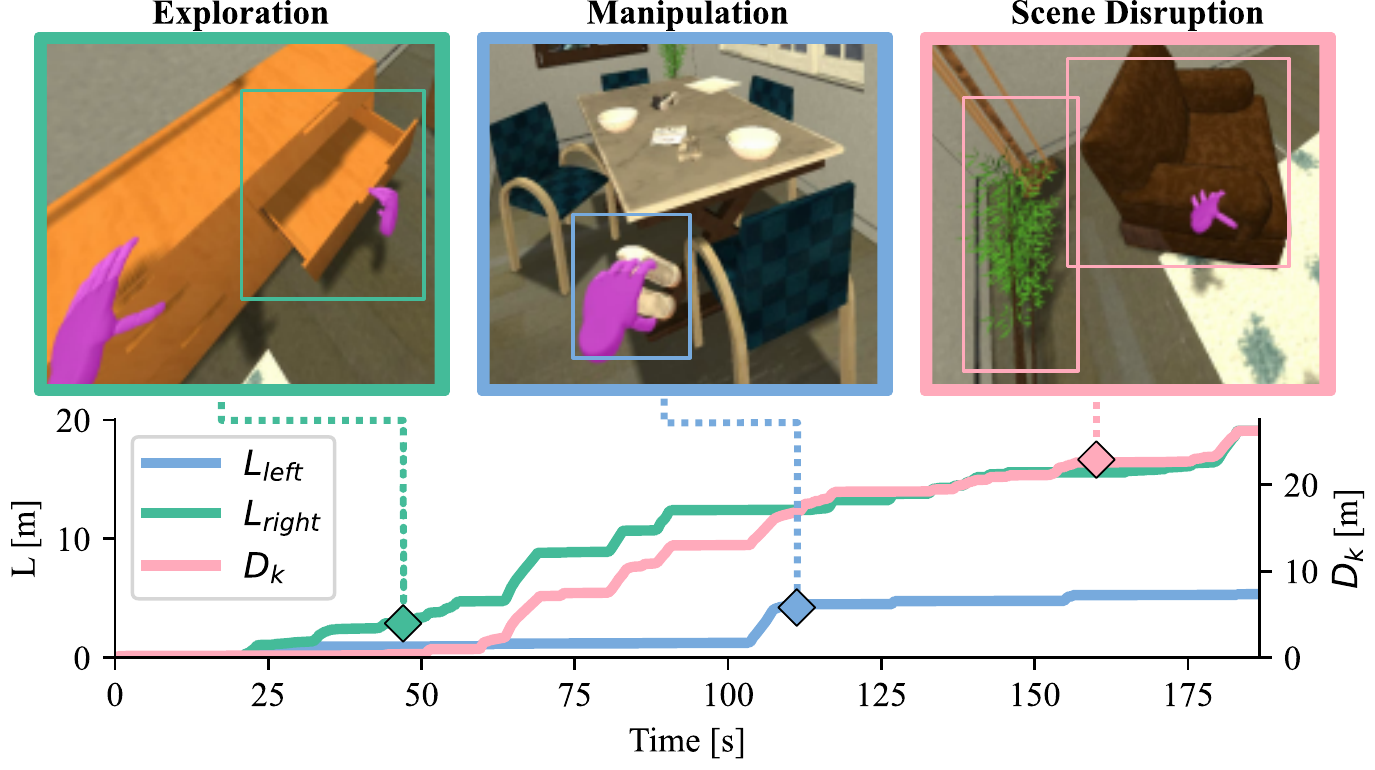}
        \caption{\textbf{Evaluation of human performance in \texttt{collect\_misplaced\_items}:} (\textit{Left}) success score, $Q$; (\textit{Right}) efficiency metrics: kinematic disarrangement, ($D_k$, dotted), hand interaction displacement ($L_\mathit{right}$, green, and $L_\mathit{left}$, blue); frames at the top depict significant events detected by the metrics; the success score detects the completion of activity-relevant steps; exploration, manipulation and scene disruption events are captured by the efficiency metrics that provide complementary information about the performance of the agent}
        \label{fig:efficiency}
        \vspace{-1em}
\end{figure}

An activity is complete when all literals in \textit{at least one} $C_i$ of its goal condition are satisfied, achieving $Q=1$ (100\%). Fig.~\ref{fig:efficiency} left depicts time evolution of $Q$ during an activity execution. $Q$ extends the fraction of objects in acceptable poses proposed as metric in~\cite{batra2020rearrangement}, generalized to any type of activity. 

\paragraph{Secondary Metrics -- Efficiency:}
Beyond success, efficiency is critical to evaluation; a successful solution in real-world tasks may be ineffective if it takes too long or causes scene disruption. %
We propose six secondary metrics that complement the primary metric (see Fig.~\ref{fig:efficiency}, right, for examples):
\begin{itemize}[
    nosep,
    noitemsep,
    leftmargin=2pt,
    itemindent=12pt]
    \item \textit{Simulated time}, $T_\mathit{sim}$: Accumulated time in simulation during execution as the number of simulated steps times the average simulated time per step. $T_\mathit{sim}$ is independent of the computer used.
    \item \textit{Kinematic disarrangement}, $\mathit{D}_\textit{K}$: Displacement caused by the agent in the environment. This can be \textit{accumulated} over time, or \textit{differential}, i.e. computed between two time steps, e.g. initial, final.
    \item \textit{Logical disarrangement},
    $\mathit{D}_\textit{L}$: Amount of changes caused by the agent in the logical state of the environment. This can be \textit{accumulated} over time or \textit{differential} between two time steps.%
    \item \textit{Distance navigated}, $L_\mathit{body}$: Accumulated distance traveled by the agent's base body. This metric evaluates the efficiency of the agent in navigating the environment.
    \item \textit{Displacement of hands}, $L_\mathit{left}$ and $L_\mathit{right}$: Accumulated displacement of each of the agent's hands while in contact with another object for manipulation (i.e., grasping, pushing, etc). This metric evaluates the efficiency of the agent in its interaction with the environment.
\end{itemize}

These efficiency metrics above can be quantified in absolute units (e.g., distance, time) for scene- and activity-specific comparisons (\textbf{general efficiency}). To enable fair comparisons cross diverse activities in \model, we also propose normalization relative to human performance (\textbf{human-centric efficiency}); given a human demonstration for an activity instance in VR, each secondary metric can be expressed as a \textit{fraction of the maximum human performance} on that metric.

For this purpose, we present the \model Dataset of Human Demonstrations with \numDemos successful demonstrations of \model activities in VR (\totalTime min). Humans are immersed in \ig, controlling the same embodiment used by the AI agents (details in Sec.~\ref{ss:hd_vr_appendix}). 
The dataset includes a complete record of human actions including manipulation, navigation, and gaze tracking data (Fig.~\ref{fig:vrstatistics}, Fig.~\ref{fig:trajectory_map}, and Fig.~\ref{fig:gaze_stats}), supporting analysis and subactivity segmentation (Fig.~\ref{fig:task_segmentation}). Sec.~\ref{sss:vr_stats2} presents a comprehensive analysis of these data; we quantify human performance in \model efficiency metrics (see Fig.~\ref{fig:vrstatistics}), and Fig.~\ref{fig:vrstatistics2} provides a further decomposition of room occupancy and hand usage across each BEHAVIOR activity.
To our knowledge, this is the largest available dataset of human behavior in VR; these data can facilitate development of new solutions for embodied AI (e.g., imitation learning) and also support studies of human cognition, planning, and motor control in ecological environments.

\section{Evaluating Reinforcement Learning in \model}
\label{s:eval}

In this section, we aim to experimentally demonstrate the challenges imposed by \model's realism, diversity, and complexity by evaluating the performance of some current state-of-the-art embodied AI solutions. 
While \model is a benchmark for all kinds of embodied AI methods, here we evaluate two reinforcement learning (RL) algorithms that have demonstrated excellent results in simpler embodied AI tasks with continuous or discrete action spaces~\cite{martin2019variable,vinyals2019grandmaster,andrychowicz2020learning,rajeswaran2017learning,hwangbo2019learning,akkaya2019solving,haarnoja2018learning,kiran2021deep}: Soft-Actor Critic (SAC~\cite{haarnoja2018soft}) and Proximal-Policy Optimization (PPO~\cite{schulman2017proximal}). We use SAC to train policies in the original low-level continuous action space of the agent, and PPO for experiments using our implemented action primitives (for details on the agents, see Sec.~\ref{s:instan}). Due to limited computational resources, we run our evaluation on the 12 most simple activities (based on involved types of state changes) until convergence.
Reward is given by our staggered success score $Q$.
We use as input to the policies a subset of the realistic agent's observations, RGB, depth and proprioception (excluding LiDAR, segmentation, etc.). Sec.~\ref{ss:expdetails} includes more experimental details.

\paragraph{Results in the original activities:} The first row of Table~\ref{tab:table_results1_am1} shows the results of SAC (mean $Q$ at the end of training for 3 seeds) on the original 12 activities with the standard setup: realistic robot actions and onboard sensing. Even for these ``simpler'' activities, \model is too great a challenge: the training agents do not fulfill any predicate in the goal condition ($Q=0$). In the following, we will analyze how each dimension of difficulty (realism, diversity, complexity) contributes to these results.

\paragraph{Effect of complexity (activity length):} In the first experiment, we evaluate the impact of the activity complexity (time length) in robot learning performance.
First, we evaluate the performance of an RL algorithm using our implemented action primitives based on motion planning. These are temporally extended actions that effectively shorten the horizon and length of the activity. 
The results of training with PPO are depicted in the second row of Table~\ref{tab:table_results1_am1}. Even in these simpler conditions, agents fail in all but one activity ({\small\texttt{bringingInWood}}, $Q=0.13$). 
In a second oracle-driven experiment, we take a successful human demonstration for each activity from the \model Dataset and save the state of the environment a few seconds before its successful execution at $T$. We use this as initial state and train agents with SAC: rows 3 to 6 of Table~\ref{tab:table_results1_am1} show the mean success rate ($\mathit{SR}$, full accomplishment of the activity) in 100 evaluation episodes for the final policy resulting from training with three different random seeds ($Q$ starts here close to 1 and is less informative). Even when starting \SI{1}{\second} away from a goal state, most learning agents fail to achieve the tasks. A few achieve better success but their performance decreases quickly as we start further away from the successful execution, being zero for all activities at \SI{10}{\second}. This indicates that the long-horizon of the activities in \model is in fact a paramount challenge for reinforcement learning.
We hypothesize that Embodied AI solutions with a hierarchical structure such as hierarchical-RL or task-and-motion-planning (TAMP) may help to overcome the challenges of high complexity (length) of the \model activities~\cite{barto2003recent,vezhnevets2017feudal,li2020hrl4in,garrett2015ffrob}.

\begin{wraptable}{r}{0.6\textwidth}
%\begin{table}
\vspace{-2.5em}
    \centering
    \resizebox{\linewidth}{!}{
    \begin{tabular}{|c|c|cccccccccccc|}
    \multicolumn{2}{c}{} & \rot{\footnotesize{bringingInWood}}& \rot{\footnotesize{collectMisplacedItems}}& \rot{\footnotesize{movingBoxesToStorage}}& \rot{\footnotesize{organizingFileCabinet}}& \rot{\footnotesize{throwingAwayLeftovers}}& \rot{\footnotesize{puttingDishesAway}}& \rot{\footnotesize{puttingleftoversAway}}& \rot{\footnotesize{re-shelvingLibraryBooks}}& \rot{\footnotesize{layingTileFloors}}& \rot{\footnotesize{settingUpCandles}}& \rot{\footnotesize{pickingUpTrash}}& \rot{\footnotesize{storingFood}}\\\cline{2-14}
    \multicolumn{1}{c|}{} & $Q^\mathit{ca}$ & 0 & 0 & 0 & 0 & 0 & 0 & 0 & 0 & 0 & 0 & 0 & 0 \\\hline
    \multirow{5}*{\rotatebox{90}{complexity}} & \cellcolor[rgb]{.8,.8,.8}$Q^\mathit{ap}$ & \cellcolor[rgb]{.8,.8,.8}0.13 & \cellcolor[rgb]{.8,.8,.8}0 & \cellcolor[rgb]{.8,.8,.8}0 & \cellcolor[rgb]{.8,.8,.8}0 & \cellcolor[rgb]{.8,.8,.8}0 & \cellcolor[rgb]{.8,.8,.8}0 & \cellcolor[rgb]{.8,.8,.8}0 & \cellcolor[rgb]{.8,.8,.8}0 & \cellcolor[rgb]{.8,.8,.8}0 & \cellcolor[rgb]{.8,.8,.8}0 & \cellcolor[rgb]{.8,.8,.8}0 & \cellcolor[rgb]{.8,.8,.8}0 \\\cline{2-14}
     & $\mathit{SR}^\mathit{ca}$@${T-\SI{1}{\second}}$ & 1 & 1 & 1 & 0 & 0 & 0 & 0 & 0 & 1 & 1 & 0.97 & 1\\\cline{2-14}
    & $\mathit{SR}^\mathit{ca}$@${T-\SI{2}{\second}}$ & 1 & 0.07 & 1 & 0 & 0 & 0 & 0 & 0 & 1 & 1 & 0.01 & 0\\\cline{2-14}
    & $\mathit{SR}^\mathit{ca}$@${T-\SI{3}{\second}}$ & 1 & 0.21 & 1 & 0 & 0 & 0 & 0 & 0 & 1 & 0.01 & 0 & 0\\\cline{2-14}
    & $\mathit{SR}^\mathit{ca}$@${T-\SI{10}{\second}}$ & 0 & 0 & 0 & 0 & 0 & 0 & 0 & 0 & 0 & 0 & 0 & 0\\\hline
    \multirow{4}*{\rotatebox{90}{realism}} & \cellcolor[rgb]{1,1,1}$Q^\mathit{ca}_\mathit{FullObs}$ & \cellcolor[rgb]{1,1,1}0 & \cellcolor[rgb]{1,1,1}0 & \cellcolor[rgb]{1,1,1}0 & \cellcolor[rgb]{1,1,1}0 & \cellcolor[rgb]{1,1,1}0 & \cellcolor[rgb]{1,1,1}0 & \cellcolor[rgb]{1,1,1}0 & \cellcolor[rgb]{1,1,1}0 & \cellcolor[rgb]{1,1,1}0 & \cellcolor[rgb]{1,1,1}0 & \cellcolor[rgb]{1,1,1}0 & \cellcolor[rgb]{1,1,1}0\\\cline{2-14}
    & \cellcolor[rgb]{.8,.8,.8}$Q^\mathit{ap}_\mathit{FullObs}$ & \cellcolor[rgb]{.8,.8,.8} 0.20 & \cellcolor[rgb]{.8,.8,.8}0.02 & \cellcolor[rgb]{.8,.8,.8}0.49 & \cellcolor[rgb]{.8,.8,.8}0 & \cellcolor[rgb]{.8,.8,.8}0 & \cellcolor[rgb]{.8,.8,.8}0 & \cellcolor[rgb]{.8,.8,.8}0 & \cellcolor[rgb]{.8,.8,.8}0.13 & \cellcolor[rgb]{.8,.8,.8}0 & \cellcolor[rgb]{.8,.8,.8}0.09 & \cellcolor[rgb]{.8,.8,.8}0 & \cellcolor[rgb]{.8,.8,.8}0\\\cline{2-14}
    & \cellcolor[rgb]{.8,.8,.8}$Q^\mathit{ap}_\mathit{NoPhys}$ & \cellcolor[rgb]{.8,.8,.8}0.92 & \cellcolor[rgb]{.8,.8,.8} 0.47 & \cellcolor[rgb]{.8,.8,.8}0.73 & \cellcolor[rgb]{.8,.8,.8}0 & \cellcolor[rgb]{.8,.8,.8}0.32 & \cellcolor[rgb]{.8,.8,.8}0.55 & \cellcolor[rgb]{.8,.8,.8}0.44 & \cellcolor[rgb]{.8,.8,.8}0.04 & \cellcolor[rgb]{.8,.8,.8}0 & \cellcolor[rgb]{.8,.8,.8}0.27 & \cellcolor[rgb]{.8,.8,.8}0 & \cellcolor[rgb]{.8,.8,.8}0.32\\\cline{2-14}
    & \cellcolor[rgb]{.8,.8,.8}$Q^\mathit{ap}_\mathit{NoPhys, FullObs}$ & \cellcolor[rgb]{.8,.8,.8}1.0 & \cellcolor[rgb]{.8,.8,.8}0.95 & \cellcolor[rgb]{.8,.8,.8}0.83 & \cellcolor[rgb]{.8,.8,.8}0 & \cellcolor[rgb]{.8,.8,.8}0.56 & \cellcolor[rgb]{.8,.8,.8}0.94 & \cellcolor[rgb]{.8,.8,.8}0.55 & \cellcolor[rgb]{.8,.8,.8}0.56 & \cellcolor[rgb]{.8,.8,.8}0 & \cellcolor[rgb]{.8,.8,.8}0.5 & \cellcolor[rgb]{.8,.8,.8}0.67 & \cellcolor[rgb]{.8,.8,.8}1.0\\\hline
    \end{tabular}
  }
\captionof{table}{\textbf{Evaluation of state-of-the-art RL algorithms on \model} \textit{Fully realistic, diverse and complex (row 1):} SAC~\cite{haarnoja2018soft} for visuomotor continuous actions (superindex $\mathit{ca}$) performs poorly in all activities; \textit{Complexity analysis (rows 2-6):} reducing complexity (horizon) with temporally extended action primitives (superindex $\mathit{ap}$ and gray cells, trained with PPO~\cite{schulman2017proximal}) or by starting few seconds away from a goal state, lead to some non-zero success rate ($\mathit{SR}$). \textit{Realism analysis (rows 7-10):} Only by reducing realism in observations and physics, and complexity through action primitives, RL achieves significant success in a handful of the activities.}
\label{tab:table_results1_am1}
\vspace{-1em}
\end{wraptable}

\paragraph{Effect of realism (in sensing and actuation):}
In a third experiment, we evaluate how much the realism in actuation and sensing affects the performance of embodied AI solutions.
To evaluate the effect of realistic observability of the \model activities in the performance of robot learning approaches, we train agents with continuous motion control (SAC), and motion primitives (PPO) assuming full-observability of the state.
Tables~\ref{tab:table_results1_am1} (rows 7-8, subindex $\mathit{FullObs}$) depict the results. We observe that even with full observability the complexity dominates policies in the original action space and they do not accomplish any part of the activities. For policies selecting among action primitives, there is some partial success only in five of the activities indicating that the perceptual problem is part of the difficulty in \model.
To evaluate the effect of realistic actuation, we train an agent using action primitives that execute without physics simulation, achieving their expected outcome (e.g. grasp an object, or place it somewhere). Tables~\ref{tab:table_results1_am1} (row 9-10, subindex $\mathit{noPhys}$) shows the results, also in combination with unrealistic full-observability. We observe that without the difficulties of realistic physics and actuation, the learning agents achieve an important part of most activities, accomplishing consistently two of them ($Q=1$) when full-observability of the state is also granted.
This indicates that the generation of the correct actuation is a critical challenge for embodied AI solutions, even when they infer the right next step at the task-planning level, supporting the importance of benchmarks with realistically action execution over predefined action outcomes.

\begin{wraptable}{r}{0.6\textwidth}
%\begin{table}
\vspace{-1em}
    \centering
    \resizebox{\linewidth}{!}{
    \begin{tabular}{|c|c|c|c|c|c|c|}
    \cline{1-2}
    \multicolumn{2}{|c|}{Diversity in\ldots} & \multicolumn{5}{c}{}\\
    \hline
    object poses & object instances & \texttt{ontop} & \texttt{sliced} & \texttt{soaked} & \texttt{stained} & \texttt{cooked}\\
    \hline
    \textcolor{red}\faTimes & \textcolor{red}\faTimes & 1 & 0.15 & 1 & 1 & 1 \\
    \hline
    \textcolor{indiagreen}\faCheck & \textcolor{red}\faTimes & 0.825 & 0 & 0.935 & 0.28 & 0.66 \\
    \hline
    \textcolor{indiagreen}\faCheck & \textcolor{indiagreen}\faCheck & 0.46 & 0 & 0.925 & 0.11 & 0.265 \\
    \hline
    \end{tabular}
  }
\captionof{table}{\textbf{Evaluation of the effect of \model's diversity:} Results of training agents with SAC~\cite{haarnoja2018soft} in single-predicate activities of increasing diversity; Even in these simple activities, performance degrades quickly indicating that current state-of-the-art cannot cope with the dimensions of diversity spanned in \model}
\label{tab:diversity_exp}
\end{wraptable}

\paragraph{Effect of diversity (in activity instance and objects):} Another cause of the poor performance of robot learning solutions in the 12 \model activities may be the high diversity in multiple dimensions, such as scenes, objects, and initial states.
This diversity forces embodied AI solutions to generalize to all possible conditions.
In a second experiment, we evaluate the effect of \model's diversity on performance.
To present diversity across activities while alleviating their complexity, we train RL agents to complete five single-literal activities involving only one or two objects.
Note that these activities are not part of \model. We evaluate training with RL (SAC) for each activity under diverse instantiations: initialization of the activity (object poses) and object instances. The results are shown in Table~\ref{tab:diversity_exp}, where we report $Q$. First, we train without any diversity as baseline to understand the ground complexity of the single-literal activities. All agents achieve success. Then, we evaluate how well the RL policies train for a diverse set of instances of the activities, first changing objects' initial pose, then changing the object. Performance in all activities decreases rapidly, especially in {\small\texttt{sliced}} and {\small\texttt{stained}}. These experiments indicate that the diversity in \model goes beyond what current RL algorithms can handle even in simplified activities, and poses a challenge for generalization in embodied AI.

\section{Conclusion and Future Work}
\label{sec:conclusion}

We presented \model, a novel benchmark for embodied AI solutions of household activities. 
\model presents  \numActivities realistic, diverse and complex activities with a new logic-symbolic representation, a fully functional simulation-based implementation, and a set of human-centric metrics based on the performance of humans on the same activities in VR.
The activities push the state-of-the-art in benchmarking adding new types of state changes that the agent needs to be able to cause, such as cleaning surfaces or changing object temperatures.
Our experiments with two state-of-the-art RL baselines shed light on the challenges presented by \model's level of realism, diversity and complexity. \model will be open-source and free to use; we hope it facilitates participation and fair access to research tools, and paves the way towards a new generation of embodied AI.%

\acknowledgments{
We would like to thank Bokui Shen, Xi Jia Zhou, and Jim Fan for comments, ideas, and support in data collection. % todo expand
This work is in part supported by Toyota Research Institute (TRI), ARMY MURI grant W911NF-15-1-0479, Samsung, Amazon, and Stanford Institute for Human-Centered AI (SUHAI). % todo 
%S. S. is supported by the National Science Foundation Graduate Research Fellowship Program (NSF GRFP). S. B. is supported by a National Defense Science and Engineering Graduate (NDSEG) fellowship. % todo expand
S. S. and C. L. are supported by SUHAI Award \#202521. S. S. is also supported by the National Science Foundation Graduate Research Fellowship Program (NSF GRFP). R. M-M. and S. B. are supported by SAIL TRI Center – Award \# S-2018-28-Savarese-Robot-Learn. S. B. is also supported by a National Defense Science and Engineering Graduate (NDSEG) fellowship, SAIL TRI Center – Award \# S-2018-27-Niebles, and SAIL TRI Center – Award \# TRI Code 44. S. S. and S. B. are supported by a Department of Navy award (N00014-16-1-2127) issued by the Office of Naval Research (ONR). F. X. is supported by the Qualcomm Innovation Fellowship and Stanford Graduate Fellowship. 
This article solely reflects the opinions and conclusions of its authors and not any other entity.

}
{
\bibliography{bibl.bib}
}

\newpage

\renewcommand{\thesection}{\Alph{section}}
\renewcommand{\thesubsection}{A.\arabic{subsection}}

\setcounter{figure}{0} \renewcommand{\thefigure}{A.\arabic{figure}}

\setcounter{table}{0}
\renewcommand{\thetable}{A.\arabic{table}}

\newpage

\section*{Appendix for BEHAVIOR: Benchmark for Everyday Household Activities in Virtual, Interactive, and Ecological Environments}
\label{s:appendix}

\subsection{Visualizing \numActivities \model Activities}
\label{ss:activityviz}

\newcommand{\tikzpicbis}[3]{
\begin{scope}[xshift=#1]
\node [below right, outer sep=0pt, inner sep=0] (image) at (0,0){\includegraphics[width=0.115\textwidth]{figures/acts_mosaic/#2}};%
\node [below right, outer sep=0pt, inner sep=0] (image) at
(0.12\textwidth,0){\includegraphics[width=0.115\textwidth]{figures/acts_mosaic/real_life/#2}};%
\node [font=\fontsize{6pt}{5}\selectfont, below,inner sep=1pt, align=center, minimum width=0.24\textwidth] at (0.115\textwidth, -0.09\textwidth){#3};%
\end{scope}
}%
\newcommand{\myheightbtwtikz}{2pt}
\begin{figure}[H]
\begin{tikzpicture}
\tikzpicbis{0.00\textwidth}{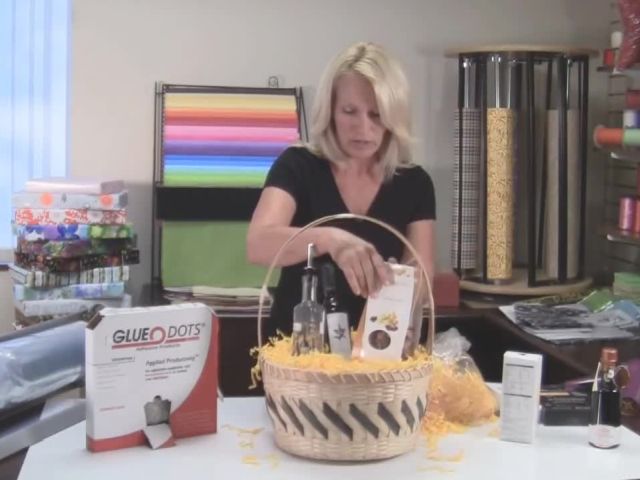}{Assembling gift baskets}%
\tikzpicbis{0.25\textwidth}{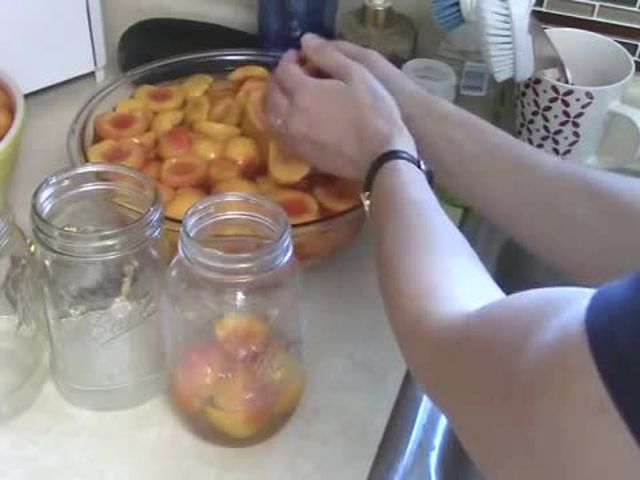}{Bottling fruit}%
\tikzpicbis{0.50\textwidth}{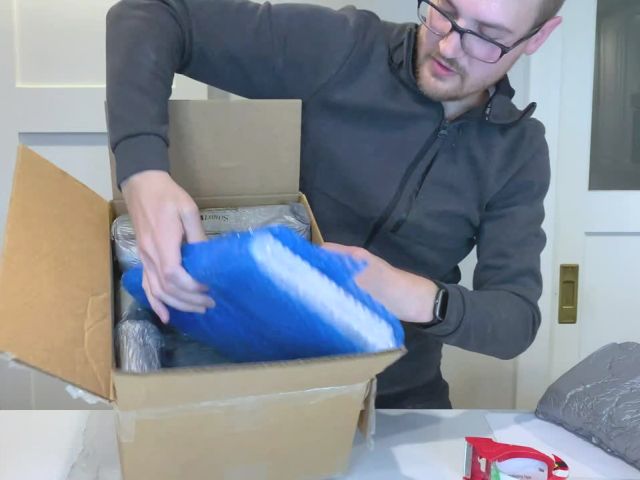}{Boxing books up for storage}%
\tikzpicbis{0.75\textwidth}{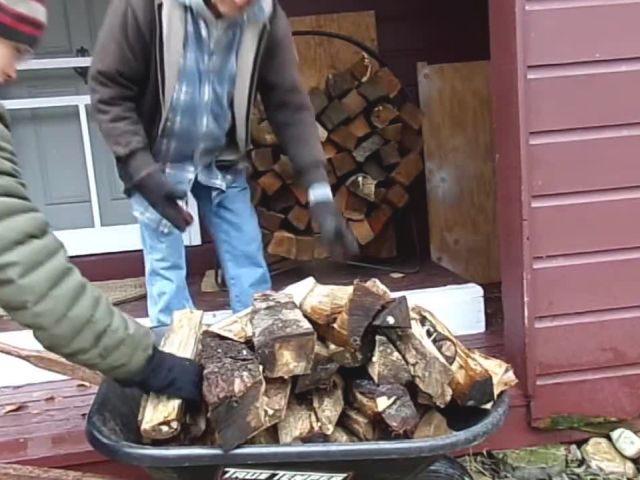}{Bringing in wood}%
\end{tikzpicture}%
\vspace{\myheightbtwtikz}
\begin{tikzpicture}
\tikzpicbis{0.00\textwidth}{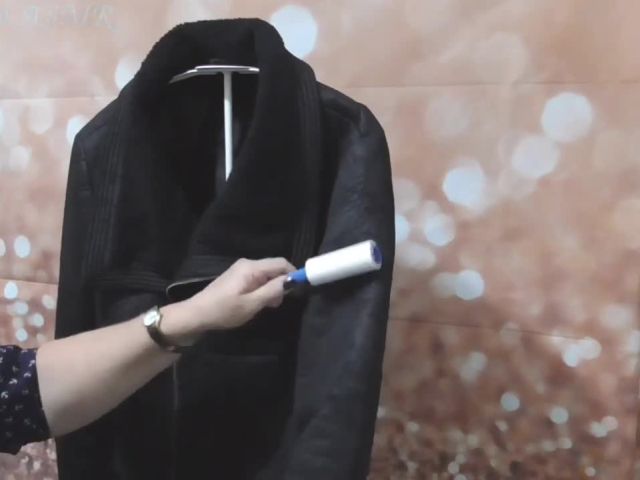}{Brushing lint off clothing}%
\tikzpicbis{0.25\textwidth}{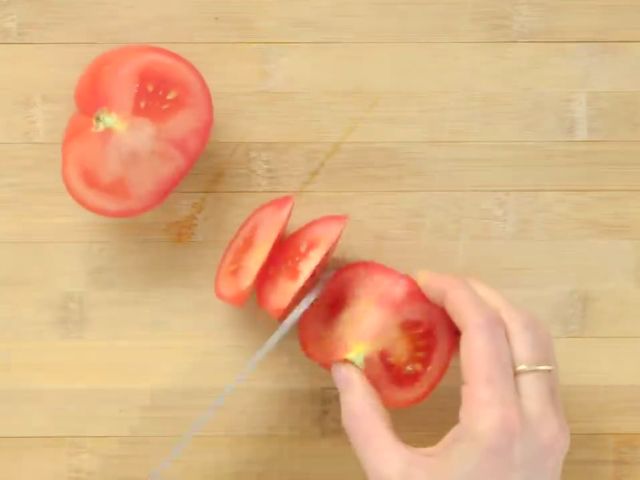}{Chopping vegetables}%
\tikzpicbis{0.50\textwidth}{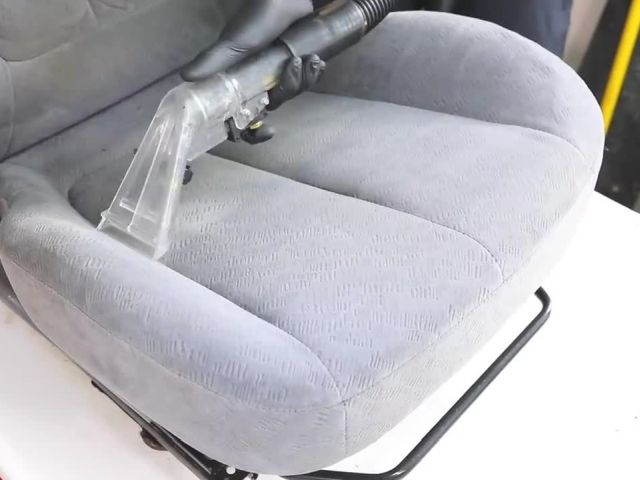}{Cleaning a car}%
\tikzpicbis{0.75\textwidth}{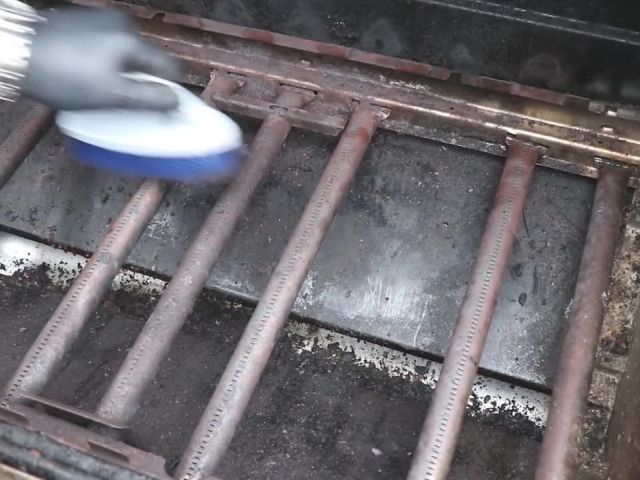}{Cleaning barbecue grill}%
\end{tikzpicture}%
\vspace{\myheightbtwtikz}
\begin{tikzpicture}
\tikzpicbis{0.00\textwidth}{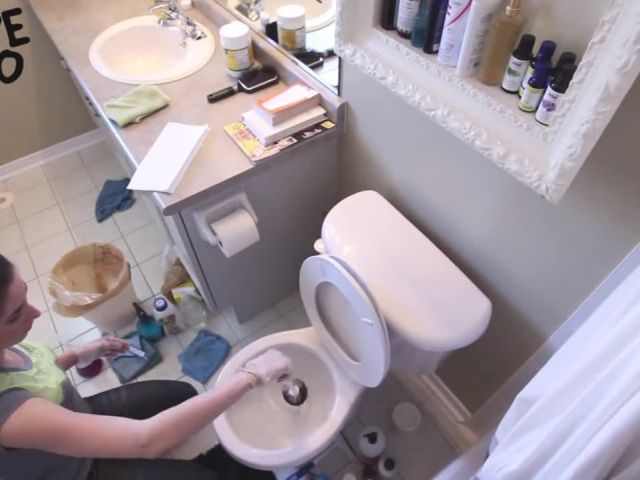}{Cleaning bathrooms}%
\tikzpicbis{0.25\textwidth}{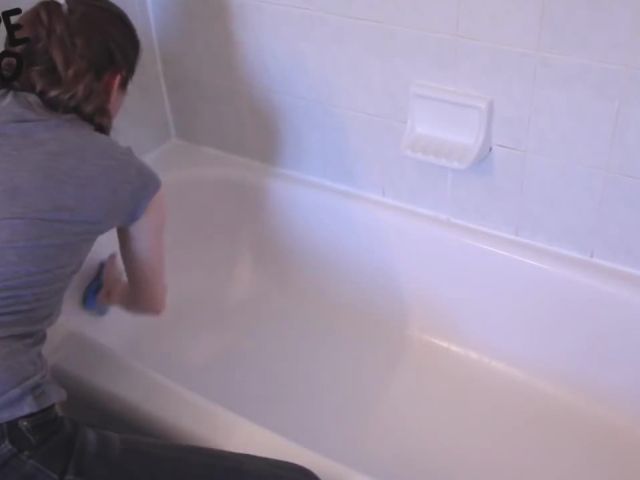}{Cleaning bathtub}%
\tikzpicbis{0.50\textwidth}{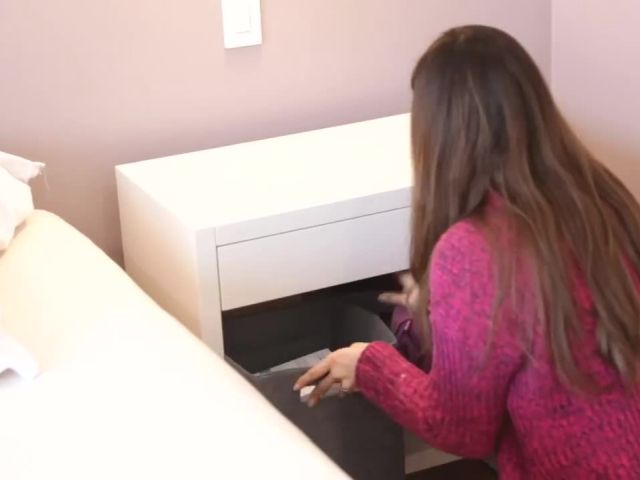}{Cleaning bedroom}%
\tikzpicbis{0.75\textwidth}{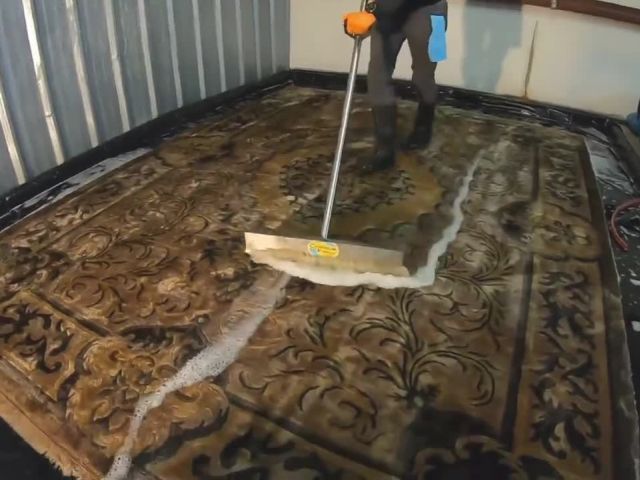}{Cleaning carpets}%
\end{tikzpicture}%
\vspace{\myheightbtwtikz}
\begin{tikzpicture}
\tikzpicbis{0.00\textwidth}{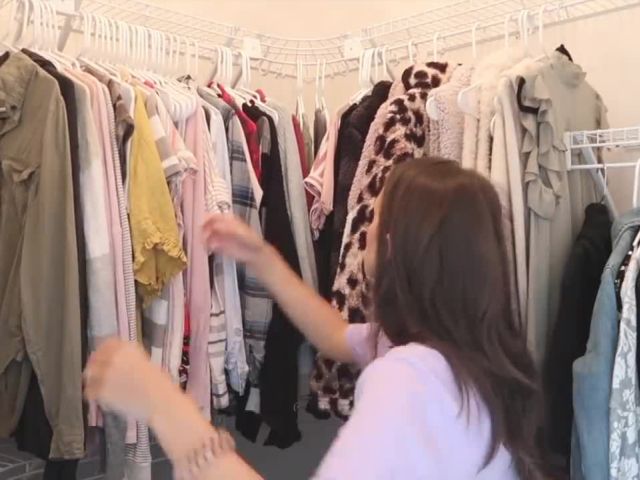}{Cleaning closet}%
\tikzpicbis{0.25\textwidth}{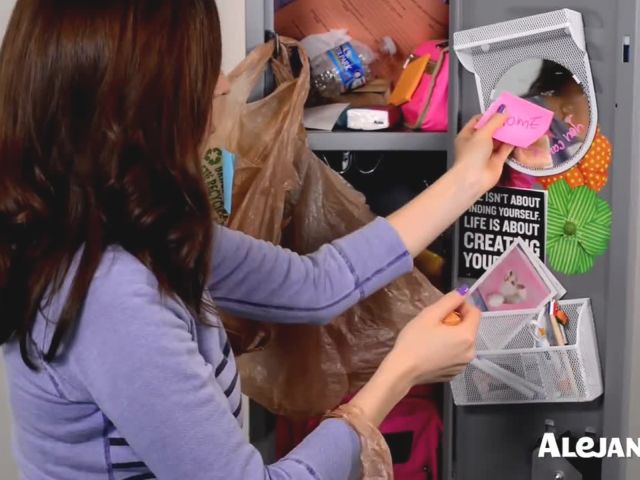}{Cleaning cupboards}%
\tikzpicbis{0.50\textwidth}{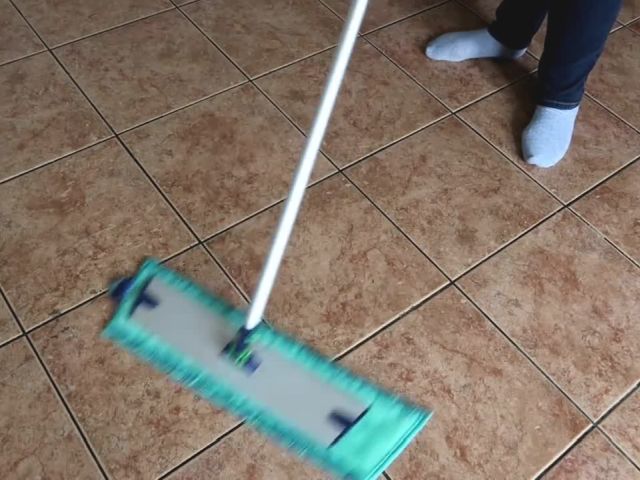}{Cleaning floors}%
\tikzpicbis{0.75\textwidth}{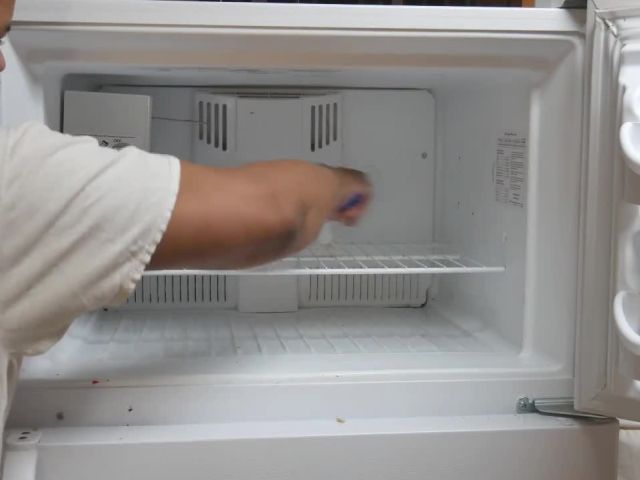}{Cleaning freezer}%
\end{tikzpicture}%
\vspace{\myheightbtwtikz}
\begin{tikzpicture}
\tikzpicbis{0.00\textwidth}{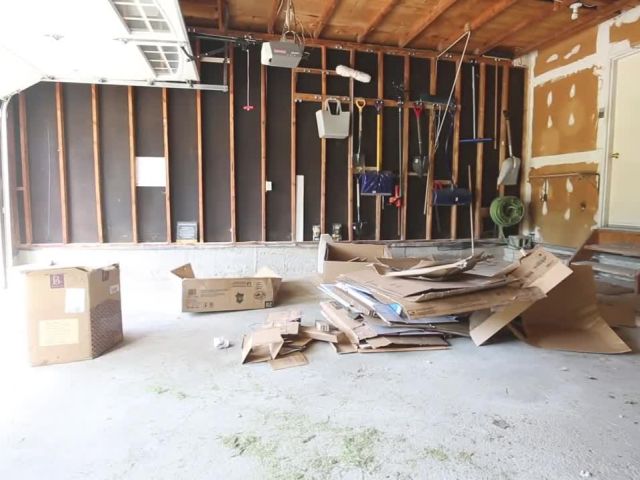}{Cleaning garage}%
\tikzpicbis{0.25\textwidth}{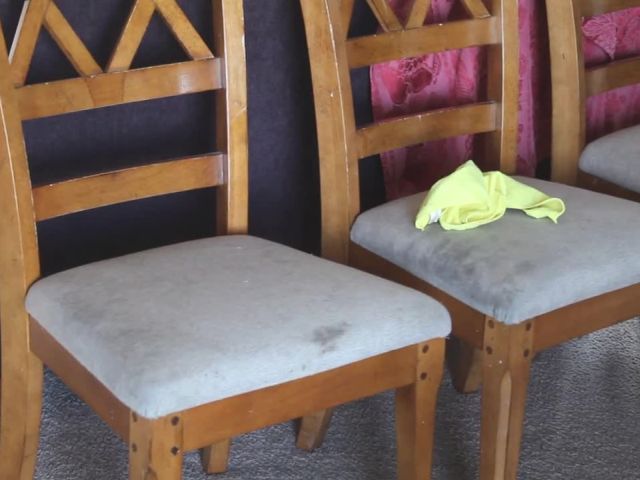}{Cleaning high chair}%
\tikzpicbis{0.50\textwidth}{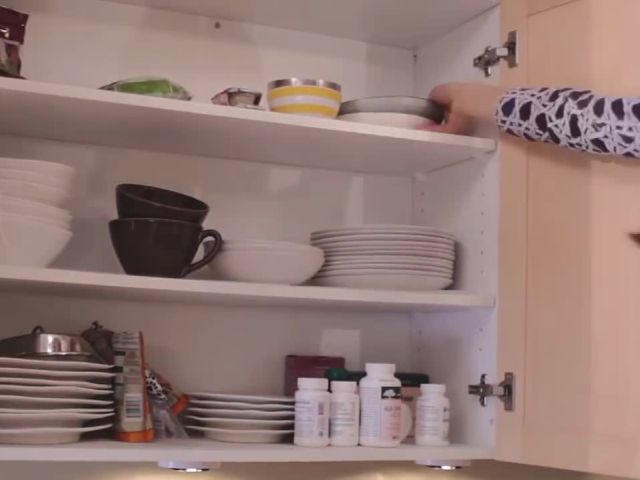}{Cleaning kitchen cupboard}%
\tikzpicbis{0.75\textwidth}{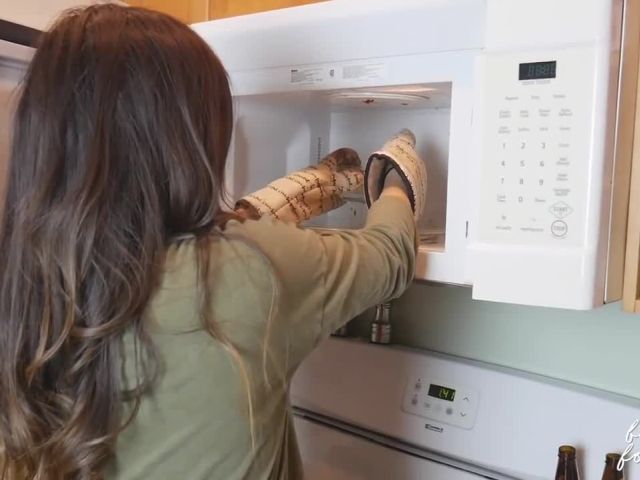}{Cleaning microwave oven}%
\end{tikzpicture}%
\vspace{\myheightbtwtikz}
\begin{tikzpicture}
\tikzpicbis{0.00\textwidth}{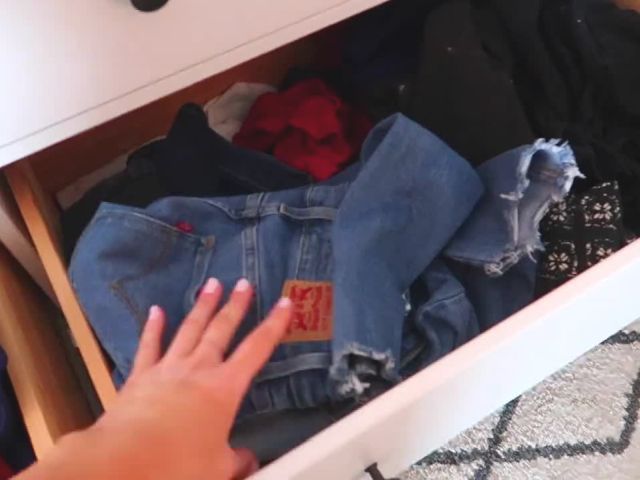}{Cleaning out drawers}%
\tikzpicbis{0.25\textwidth}{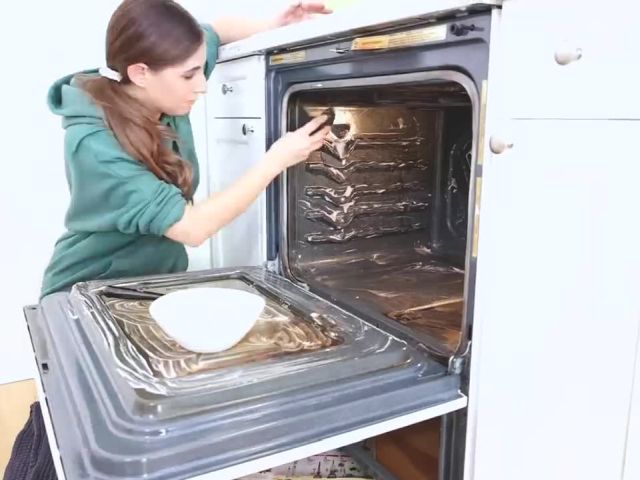}{Cleaning oven}%
\tikzpicbis{0.50\textwidth}{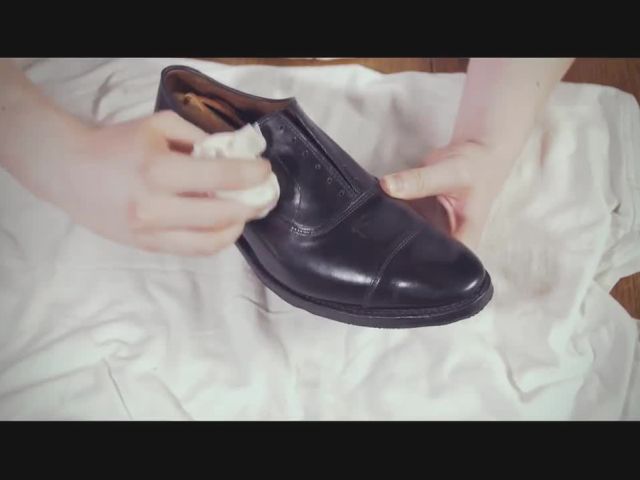}{Cleaning shoes}%
\tikzpicbis{0.75\textwidth}{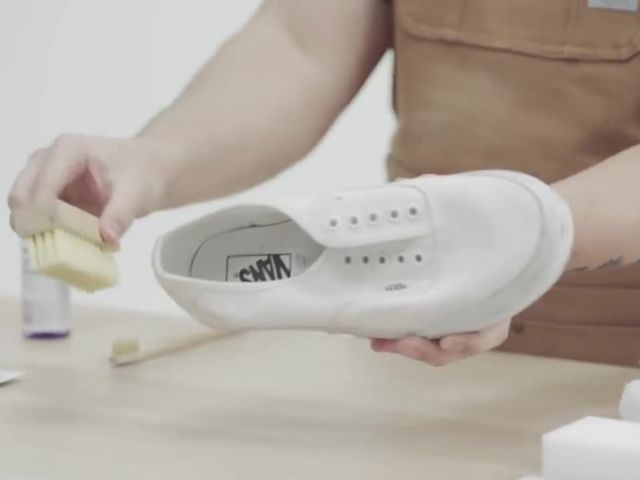}{Cleaning sneakers}%
\end{tikzpicture}%
\vspace{\myheightbtwtikz}
\begin{tikzpicture}
\tikzpicbis{0.00\textwidth}{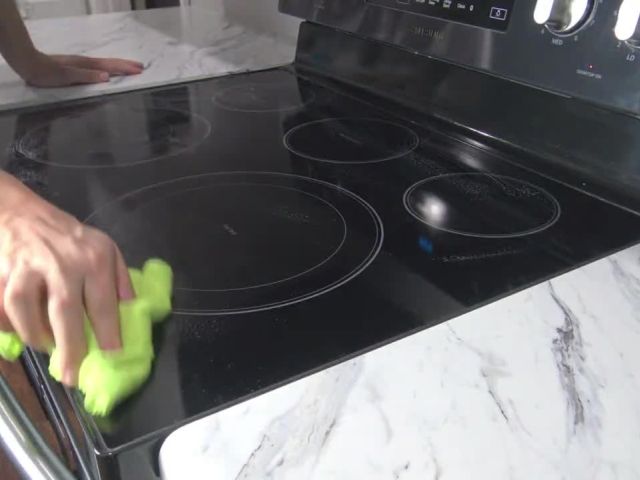}{Cleaning stove}%
\tikzpicbis{0.25\textwidth}{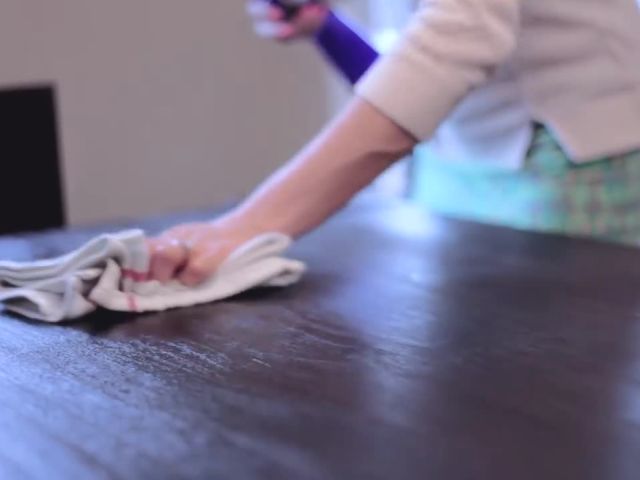}{Cleaning table after clearing}%
\tikzpicbis{0.50\textwidth}{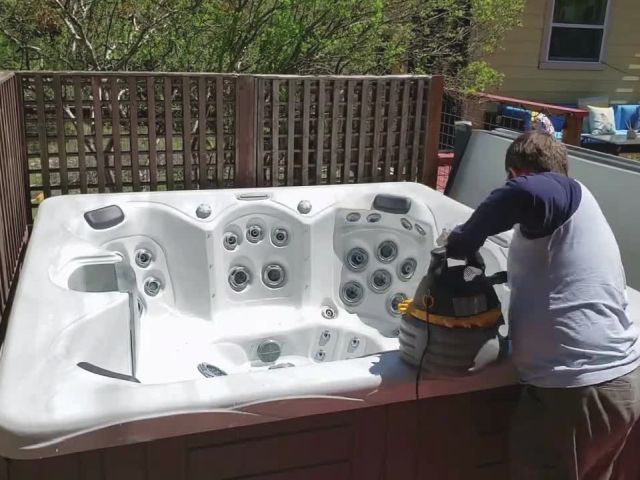}{Cleaning the hot tub}%
\tikzpicbis{0.75\textwidth}{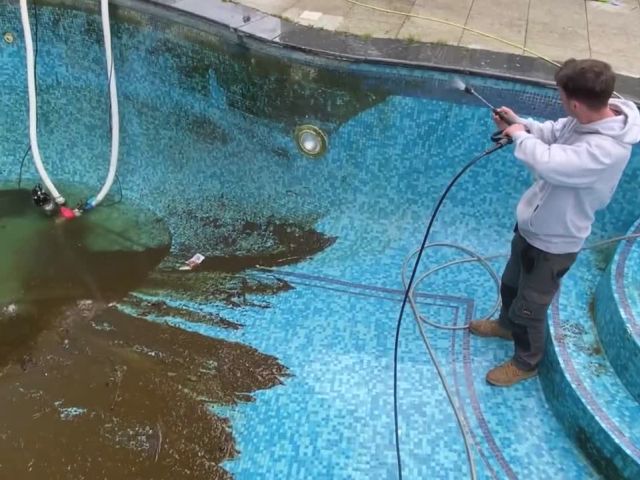}{Cleaning the pool}%
\end{tikzpicture}%
\vspace{\myheightbtwtikz}
\begin{tikzpicture}
\tikzpicbis{0.00\textwidth}{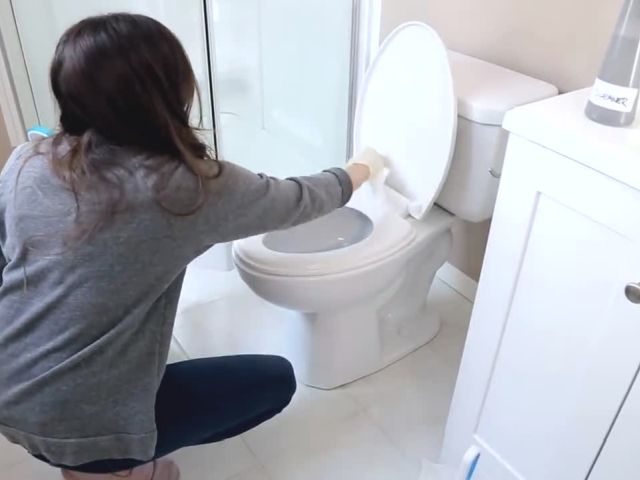}{Cleaning toilet}%
\tikzpicbis{0.25\textwidth}{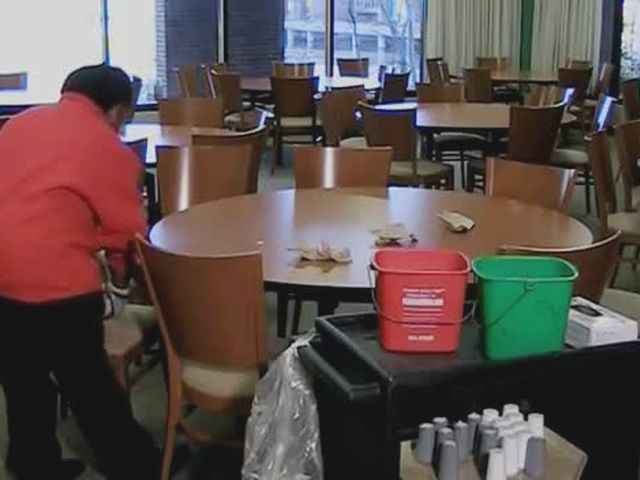}{Cleaning up after a meal}%
\tikzpicbis{0.50\textwidth}{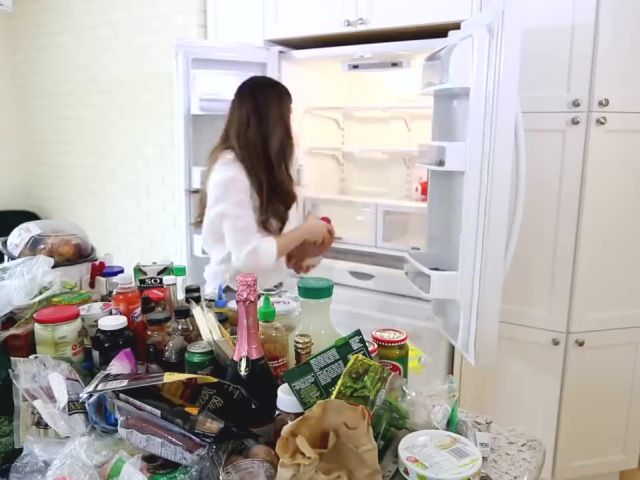}{Cleaning up refrigerator}%
\tikzpicbis{0.75\textwidth}{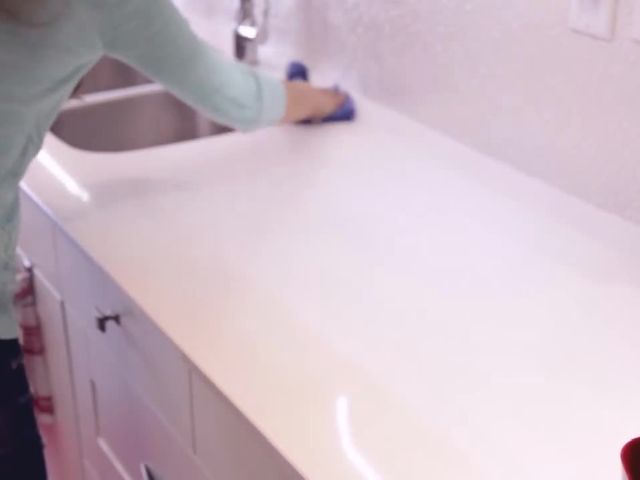}{Cleaning up the kitchen only}%
\end{tikzpicture}%
\vspace{\myheightbtwtikz}
\begin{tikzpicture}
\tikzpicbis{0.00\textwidth}{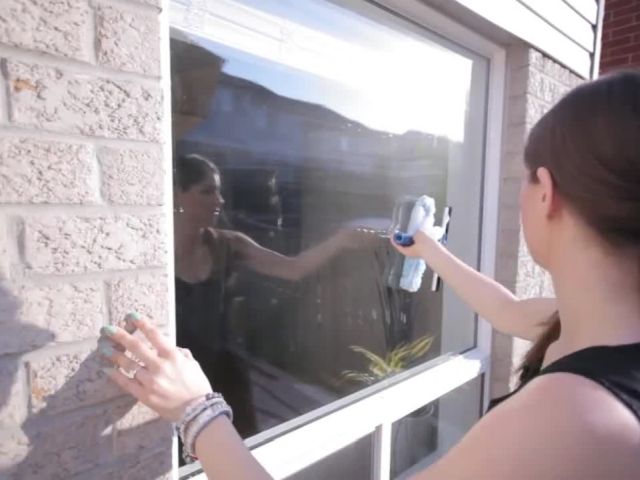}{Cleaning windows}%
\tikzpicbis{0.25\textwidth}{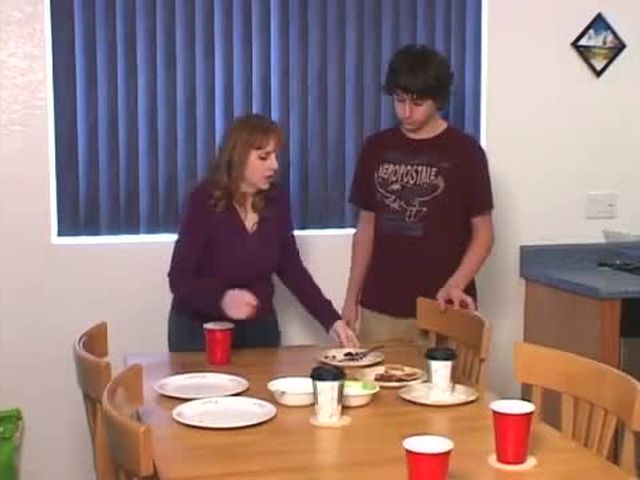}{Clearing dinner table}%
\tikzpicbis{0.50\textwidth}{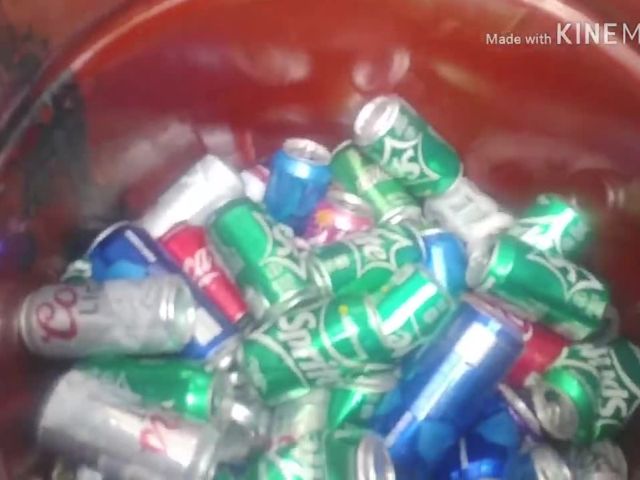}{Collecting aluminum cans}%
\tikzpicbis{0.75\textwidth}{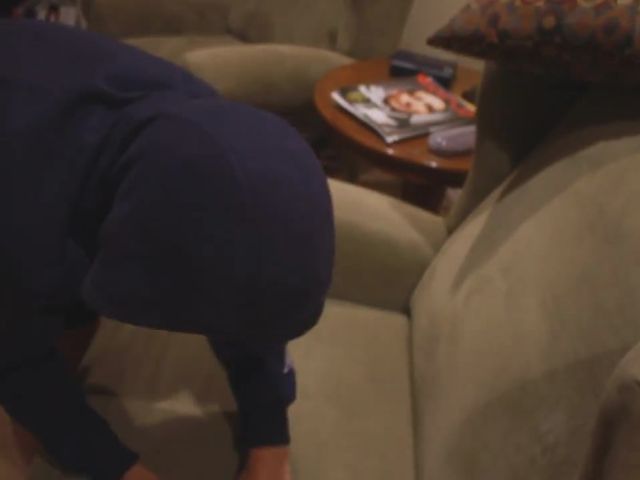}{Collect misplaced items}%
\end{tikzpicture}%
\vspace{\myheightbtwtikz}
\begin{tikzpicture}
\tikzpicbis{0.00\textwidth}{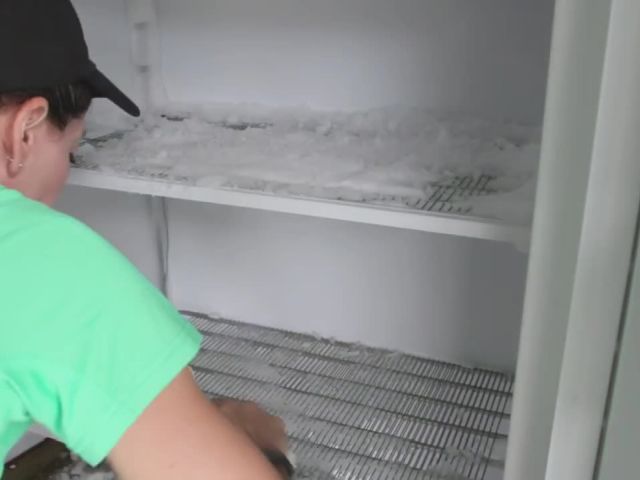}{Defrosting freezer}%
\tikzpicbis{0.25\textwidth}{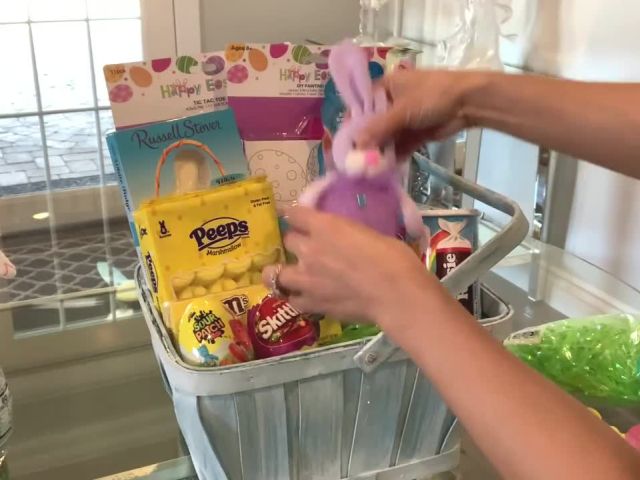}{Filling an Easter basket}%
\tikzpicbis{0.50\textwidth}{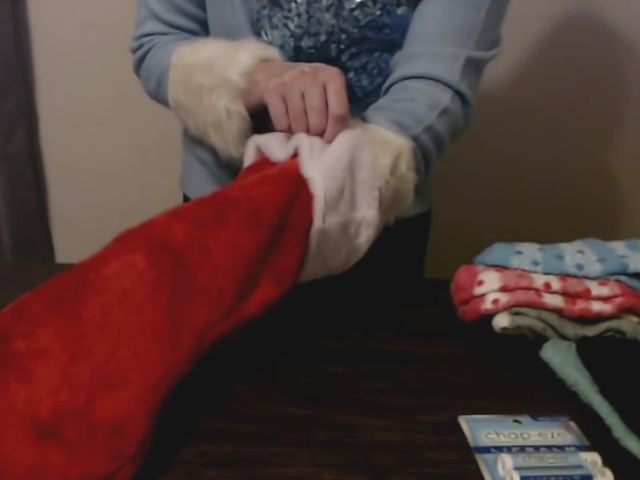}{Filling a Christmas stocking}%
\tikzpicbis{0.75\textwidth}{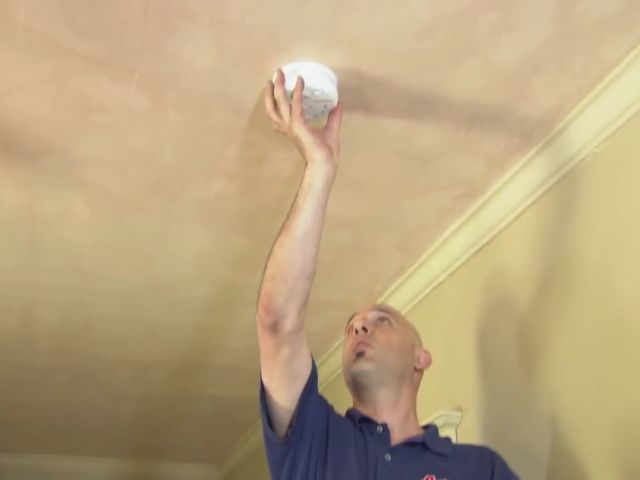}{Installing alarms}%
\end{tikzpicture}%
\vspace{\myheightbtwtikz}
\begin{tikzpicture}
\tikzpicbis{0.00\textwidth}{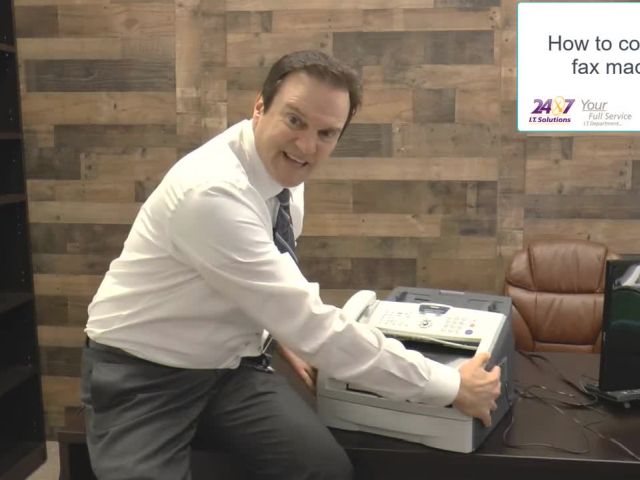}{Installing a fax machine}%
\tikzpicbis{0.25\textwidth}{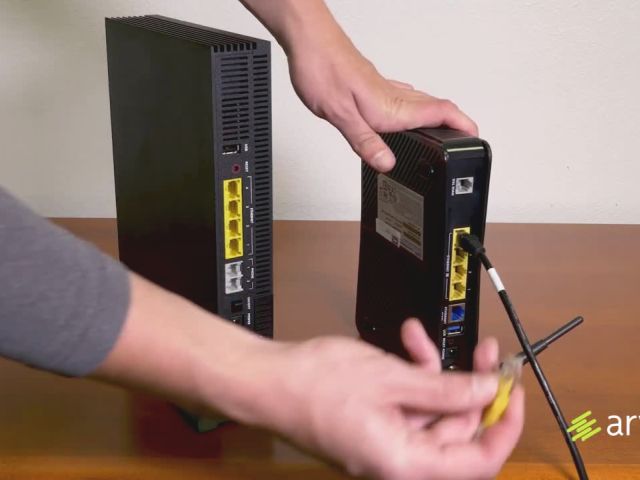}{Installing a modem}%
\tikzpicbis{0.50\textwidth}{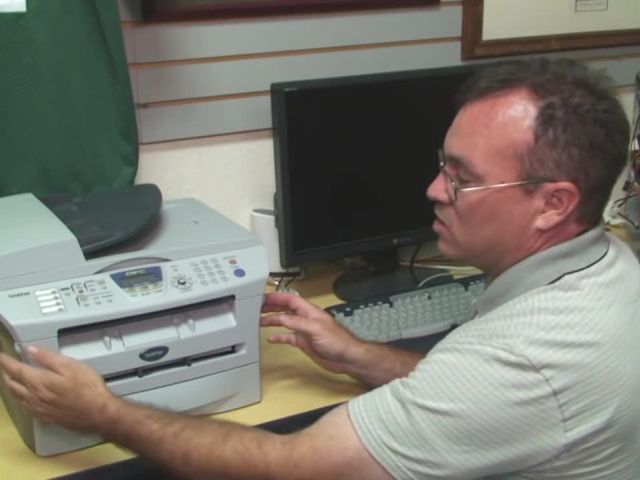}{Installing a printer}%
\tikzpicbis{0.75\textwidth}{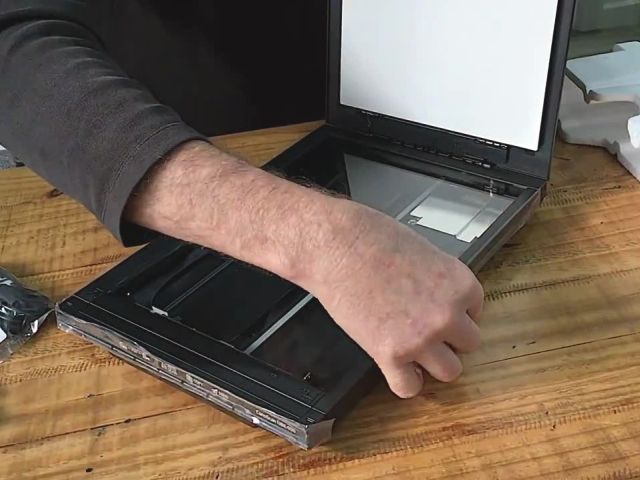}{Installing a scanner}%
\end{tikzpicture}%
\vspace{\myheightbtwtikz}
\begin{tikzpicture}
\tikzpicbis{0.00\textwidth}{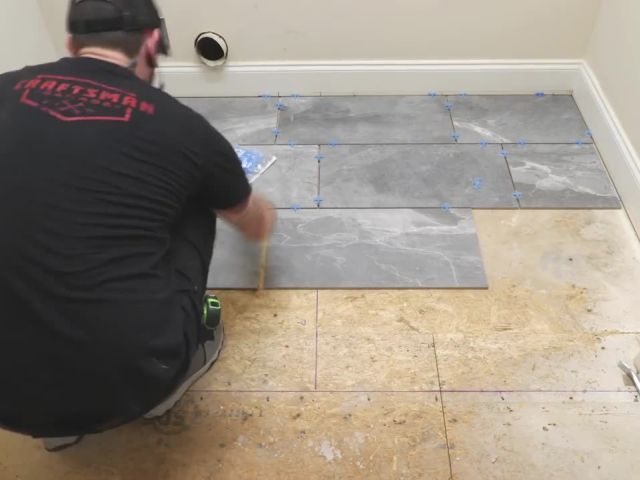}{Laying tile floors}%
\tikzpicbis{0.25\textwidth}{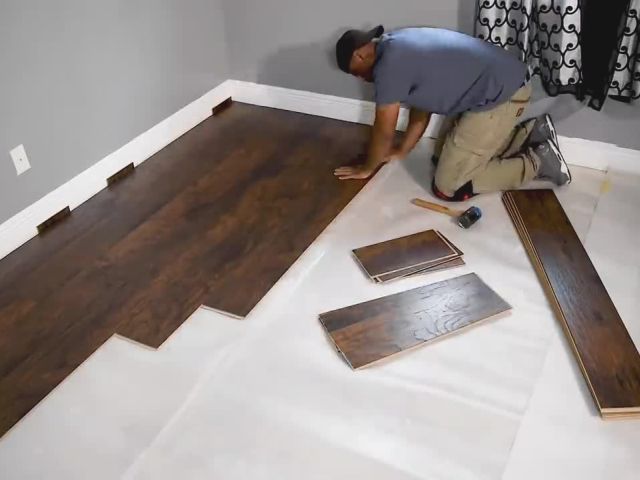}{Laying wood floors}%
\tikzpicbis{0.50\textwidth}{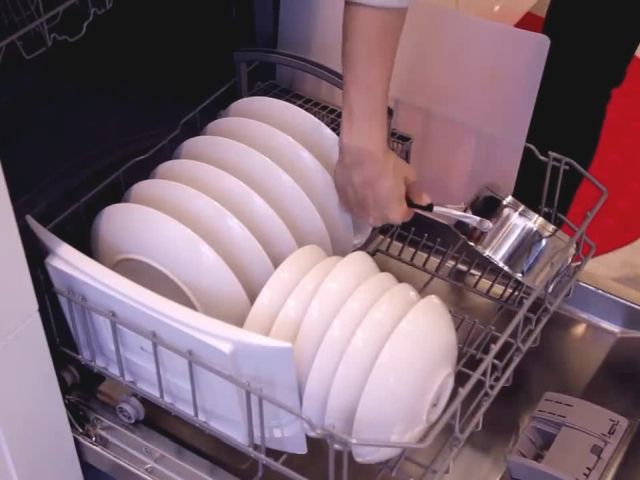}{Loading the dishwasher}%
\tikzpicbis{0.75\textwidth}{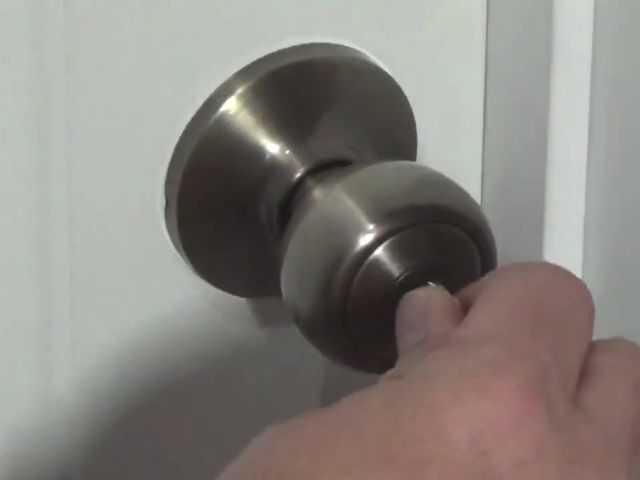}{Locking every door}%
\end{tikzpicture}%
\hfill
\caption{\textbf{\model \numActivities activities:} Each pair of images depict a frame of the execution of the activity in \model from the agent's perspective in virtual reality (\textit{left}) and the same activity in real-life from a YouTube video (\textit{right}). All activities are selected from the American Time Use Survey~\cite{atus}, and correspond to simulatable household chores relevant in human's everyday life. The set of activities cover common areas like cleaning, maintenance, preparation for social activities, or household management.
}
    \label{fig:behavior_acts_page1}
\end{figure}

\begin{figure}[H]
\ContinuedFloat
\begin{tikzpicture}
\tikzpicbis{0.00\textwidth}{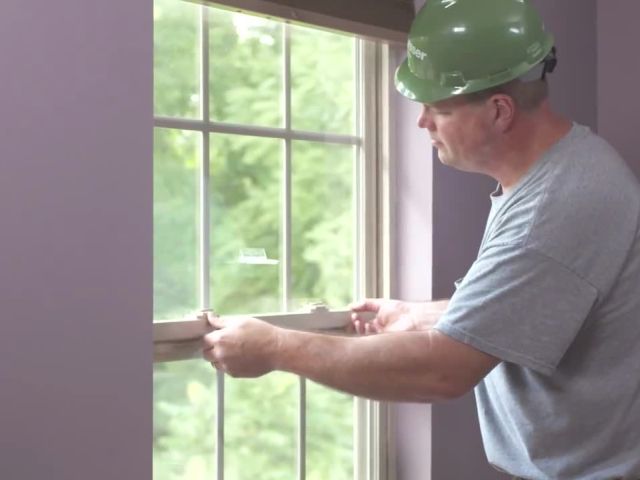}{Locking every window}%
\tikzpicbis{0.25\textwidth}{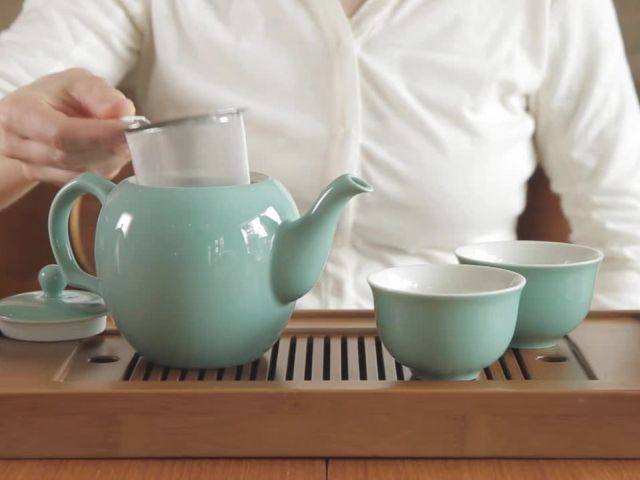}{Making tea}%
\tikzpicbis{0.50\textwidth}{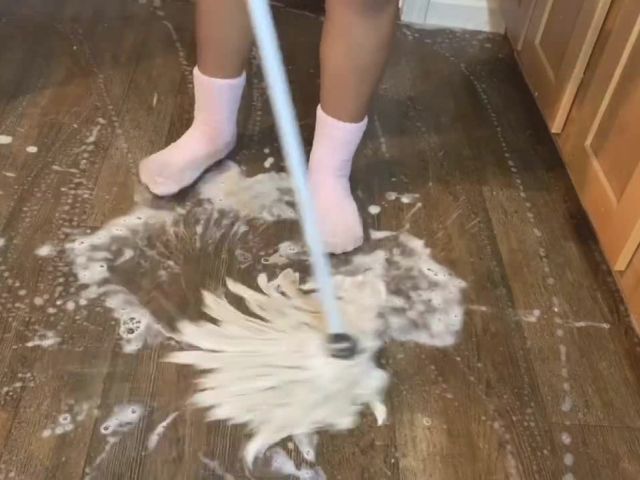}{Mopping floors}%
\tikzpicbis{0.75\textwidth}{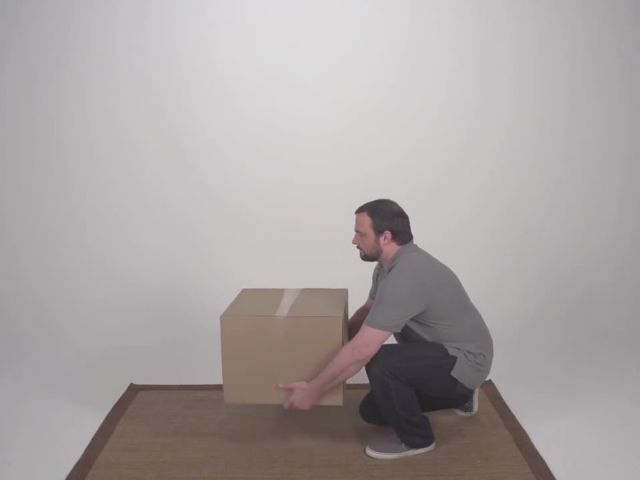}{Moving boxes to storage}%
\end{tikzpicture}%
\vspace{\myheightbtwtikz}
\begin{tikzpicture}
\tikzpicbis{0.00\textwidth}{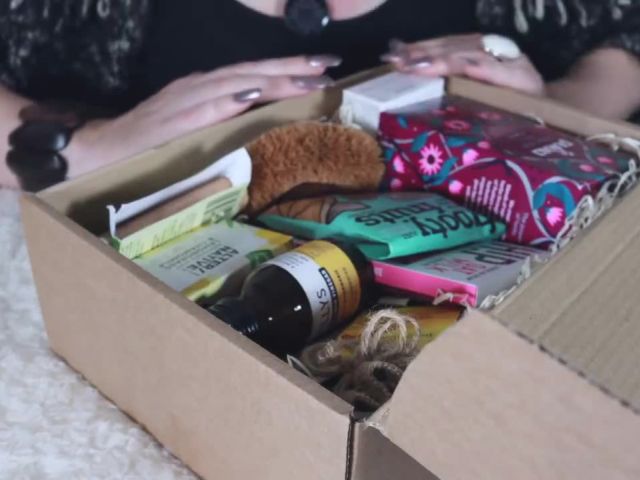}{Opening packages}%
\tikzpicbis{0.25\textwidth}{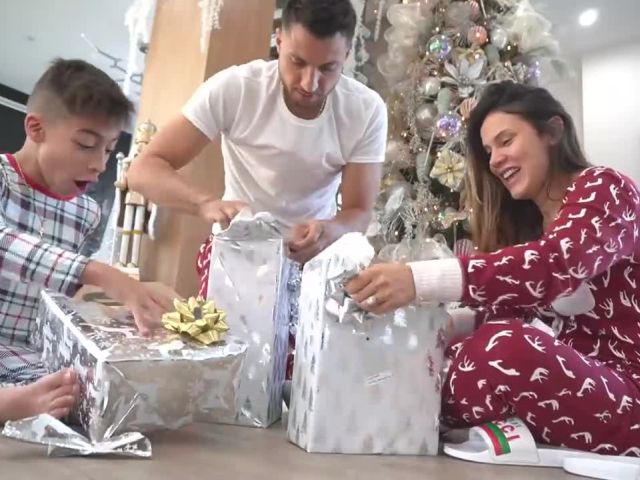}{Opening presents}%
\tikzpicbis{0.50\textwidth}{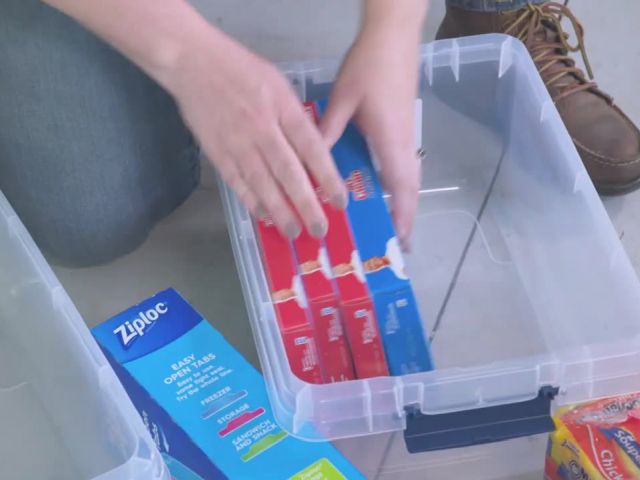}{Organizing boxes in garage}%
\tikzpicbis{0.75\textwidth}{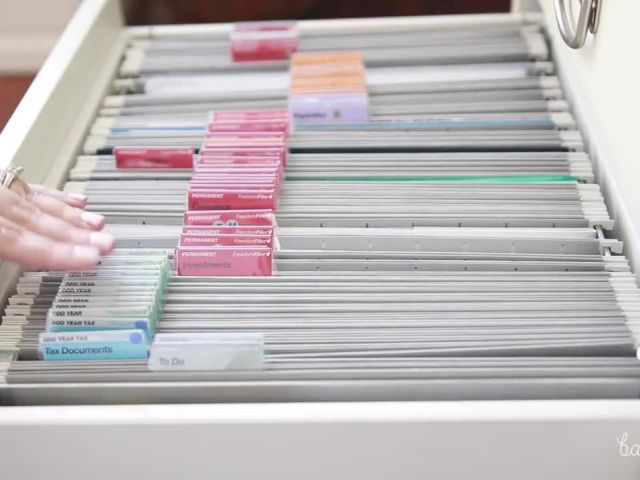}{Organizing file cabinet}%
\end{tikzpicture}%
\vspace{\myheightbtwtikz}
\begin{tikzpicture}
\tikzpicbis{0.00\textwidth}{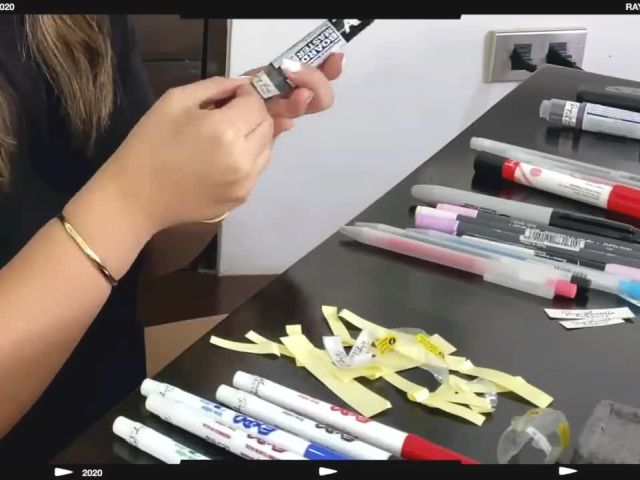}{Organizing school stuff}%
\tikzpicbis{0.25\textwidth}{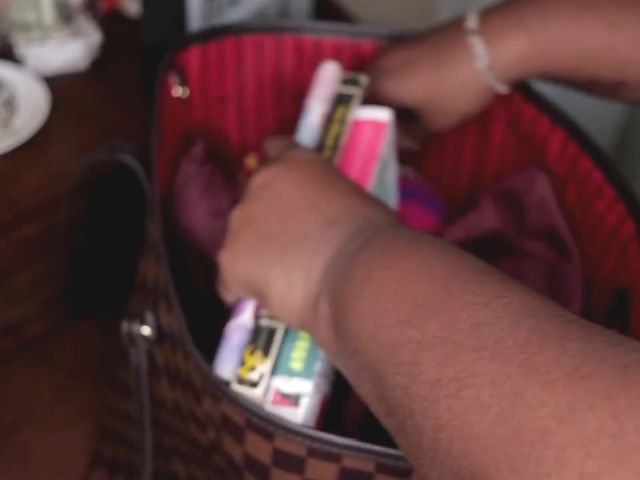}{Packing adults' bags}%
\tikzpicbis{0.50\textwidth}{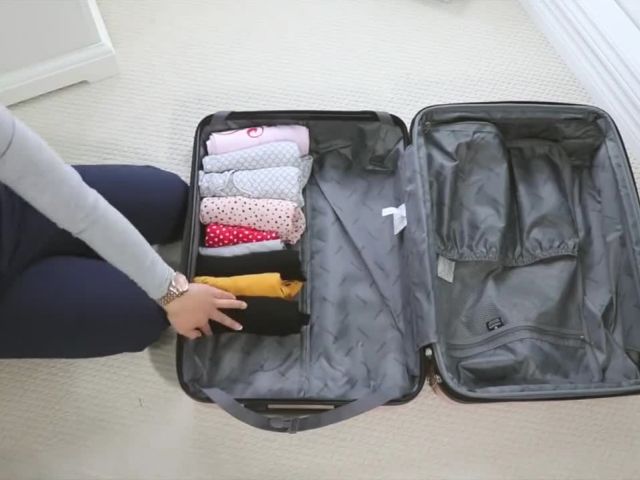}{Packing bags or suitcase}%
\tikzpicbis{0.75\textwidth}{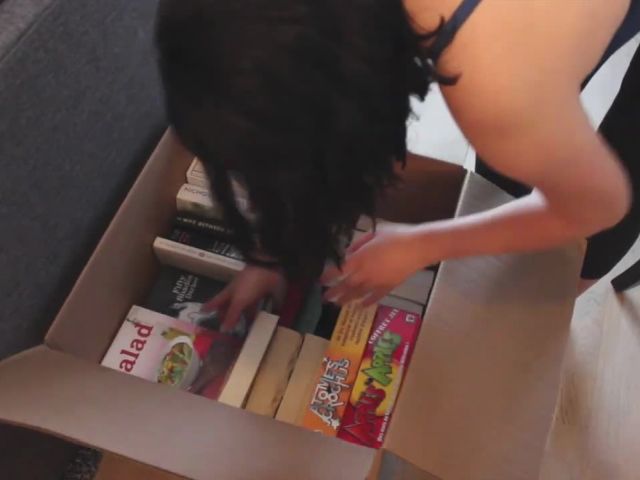}{Packing boxes for move}%
\end{tikzpicture}%
\vspace{\myheightbtwtikz}
\begin{tikzpicture}
\tikzpicbis{0.00\textwidth}{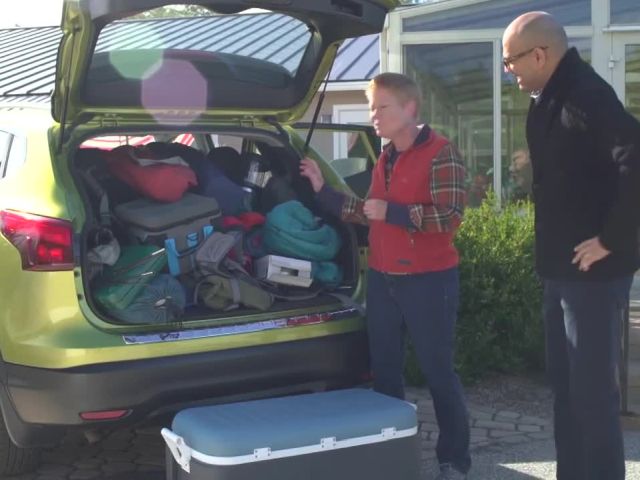}{Packing car for trip}%
\tikzpicbis{0.25\textwidth}{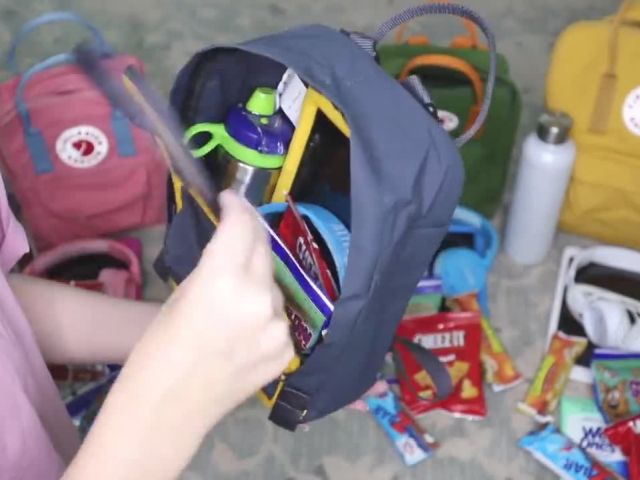}{Packing child's bag}%
\tikzpicbis{0.50\textwidth}{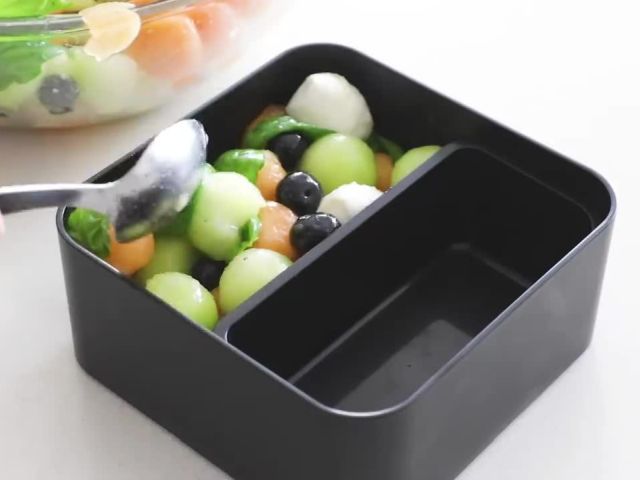}{Packing food for work}%
\tikzpicbis{0.75\textwidth}{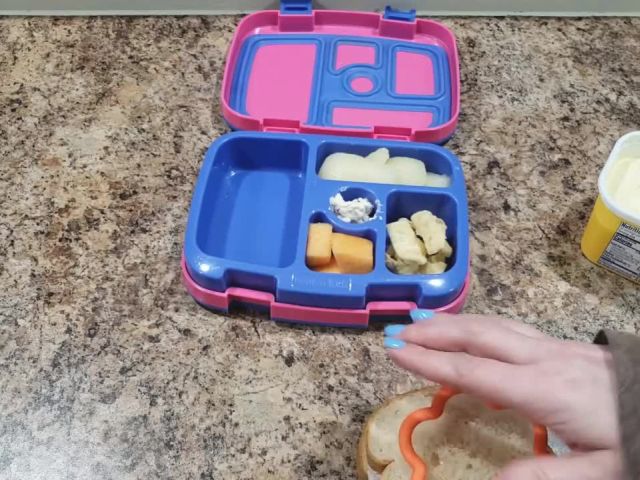}{Packing lunches}%
\end{tikzpicture}%
\vspace{\myheightbtwtikz}
\begin{tikzpicture}
\tikzpicbis{0.00\textwidth}{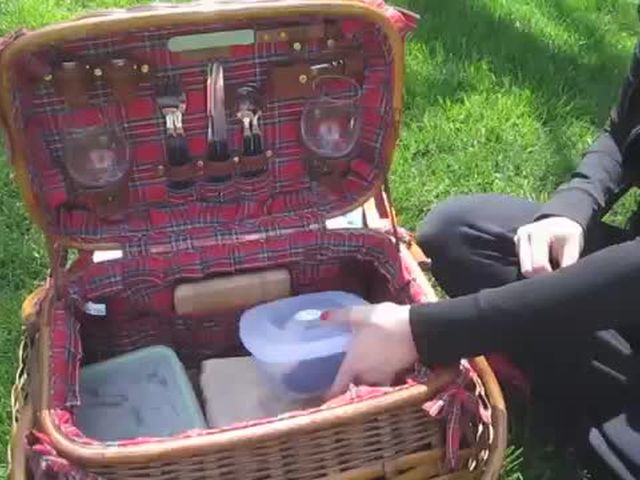}{Packing picnics}%
\tikzpicbis{0.25\textwidth}{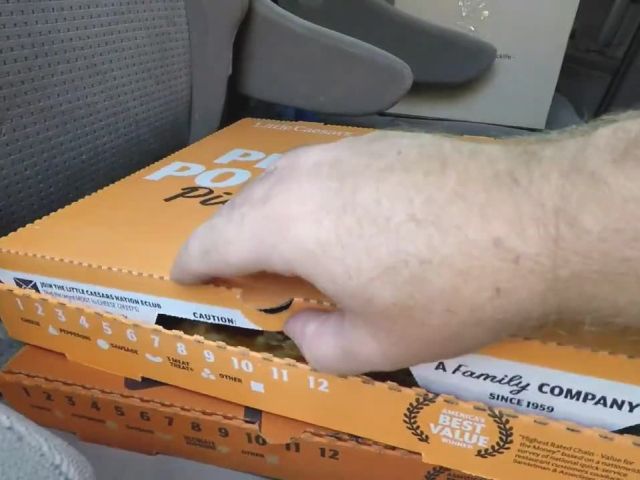}{Picking up take-out food}%
\tikzpicbis{0.50\textwidth}{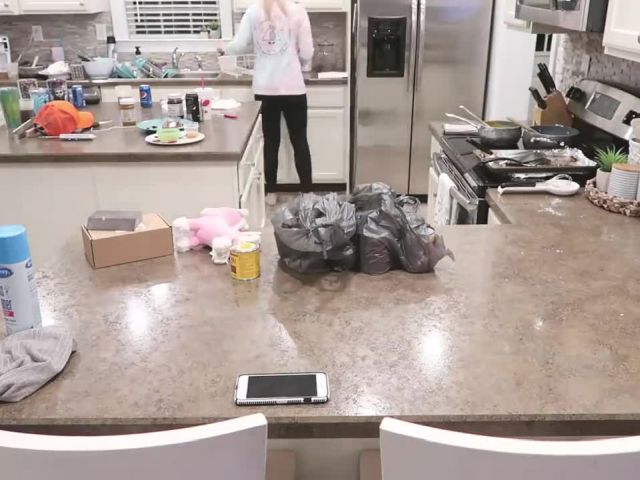}{Picking up trash}%
\tikzpicbis{0.75\textwidth}{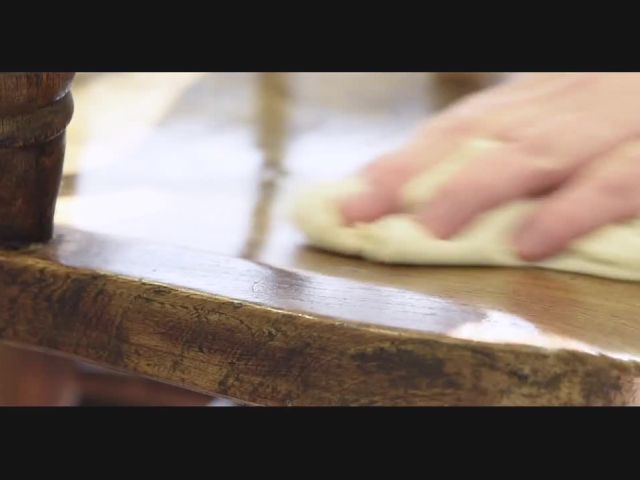}{Polishing furniture}%
\end{tikzpicture}%
\vspace{\myheightbtwtikz}
\begin{tikzpicture}
\tikzpicbis{0.00\textwidth}{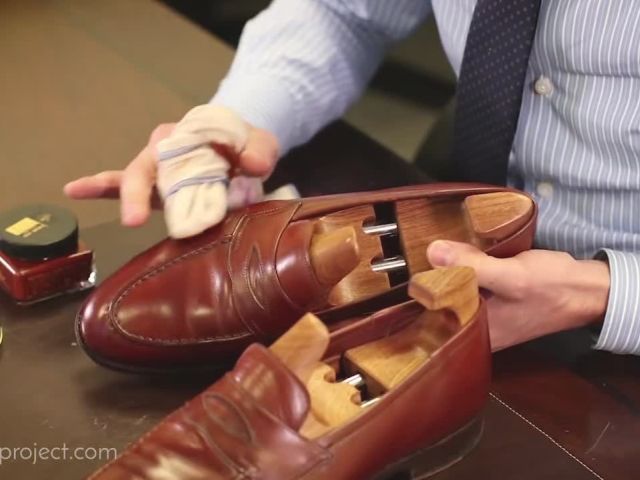}{Polishing shoes}%
\tikzpicbis{0.25\textwidth}{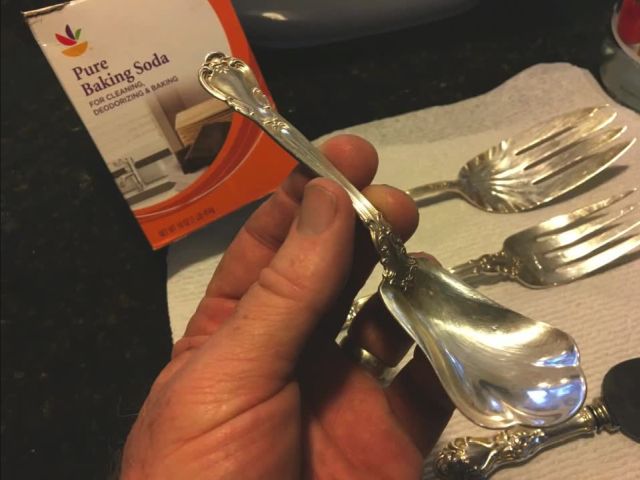}{Polishing silver}%
\tikzpicbis{0.50\textwidth}{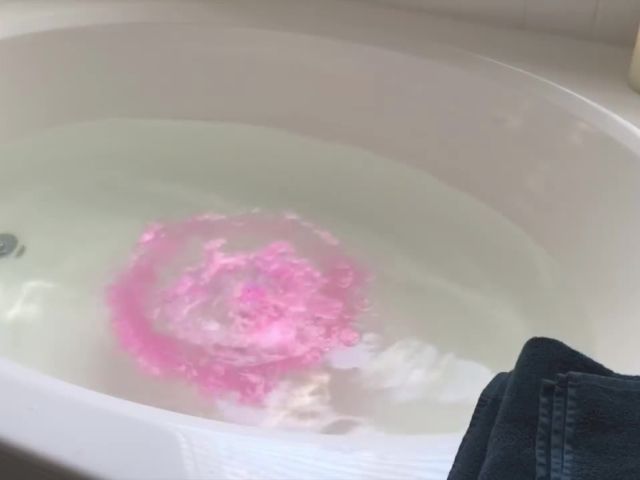}{Preparing a shower for child}%
\tikzpicbis{0.75\textwidth}{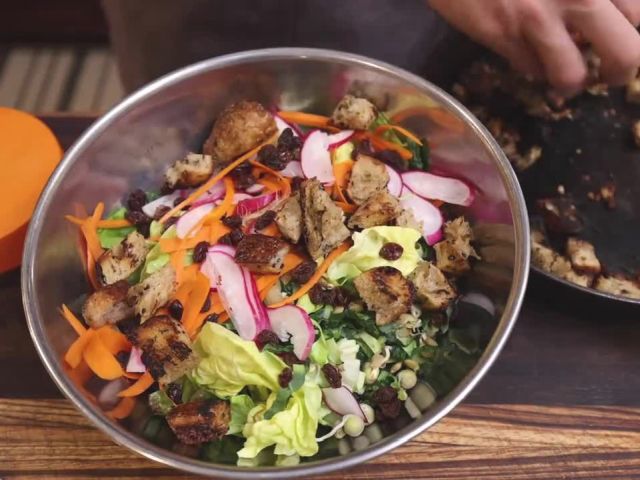}{Preparing salad}%
\end{tikzpicture}%
\vspace{\myheightbtwtikz}
\begin{tikzpicture}
\tikzpicbis{0.00\textwidth}{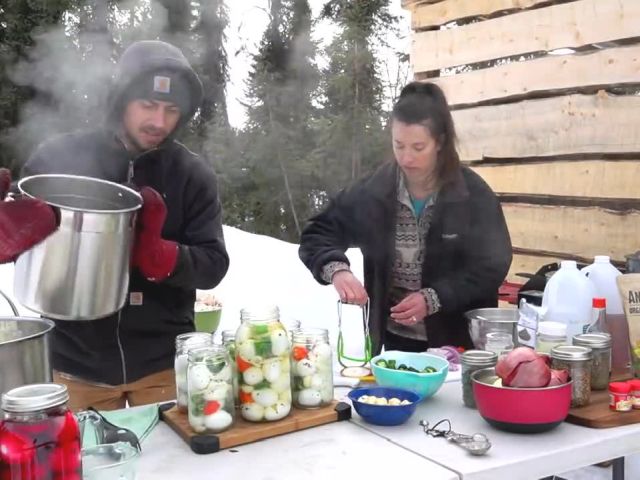}{Preserving food}%
\tikzpicbis{0.25\textwidth}{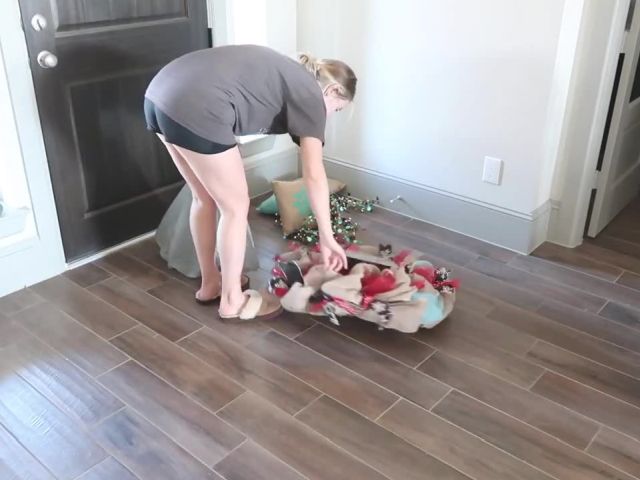}{Putting away Christmas decor}%
\tikzpicbis{0.50\textwidth}{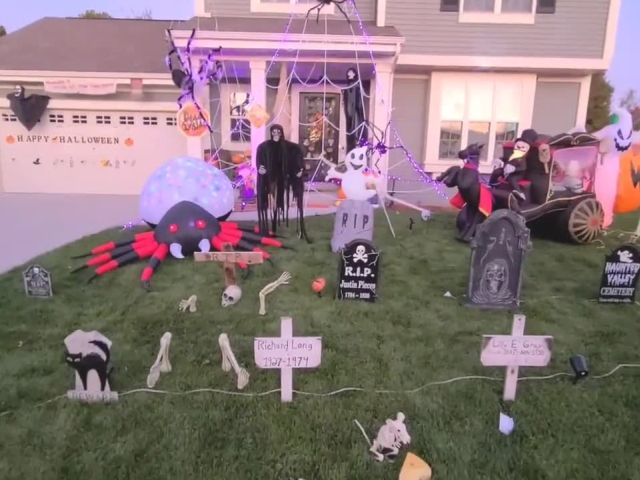}{Putting away Halloween decor}%
\tikzpicbis{0.75\textwidth}{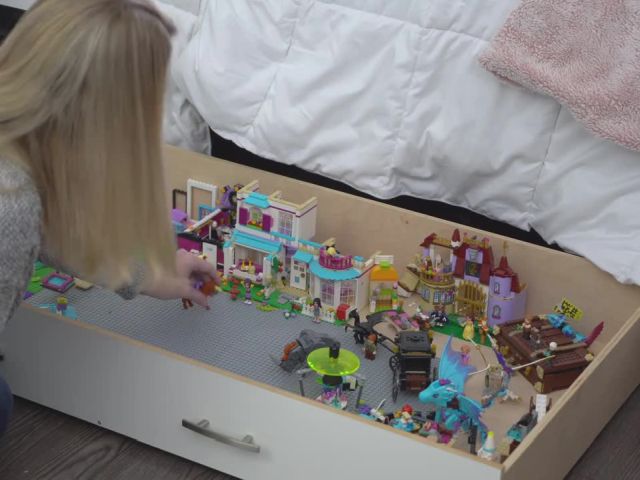}{Putting away toys}%
\end{tikzpicture}%
\vspace{\myheightbtwtikz}
\begin{tikzpicture}
\tikzpicbis{0.00\textwidth}{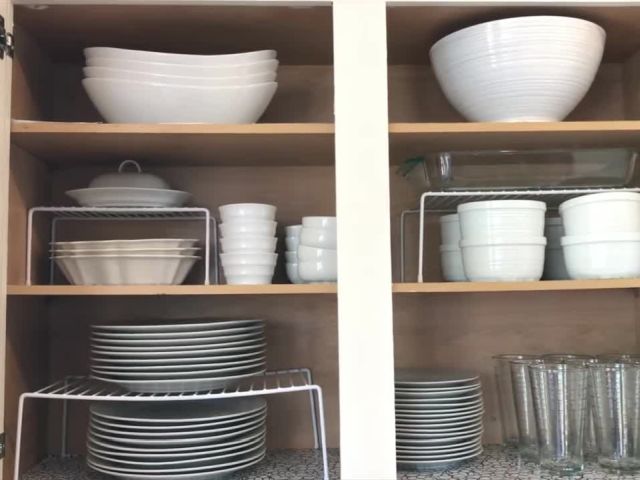}{Putting away cleaned dishes}%
\tikzpicbis{0.25\textwidth}{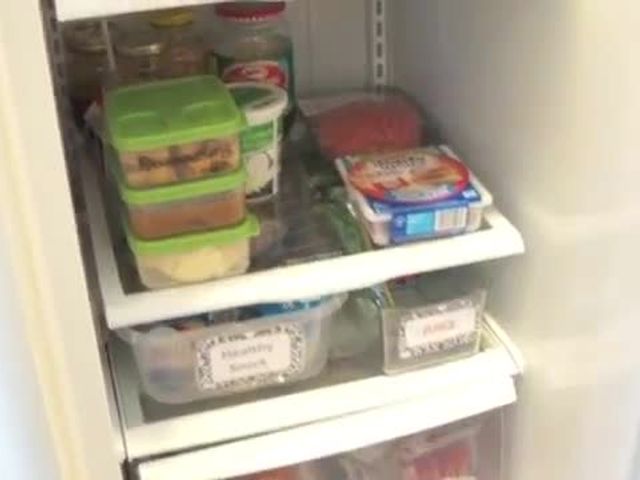}{Putting leftovers away}%
\tikzpicbis{0.50\textwidth}{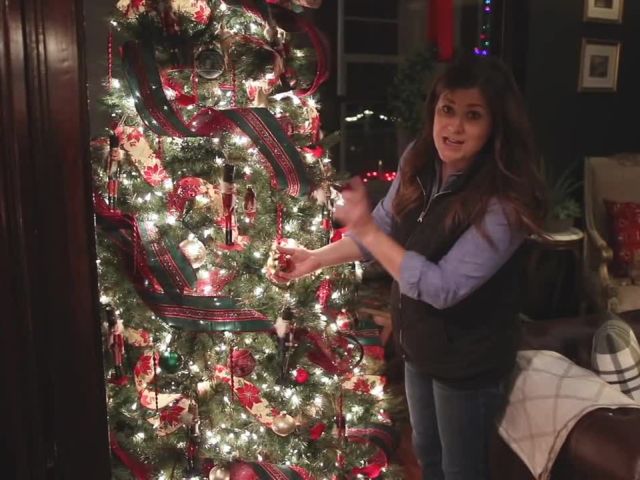}{Putting up Christmas decor}%
\tikzpicbis{0.75\textwidth}{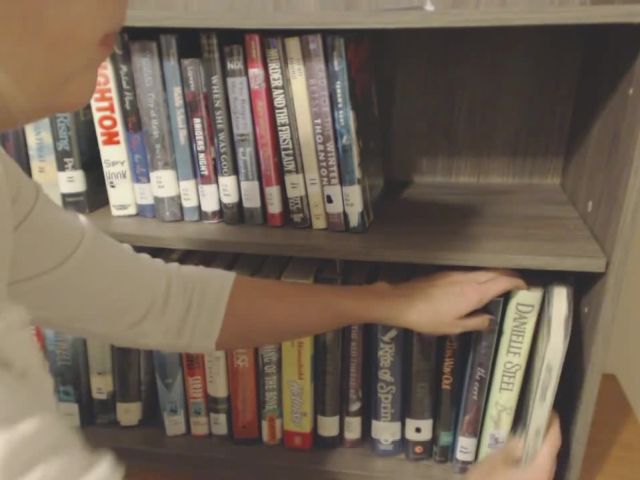}{Re-shelving library books}%
\end{tikzpicture}%
\vspace{\myheightbtwtikz}
\begin{tikzpicture}
\tikzpicbis{0.00\textwidth}{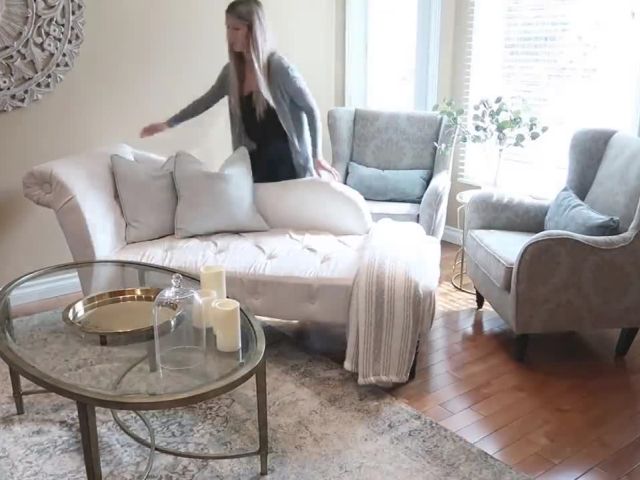}{Rearranging furniture}%
\tikzpicbis{0.25\textwidth}{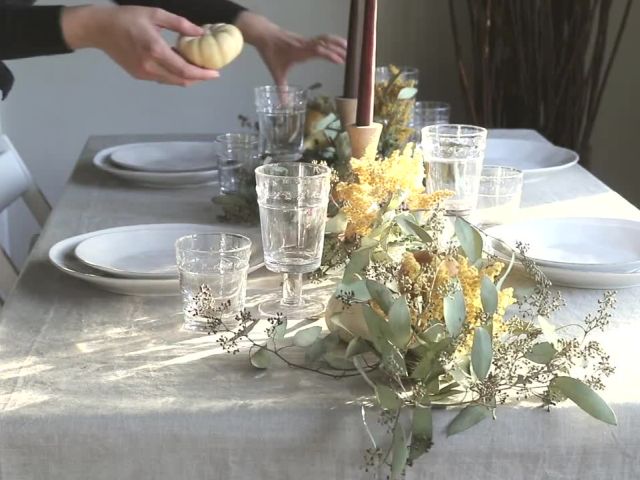}{Serving a meal}%
\tikzpicbis{0.50\textwidth}{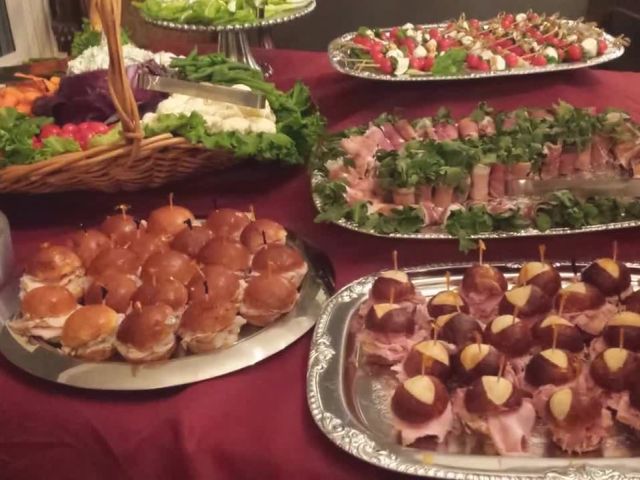}{Serving hors d oeuvres}%
\tikzpicbis{0.75\textwidth}{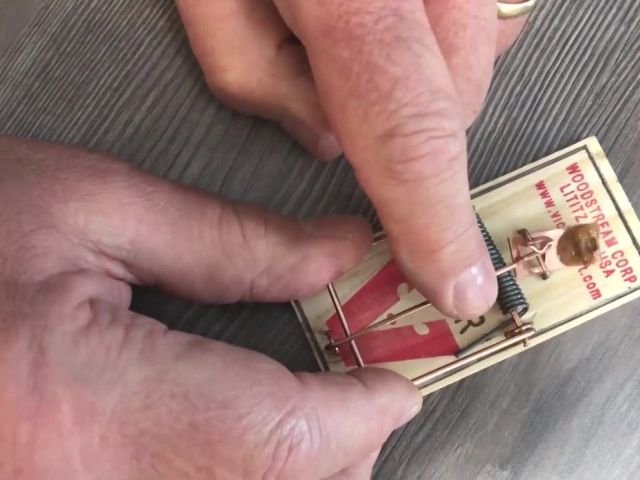}{Setting mousetraps}%
\end{tikzpicture}%
\vspace{\myheightbtwtikz}
\begin{tikzpicture}
\tikzpicbis{0.00\textwidth}{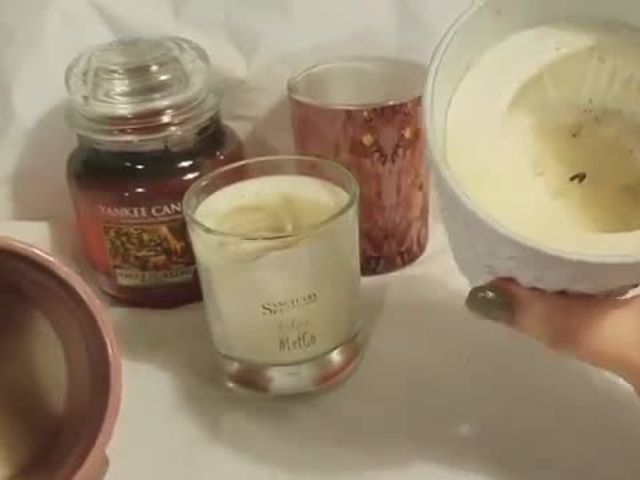}{Setting up candles}%
\tikzpicbis{0.25\textwidth}{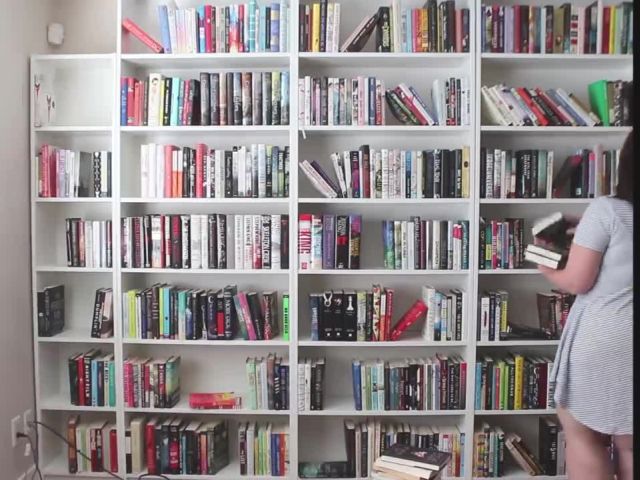}{Sorting books}%
\tikzpicbis{0.50\textwidth}{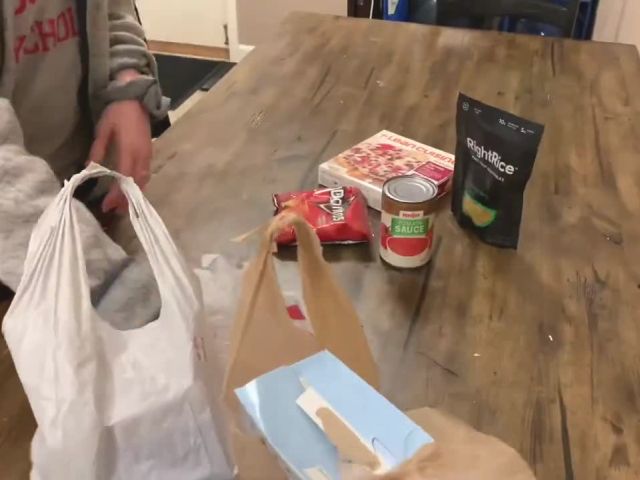}{Sorting groceries}%
\tikzpicbis{0.75\textwidth}{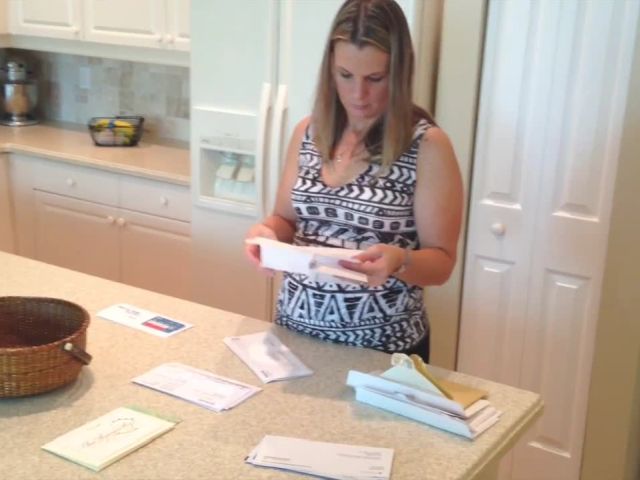}{Sorting mail}%
\end{tikzpicture}%
\vspace{\myheightbtwtikz}
\begin{tikzpicture}
\tikzpicbis{0.00\textwidth}{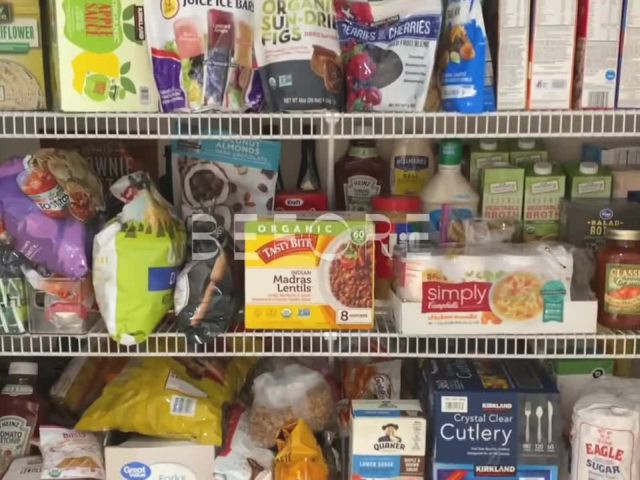}{Storing food}%
\tikzpicbis{0.25\textwidth}{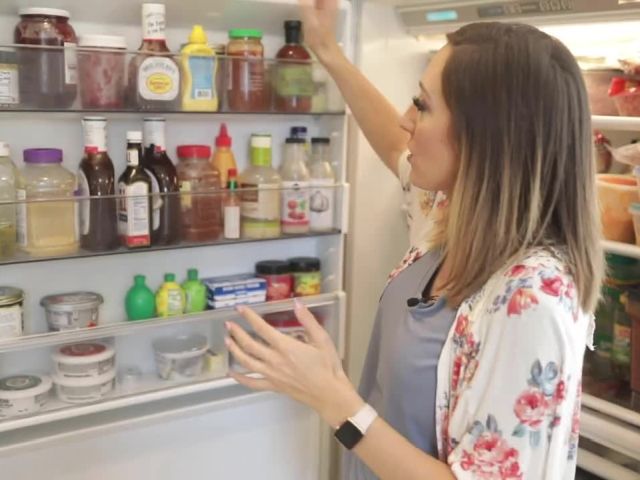}{Storing the groceries}%
\tikzpicbis{0.50\textwidth}{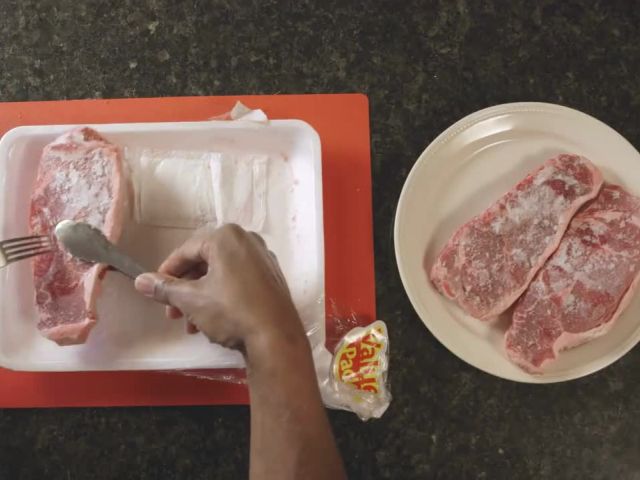}{Thawing frozen food}%
\tikzpicbis{0.75\textwidth}{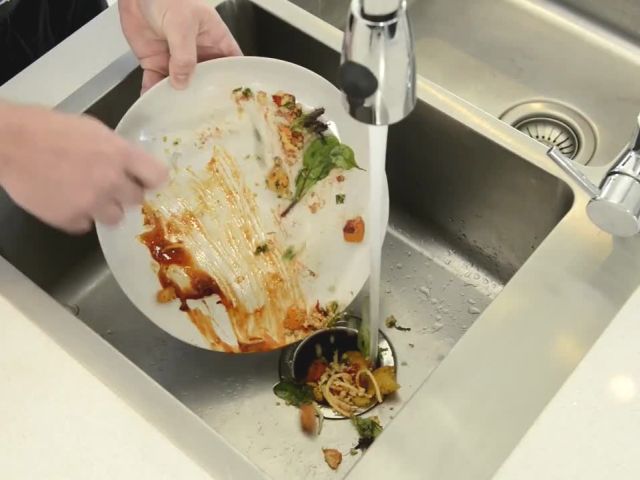}{Throwing away leftovers}%
\end{tikzpicture}%
\vspace{\myheightbtwtikz}
\begin{tikzpicture}
\tikzpicbis{0.00\textwidth}{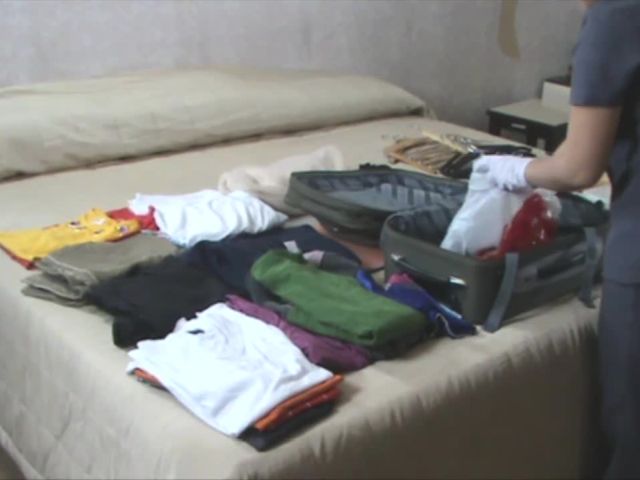}{Unpacking suitcase}%
\tikzpicbis{0.25\textwidth}{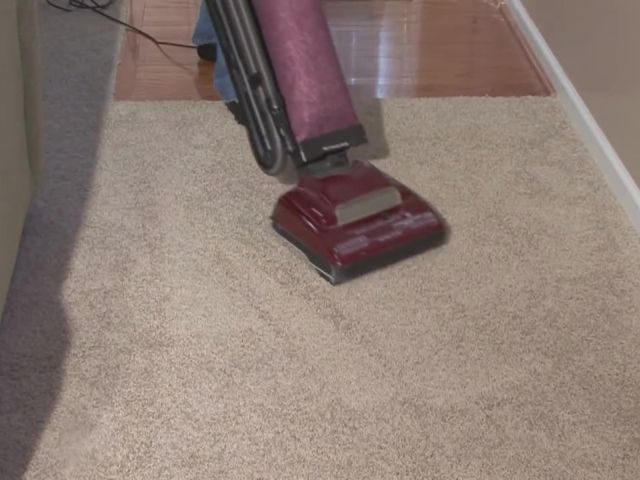}{Vacuuming floors}%
\tikzpicbis{0.50\textwidth}{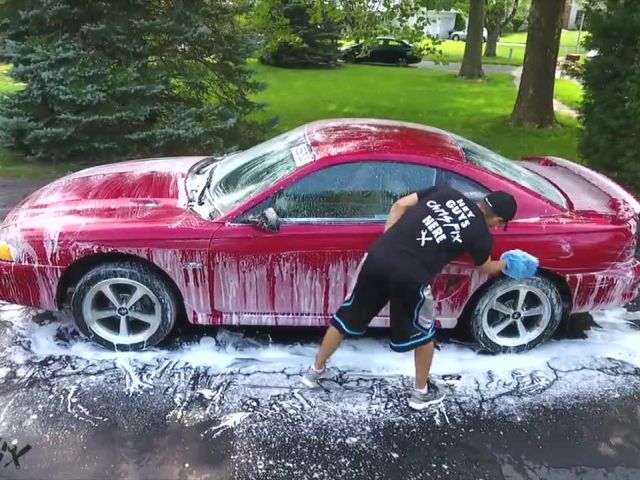}{Washing car}%
\tikzpicbis{0.75\textwidth}{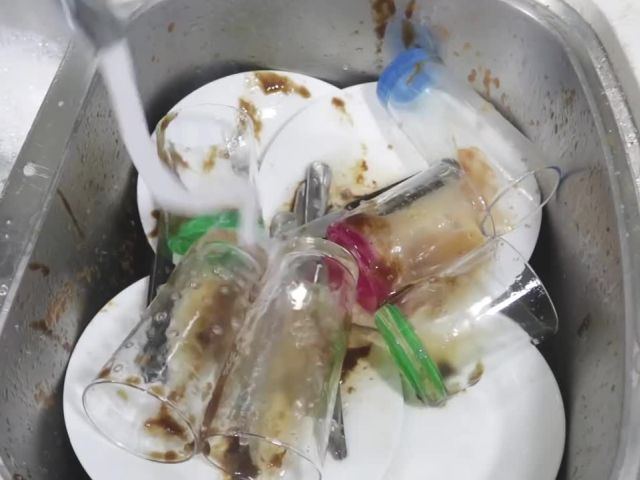}{Washing dishes}%
\end{tikzpicture}%
\vspace{\myheightbtwtikz}
\begin{tikzpicture}
\tikzpicbis{0.00\textwidth}{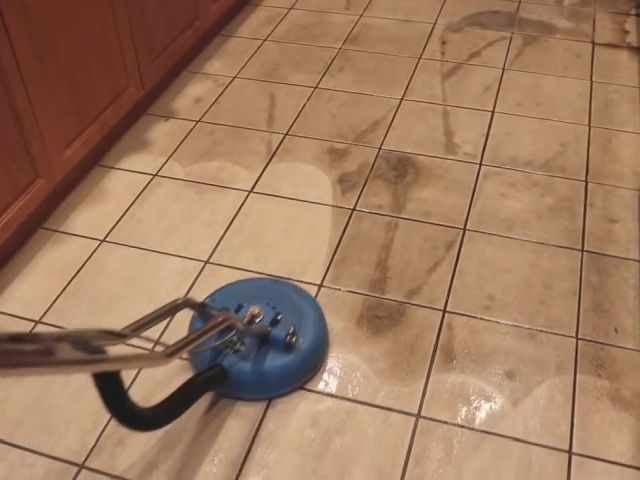}{Washing floor}%
\tikzpicbis{0.25\textwidth}{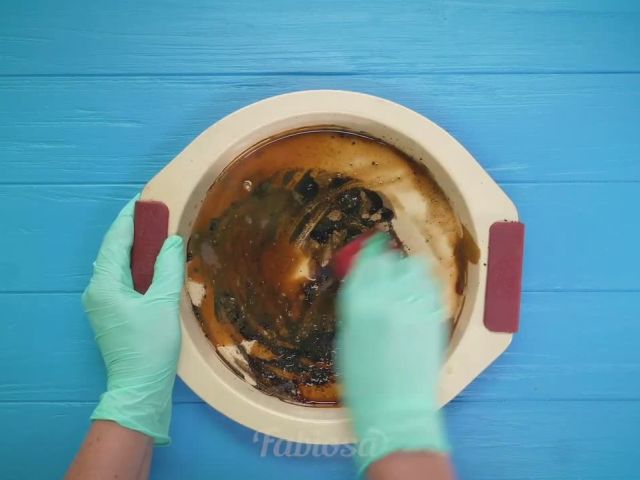}{Washing pots and pans}%
\tikzpicbis{0.50\textwidth}{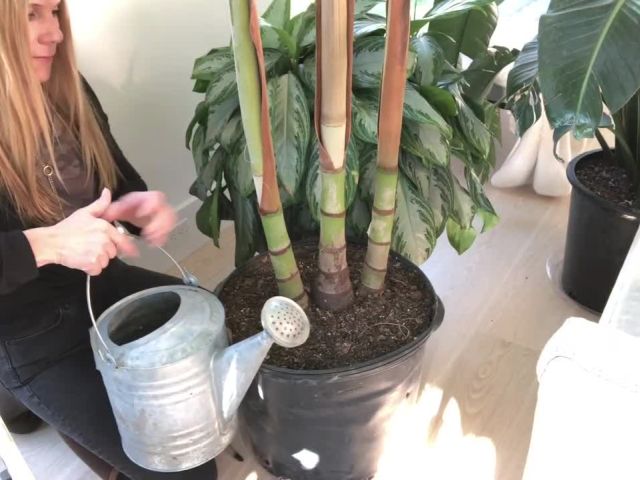}{Watering houseplants}%
\tikzpicbis{0.75\textwidth}{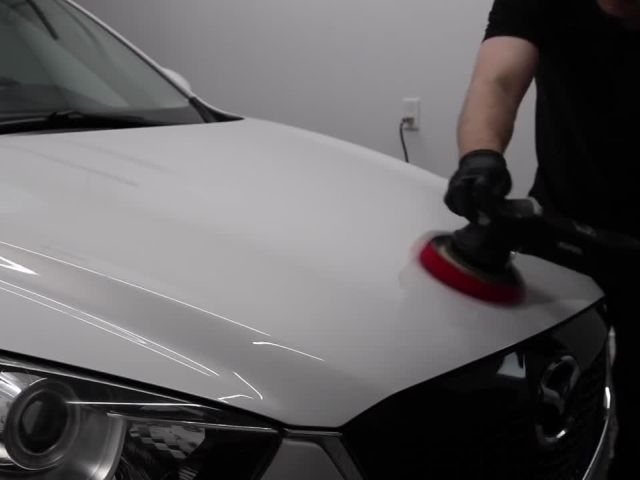}{Waxing car}%
\end{tikzpicture}%
\hfill
\caption{\textbf{\model \numActivities activities} (cont.)}
    \label{fig:behavior_acts}
\end{figure}

\subsection{Additional Comparison between \model and other Embodied AI Benchmarks}
\FloatBarrier

\begin{table}[H]
    \definecolor{myGrayCell}{rgb}{.8,.8,.8}
    \centering
    \resizebox{\linewidth}{!}{
    \begin{tabular}{|c|cc|c|ccccccccccccccccc|}
        \rot{} & 
        \rot{} & 
        \rot{} & 
        \rot{\textbf{BEHAVIOR}} & 
        \rot{AI2THOR Visual Room Rearrangement} & 
        \rot{TDW Transport Challenge} & 
        \rot{Rearrangement T5 (Habitat)} &
        \rot{ManipulaTHOR ArmPointNav} & 
        \rot{Interactive Gibson Benchmark} & 
        \rot{VirtualHome} & 
        \rot{ALFRED} & 
        \rot{OCRTOC} & 
        \rot{RLBench} & 
        \rot{Metaworld} & 
        \rot{IKEA Furniture Assembly} & 
        \rot{Robosuite} & 
        \rot{SoftGym} & 
        \rot{DeepMind Control Suite} & 
        \rot{OpenAIGym} & 
        \rot{Habitat 1.0} & 
        \rot{Gibson}
        \\\cline{1-21}
        & 
        \begin{tabular}{@{}c@{}}Realistic\\activities\end{tabular} & \cellcolor[rgb]{.8,.8,.8}\begin{tabular}{@{}c@{}}Activities  match\\human time-use\end{tabular} & 
        \cellcolor[rgb]{.8,.8,.8}\textcolor{indiagreen}\faCheck & 
        \cellcolor[rgb]{.8,.8,.8}\textcolor{red}\faTimes & 
        \cellcolor[rgb]{.8,.8,.8}\textcolor{red}\faTimes & 
        \cellcolor[rgb]{.8,.8,.8}\textcolor{red}\faTimes & 
        \cellcolor[rgb]{.8,.8,.8}\textcolor{red}\faTimes & 
        \cellcolor[rgb]{.8,.8,.8}\textcolor{red}\faTimes & 
        \cellcolor[rgb]{.8,.8,.8}\textcolor{red}\faTimes & 
        \cellcolor[rgb]{.8,.8,.8}\textcolor{red}\faTimes & 
        \cellcolor[rgb]{.8,.8,.8}\textcolor{red}\faTimes & 
        \cellcolor[rgb]{.8,.8,.8}\textcolor{red}\faTimes & 
        \cellcolor[rgb]{.8,.8,.8}\textcolor{red}\faTimes & 
        \cellcolor[rgb]{.8,.8,.8}\textcolor{red}\faTimes & 
        \cellcolor[rgb]{.8,.8,.8}\textcolor{red}\faTimes & 
        \cellcolor[rgb]{.8,.8,.8}\textcolor{red}\faTimes & 
        \cellcolor[rgb]{.8,.8,.8}\textcolor{red}\faTimes & 
        \cellcolor[rgb]{.8,.8,.8}\textcolor{red}\faTimes & 
        \cellcolor[rgb]{.8,.8,.8}\textcolor{red}\faTimes & 
        \cellcolor[rgb]{.8,.8,.8}\textcolor{red}\faTimes 
        \\\cline{2-3}
        & 
        \multirow{3}*{\begin{tabular}{@{}c@{}}Realistic\\physics\end{tabular}} & 
        \begin{tabular}{@{}c@{}}Kinematics,\\dynamics\end{tabular} & 
        \textcolor{indiagreen}\faCheck &
        \textcolor{indiagreen}\faCheck &
        \textcolor{indiagreen}\faCheck &
         \textcolor{indiagreen}\faCheck &
         \textcolor{indiagreen}\faCheck &
         \textcolor{indiagreen}\faCheck &
         \textcolor{red}\faTimes &
         \textcolor{indiagreen}\faCheck &
         \textcolor{indiagreen}\faCheck &
         \textcolor{indiagreen}\faCheck &
         \textcolor{indiagreen}\faCheck &
         \textcolor{indiagreen}\faCheck &
         \textcolor{indiagreen}\faCheck &
         \textcolor{indiagreen}\faCheck &
         \textcolor{indiagreen}\faCheck &
         \textcolor{indiagreen}\faCheck &
         \textcolor{indiagreen}\faCheck &
         \textcolor{indiagreen}\faCheck 
        \\ 
        & 
        & 
        \cellcolor[rgb]{.8,.8,.8} Continuous temperature & 
        \cellcolor[rgb]{.8,.8,.8} \textcolor{indiagreen}\faCheck & 
        \cellcolor[rgb]{.8,.8,.8} \textcolor{red}\faTimes & 
        \cellcolor[rgb]{.8,.8,.8} \textcolor{red}\faTimes & 
        \cellcolor[rgb]{.8,.8,.8} \textcolor{red}\faTimes & 
        \cellcolor[rgb]{.8,.8,.8} \textcolor{red}\faTimes & 
        \cellcolor[rgb]{.8,.8,.8} \textcolor{red}\faTimes & 
        \cellcolor[rgb]{.8,.8,.8} \textcolor{red}\faTimes & 
        \cellcolor[rgb]{.8,.8,.8} \textcolor{red}\faTimes & 
        \cellcolor[rgb]{.8,.8,.8} \textcolor{red}\faTimes & 
        \cellcolor[rgb]{.8,.8,.8} \textcolor{red}\faTimes & 
        \cellcolor[rgb]{.8,.8,.8} \textcolor{red}\faTimes & 
        \cellcolor[rgb]{.8,.8,.8} \textcolor{red}\faTimes & 
        \cellcolor[rgb]{.8,.8,.8} \textcolor{red}\faTimes & 
        \cellcolor[rgb]{.8,.8,.8} \textcolor{red}\faTimes & 
       \cellcolor[rgb]{.8,.8,.8}  \textcolor{red}\faTimes & 
       \cellcolor[rgb]{.8,.8,.8}  \textcolor{red}\faTimes & 
       \cellcolor[rgb]{.8,.8,.8}  \textcolor{red}\faTimes & 
       \cellcolor[rgb]{.8,.8,.8}  \textcolor{red}\faTimes 
        \\ 
        & 
        & 
        Flexible materials & 
        \textcolor{red}\faTimes & 
        \textcolor{red}\faTimes & 
        \textcolor{red}\faTimes & 
        \textcolor{red}\faTimes & 
        \textcolor{red}\faTimes & 
        \textcolor{red}\faTimes & 
        \textcolor{red}\faTimes & 
        \textcolor{red}\faTimes & 
        \textcolor{red}\faTimes & 
        \textcolor{red}\faTimes & 
        \textcolor{red}\faTimes & 
        \textcolor{red}\faTimes & 
        \textcolor{red}\faTimes & 
        \textcolor{indiagreen}\faCheck & 
        \textcolor{red}\faTimes & 
        \textcolor{red}\faTimes & 
        \textcolor{red}\faTimes & 
        \textcolor{red}\faTimes 
        \\ \cline{2-3}
        
        \multirow{10}*{\rotatebox{90}{Realism}} &
        \multirow{2}*{\begin{tabular}{@{}c@{}}Realistic\\embodied AI\\agents\end{tabular}} & 
        \cellcolor[rgb]{.8,.8,.8}\begin{tabular}{@{}c@{}}Realistic action\\execution\end{tabular} & 
        \cellcolor[rgb]{.8,.8,.8}  \textcolor{indiagreen}\faCheck &
        \cellcolor[rgb]{.8,.8,.8}  \textcolor{red}\faTimes &
        \cellcolor[rgb]{.8,.8,.8}  \textcolor{indiagreen}\faCheck &
        \cellcolor[rgb]{.8,.8,.8}  \textcolor{red}\faTimes &
       \cellcolor[rgb]{.8,.8,.8}   \textcolor{indiagreen}\faCheck &
        \cellcolor[rgb]{.8,.8,.8}  \textcolor{indiagreen}\faCheck &
        \cellcolor[rgb]{.8,.8,.8}  \textcolor{red}\faTimes &
        \cellcolor[rgb]{.8,.8,.8}  \textcolor{red}\faTimes &
        \cellcolor[rgb]{.8,.8,.8}  \textcolor{red}\faTimes &
        \cellcolor[rgb]{.8,.8,.8}  \textcolor{indiagreen}\faCheck &
        \cellcolor[rgb]{.8,.8,.8}  \textcolor{indiagreen}\faCheck &
        \cellcolor[rgb]{.8,.8,.8}  \textcolor{indiagreen}\faCheck &
        \cellcolor[rgb]{.8,.8,.8}  \textcolor{indiagreen}\faCheck &
        \cellcolor[rgb]{.8,.8,.8}  \textcolor{indiagreen}\faCheck &
        \cellcolor[rgb]{.8,.8,.8}  \textcolor{indiagreen}\faCheck &
        \cellcolor[rgb]{.8,.8,.8}  \textcolor{indiagreen}\faCheck &
        \cellcolor[rgb]{.8,.8,.8}  \textcolor{indiagreen}\faCheck &
       \cellcolor[rgb]{.8,.8,.8}   \textcolor{indiagreen}\faCheck
        \\
        & 
        & 
        \begin{tabular}{@{}c@{}}Realistic\\observations\end{tabular} &
        \textcolor{indiagreen}\faCheck & 
        \textcolor{indiagreen}\faCheck & 
        \textcolor{indiagreen}\faCheck &
        \textcolor{indiagreen}\faCheck & 
        \textcolor{red}\faTimes &
        \textcolor{indiagreen}\faCheck & 
        \textcolor{red}\faTimes &
        \textcolor{red}\faTimes &
        \textcolor{indiagreen}\faCheck &
        \textcolor{indiagreen}\faCheck & 
        \textcolor{indiagreen}\faCheck &
        \textcolor{indiagreen}\faCheck & 
        \textcolor{indiagreen}\faCheck &
        \textcolor{indiagreen}\faCheck & 
        \textcolor{indiagreen}\faCheck &
        \textcolor{indiagreen}\faCheck &
        \textcolor{indiagreen}\faCheck &
        \textcolor{indiagreen}\faCheck 
        \\\cline{2-3}
        & 
        \multirow{2}*{\begin{tabular}{@{}c@{}}Realistic\\scenes\end{tabular}} & 
        \cellcolor[rgb]{.8,.8,.8}\begin{tabular}{@{}c@{}}Visually\\realistic\end{tabular} & 
        \cellcolor[rgb]{.8,.8,.8}  \textcolor{indiagreen}\faCheck &
       \cellcolor[rgb]{.8,.8,.8}   \textcolor{indiagreen}\faCheck &
       \cellcolor[rgb]{.8,.8,.8}   \textcolor{indiagreen}\faCheck &
       \cellcolor[rgb]{.8,.8,.8}   \textcolor{indiagreen}\faCheck &
       \cellcolor[rgb]{.8,.8,.8}   \textcolor{indiagreen}\faCheck &
       \cellcolor[rgb]{.8,.8,.8}   \textcolor{indiagreen}\faCheck &
       \cellcolor[rgb]{.8,.8,.8}   \textcolor{indiagreen}\faCheck &
      \cellcolor[rgb]{.8,.8,.8}    \textcolor{indiagreen}\faCheck &
      \cellcolor[rgb]{.8,.8,.8}    \textcolor{indiagreen}\faCheck &
      \cellcolor[rgb]{.8,.8,.8}    \textcolor{red}\faTimes & 
      \cellcolor[rgb]{.8,.8,.8}    \textcolor{red}\faTimes &
        \cellcolor[rgb]{.8,.8,.8}  \textcolor{indiagreen}\faCheck &
       \cellcolor[rgb]{.8,.8,.8}   \textcolor{red}\faTimes &
       \cellcolor[rgb]{.8,.8,.8}   \textcolor{red}\faTimes &
       \cellcolor[rgb]{.8,.8,.8}   \textcolor{red}\faTimes &
       \cellcolor[rgb]{.8,.8,.8}   \textcolor{red}\faTimes &
        \cellcolor[rgb]{.8,.8,.8}  \textcolor{indiagreen}\faCheck &
        \cellcolor[rgb]{.8,.8,.8}  \textcolor{indiagreen}\faCheck 
        \\
        & 
        & 
        \begin{tabular}{@{}c@{}}Scenes reconstructed\\from real homes\end{tabular} & 
        \textcolor{indiagreen}\faCheck &
        \textcolor{red}\faTimes &
        \textcolor{red}\faTimes &
        \textcolor{indiagreen}\faCheck &
        \textcolor{red}\faTimes &
        \textcolor{indiagreen}\faCheck &
        \textcolor{red}\faTimes &
        \textcolor{red}\faTimes &
        \textcolor{red}\faTimes &
        \textcolor{red}\faTimes &
        \textcolor{red}\faTimes &
        \textcolor{red}\faTimes &
        \textcolor{red}\faTimes &
        \textcolor{red}\faTimes &
        \textcolor{red}\faTimes &
        \textcolor{red}\faTimes &
        \textcolor{indiagreen}\faCheck &
        \textcolor{indiagreen}\faCheck 
        \\\cline{2-3}
        & 
        \multirow{2}*{\begin{tabular}{@{}c@{}}Realistic\\object models\end{tabular}} &
        \cellcolor[rgb]{.8,.8,.8} \begin{tabular}{@{}c@{}}Visually\\realistic\end{tabular} &
        \cellcolor[rgb]{.8,.8,.8} \textcolor{indiagreen}\faCheck & 
        \cellcolor[rgb]{.8,.8,.8} \textcolor{indiagreen}\faCheck & 
        \cellcolor[rgb]{.8,.8,.8} \textcolor{indiagreen}\faCheck & 
        \cellcolor[rgb]{.8,.8,.8} \textcolor{indiagreen}\faCheck & 
        \cellcolor[rgb]{.8,.8,.8} \textcolor{indiagreen}\faCheck & 
        \cellcolor[rgb]{.8,.8,.8} \textcolor{indiagreen}\faCheck & 
        \cellcolor[rgb]{.8,.8,.8} \textcolor{indiagreen}\faCheck & 
        \cellcolor[rgb]{.8,.8,.8} \textcolor{indiagreen}\faCheck & 
        \cellcolor[rgb]{.8,.8,.8} \textcolor{indiagreen}\faCheck & 
        \cellcolor[rgb]{.8,.8,.8} \textcolor{indiagreen}\faCheck & 
        \cellcolor[rgb]{.8,.8,.8} \textcolor{red}\faTimes & 
        \cellcolor[rgb]{.8,.8,.8} \textcolor{indiagreen}\faCheck & 
        \cellcolor[rgb]{.8,.8,.8} \textcolor{indiagreen}\faCheck & 
        \cellcolor[rgb]{.8,.8,.8} \textcolor{red}\faTimes & 
        \cellcolor[rgb]{.8,.8,.8} \textcolor{red}\faTimes & 
       \cellcolor[rgb]{.8,.8,.8}  \textcolor{red}\faTimes & 
       \cellcolor[rgb]{.8,.8,.8}  \textcolor{indiagreen}\faCheck & 
       \cellcolor[rgb]{.8,.8,.8}  N/A
        \\
        & 
        & 
         \begin{tabular}{@{}c@{}}Weight, CoM,\\texture, cook temp\end{tabular} & 
         \textcolor{indiagreen}\faCheck &
         \textcolor{red}\faTimes &
         \textcolor{red}\faTimes &
         \textcolor{red}\faTimes &
         \textcolor{red}\faTimes &
         \textcolor{red}\faTimes &
         \textcolor{red}\faTimes &
         \textcolor{red}\faTimes &
         \textcolor{red}\faTimes &
         \textcolor{red}\faTimes &
         \textcolor{red}\faTimes &
         \textcolor{red}\faTimes &
         \textcolor{red}\faTimes &
         \textcolor{red}\faTimes &
         \textcolor{red}\faTimes &
         \textcolor{red}\faTimes &
         \textcolor{red}\faTimes &
         N/A
        \\\cline{1-3}
        & 
        \multirow{2}*{\begin{tabular}{@{}c@{}}Diverse\\activities\end{tabular}} & 
        \cellcolor[rgb]{.8,.8,.8}  \begin{tabular}{@{}c@{}}\# Activities\end{tabular} & 
        \cellcolor[rgb]{.8,.8,.8}  100 & 
        \cellcolor[rgb]{.8,.8,.8}  1 & 
        \cellcolor[rgb]{.8,.8,.8}  1 & 
        \cellcolor[rgb]{.8,.8,.8}  1 &
       \cellcolor[rgb]{.8,.8,.8}   1 & 
       \cellcolor[rgb]{.8,.8,.8}   2 & 
       \cellcolor[rgb]{.8,.8,.8}   \textbf{549} & 
        \cellcolor[rgb]{.8,.8,.8}   7 & 
        \cellcolor[rgb]{.8,.8,.8}  5 & 
        \cellcolor[rgb]{.8,.8,.8}  100 & 
       \cellcolor[rgb]{.8,.8,.8}   50 & 
       \cellcolor[rgb]{.8,.8,.8}   1 & 
       \cellcolor[rgb]{.8,.8,.8}   5 & 
       \cellcolor[rgb]{.8,.8,.8}   10 & 
       \cellcolor[rgb]{.8,.8,.8}   28 & 
       \cellcolor[rgb]{.8,.8,.8}   8 & 
       \cellcolor[rgb]{.8,.8,.8}   2 & 
       \cellcolor[rgb]{.8,.8,.8}   3
        \\
        \multirow{14}*{\rotatebox{90}{~~Diversity}} & 
        & 
        \begin{tabular}{@{}c@{}}Infinite scene-\\agnostic instantiation\end{tabular} & 
        \textcolor{indiagreen}\faCheck & 
        \textcolor{red}\faTimes & 
        \textcolor{red}\faTimes & 
        \textcolor{red}\faTimes & 
        \textcolor{red}\faTimes & 
        \textcolor{red}\faTimes & 
        \textcolor{red}\faTimes & 
        \textcolor{red}\faTimes & 
        \textcolor{red}\faTimes & 
        \textcolor{red}\faTimes & 
        \textcolor{red}\faTimes & 
        \textcolor{red}\faTimes & 
        \textcolor{red}\faTimes & 
        \textcolor{red}\faTimes & 
        \textcolor{red}\faTimes & 
        \textcolor{red}\faTimes & 
        \textcolor{red}\faTimes & 
        N/A
        \\ \cline{2-3}
        &         
        \multirow{2}*{\begin{tabular}{@{}c@{}}Diverse scenes\\and objects\end{tabular}} & 
        \cellcolor[rgb]{.8,.8,.8} \begin{tabular}{@{}c@{}}\# Object models\end{tabular} & 
        \cellcolor[rgb]{.8,.8,.8} \textbf{1217} & 
        \cellcolor[rgb]{.8,.8,.8} 118 &
        \cellcolor[rgb]{.8,.8,.8} \cellcolor[rgb]{.8,.8,.8} 112 & 
        \cellcolor[rgb]{.8,.8,.8} YCB &
        \cellcolor[rgb]{.8,.8,.8} \cellcolor[rgb]{.8,.8,.8} 150 & 
        \cellcolor[rgb]{.8,.8,.8} 152 & 
        \cellcolor[rgb]{.8,.8,.8} &
        \cellcolor[rgb]{.8,.8,.8} 84 &
       \cellcolor[rgb]{.8,.8,.8}  101 + YCB & 
       \cellcolor[rgb]{.8,.8,.8}  73+ & 
       \cellcolor[rgb]{.8,.8,.8}  28 & 
       \cellcolor[rgb]{.8,.8,.8}  80 & 
        \cellcolor[rgb]{.8,.8,.8} 10 & 
        \cellcolor[rgb]{.8,.8,.8} 4 & 
       \cellcolor[rgb]{.8,.8,.8}  4 & 
       \cellcolor[rgb]{.8,.8,.8}  4 & 
       \cellcolor[rgb]{.8,.8,.8} \cellcolor[rgb]{.8,.8,.8}  Matterport & 
       \cellcolor[rgb]{.8,.8,.8}  N/A
        \\ 
        & & 
        \# Scenes / Rooms & 
        \begin{tabular}{@{}c@{}}\textbf{15 / }\\\textbf{100}\end{tabular} & 
        \begin{tabular}{@{}c@{}}\textbf{- /} \\\textbf{120}\end{tabular} & 
        \begin{tabular}{@{}c@{}}\textbf{15 /} \\\textbf{90-120}\end{tabular} & 
        \begin{tabular}{@{}c@{}}55 static / \\-\end{tabular} & 
        \begin{tabular}{@{}c@{}}- /\\30\end{tabular} & 
        \begin{tabular}{@{}c@{}} 10 /\\-\end{tabular} & 
        \begin{tabular}{@{}c@{}}7 / \\-\end{tabular} &
        \begin{tabular}{@{}c@{}}- / \\120\end{tabular} & 
        \begin{tabular}{@{}c@{}}1 / \\-\end{tabular} & 
        \begin{tabular}{@{}c@{}}1 / \\-\end{tabular} & 
        \begin{tabular}{@{}c@{}}1 / \\-\end{tabular} & 
        \begin{tabular}{@{}c@{}}1 / \\-\end{tabular} & 
        \begin{tabular}{@{}c@{}}1 / \\-\end{tabular} & 
        \begin{tabular}{@{}c@{}}1 / \\-\end{tabular} & 
        \begin{tabular}{@{}c@{}}1 / \\-\end{tabular} & 
        \begin{tabular}{@{}c@{}}1 / \\-\end{tabular} & 
        \begin{tabular}{@{}c@{}}Matterport\\+ Gibson\end{tabular} &
         \begin{tabular}{@{}c@{}}572\\static\end{tabular} 
        \\ \cline{2-3}
        & 
        \multirow{9}*{\begin{tabular}{@{}c@{}}Diverse skills\\and activity reqs:\\Benchmark\\requires\\manipulating\ldots\end{tabular}} &  \cellcolor[rgb]{.8,.8,.8} \begin{tabular}{@{}c@{}}objects' pose\end{tabular} & 
        \cellcolor[rgb]{.8,.8,.8}  \textcolor{indiagreen}\faCheck & 
        \cellcolor[rgb]{.8,.8,.8}  \textcolor{indiagreen}\faCheck & 
        \cellcolor[rgb]{.8,.8,.8}  \textcolor{indiagreen}\faCheck & 
        \cellcolor[rgb]{.8,.8,.8}  \textcolor{indiagreen}\faCheck & 
        \cellcolor[rgb]{.8,.8,.8}  \textcolor{indiagreen}\faCheck & 
        \cellcolor[rgb]{.8,.8,.8}  \textcolor{indiagreen}\faCheck & 
       \cellcolor[rgb]{.8,.8,.8}   \textcolor{indiagreen}\faCheck & 
        \cellcolor[rgb]{.8,.8,.8}  \textcolor{indiagreen}\faCheck & 
       \cellcolor[rgb]{.8,.8,.8}   \textcolor{indiagreen}\faCheck & 
       \cellcolor[rgb]{.8,.8,.8}   \textcolor{indiagreen}\faCheck & 
       \cellcolor[rgb]{.8,.8,.8}   \textcolor{indiagreen}\faCheck & 
       \cellcolor[rgb]{.8,.8,.8}   \textcolor{indiagreen}\faCheck & 
       \cellcolor[rgb]{.8,.8,.8}   \textcolor{indiagreen}\faCheck & 
       \cellcolor[rgb]{.8,.8,.8}   \textcolor{indiagreen}\faCheck & 
       \cellcolor[rgb]{.8,.8,.8}   \textcolor{indiagreen}\faCheck &
         \cellcolor[rgb]{.8,.8,.8}   \textcolor{indiagreen}\faCheck &  
       \cellcolor[rgb]{.8,.8,.8}   \textcolor{red}\faTimes & 
       \cellcolor[rgb]{.8,.8,.8}   \textcolor{red}\faTimes
        \\
        & & \begin{tabular}{@{}c@{}}agent's global pose\end{tabular} & 
        \textcolor{indiagreen}\faCheck & 
        \textcolor{indiagreen}\faCheck & 
        \textcolor{indiagreen}\faCheck & 
        \textcolor{indiagreen}\faCheck & 
        \textcolor{indiagreen}\faCheck & 
        \textcolor{indiagreen}\faCheck & 
        \textcolor{indiagreen}\faCheck & 
        \textcolor{indiagreen}\faCheck & 
        \textcolor{red}\faTimes & 
        \textcolor{red}\faTimes & 
        \textcolor{red}\faTimes & 
        \textcolor{red}\faTimes & 
        \textcolor{red}\faTimes & 
        \textcolor{red}\faTimes & 
        \textcolor{indiagreen}\faCheck & 
        \textcolor{red}\faTimes & 
        \textcolor{indiagreen}\faCheck & 
        \textcolor{indiagreen}\faCheck 
        \\
        & & 
        \cellcolor[rgb]{.8,.8,.8} \begin{tabular}{@{}c@{}}objects's joint config\end{tabular} & 
        \cellcolor[rgb]{.8,.8,.8}  \textcolor{indiagreen}\faCheck & 
        \cellcolor[rgb]{.8,.8,.8}  \textcolor{indiagreen}\faCheck & 
        \cellcolor[rgb]{.8,.8,.8}  \textcolor{indiagreen}\faCheck & 
       \cellcolor[rgb]{.8,.8,.8}   \textcolor{indiagreen}\faCheck & 
       \cellcolor[rgb]{.8,.8,.8}   \textcolor{indiagreen}\faCheck & 
       \cellcolor[rgb]{.8,.8,.8}   \textcolor{indiagreen}\faCheck & 
       \cellcolor[rgb]{.8,.8,.8}   \textcolor{indiagreen}\faCheck & 
       \cellcolor[rgb]{.8,.8,.8}   \textcolor{indiagreen}\faCheck & 
       \cellcolor[rgb]{.8,.8,.8}   \textcolor{red}\faTimes & 
       \cellcolor[rgb]{.8,.8,.8}   \textcolor{indiagreen}\faCheck & 
       \cellcolor[rgb]{.8,.8,.8}   \textcolor{indiagreen}\faCheck & 
       \cellcolor[rgb]{.8,.8,.8}   \textcolor{indiagreen}\faCheck & 
       \cellcolor[rgb]{.8,.8,.8}   \textcolor{red}\faTimes & 
        \cellcolor[rgb]{.8,.8,.8}  \textcolor{indiagreen}\faCheck &  
        \cellcolor[rgb]{.8,.8,.8}  \textcolor{red}\faTimes & 
        \cellcolor[rgb]{.8,.8,.8}  \textcolor{indiagreen}\faCheck & 
        \cellcolor[rgb]{.8,.8,.8}  \textcolor{red}\faTimes & 
        \cellcolor[rgb]{.8,.8,.8}  \textcolor{red}\faTimes 
        \\
        & & \begin{tabular}{@{}c@{}}objects's geom.\end{tabular} & 
        \textcolor{indiagreen}\faCheck & 
        \textcolor{indiagreen}\faCheck & 
        \textcolor{red}\faTimes & 
        \textcolor{red}\faTimes & 
        \textcolor{red}\faTimes & 
        \textcolor{red}\faTimes & 
        \textcolor{indiagreen}\faCheck & 
        \textcolor{indiagreen}\faCheck & 
        \textcolor{red}\faTimes & 
        \textcolor{red}\faTimes & 
        \textcolor{red}\faTimes & 
        \textcolor{indiagreen}\faCheck & 
        \textcolor{red}\faTimes & 
        \textcolor{indiagreen}\faCheck & 
        \textcolor{red}\faTimes & 
        \textcolor{red}\faTimes & 
        \textcolor{red}\faTimes & 
        \textcolor{red}\faTimes 
        \\
        & & 
        \cellcolor[rgb]{.8,.8,.8}  \begin{tabular}{@{}c@{}}with two hands\end{tabular} & 
        \cellcolor[rgb]{.8,.8,.8}  \textcolor{indiagreen}\faCheck & 
        \cellcolor[rgb]{.8,.8,.8}  \textcolor{red}\faTimes & 
        \cellcolor[rgb]{.8,.8,.8}  \textcolor{red}\faTimes & 
        \cellcolor[rgb]{.8,.8,.8}  \textcolor{red}\faTimes & 
        \cellcolor[rgb]{.8,.8,.8}  \textcolor{red}\faTimes & 
        \cellcolor[rgb]{.8,.8,.8}  \textcolor{red}\faTimes & 
        \cellcolor[rgb]{.8,.8,.8}  \textcolor{indiagreen}\faCheck & 
        \cellcolor[rgb]{.8,.8,.8}  \textcolor{indiagreen}\faCheck & 
        \cellcolor[rgb]{.8,.8,.8}  \textcolor{red}\faTimes & 
        \cellcolor[rgb]{.8,.8,.8}  \textcolor{red}\faTimes & 
        \cellcolor[rgb]{.8,.8,.8}  \textcolor{red}\faTimes & 
        \cellcolor[rgb]{.8,.8,.8}  \textcolor{indiagreen}\faCheck & 
        \cellcolor[rgb]{.8,.8,.8}  \textcolor{red}\faTimes & 
        \cellcolor[rgb]{.8,.8,.8}  \textcolor{red}\faTimes & 
        \cellcolor[rgb]{.8,.8,.8}  \textcolor{red}\faTimes & 
        \cellcolor[rgb]{.8,.8,.8}  \textcolor{red}\faTimes & 
        \cellcolor[rgb]{.8,.8,.8}  \textcolor{red}\faTimes & 
        \cellcolor[rgb]{.8,.8,.8}  \textcolor{red}\faTimes 
        \\
        & & \begin{tabular}{@{}c@{}}objects's functional\\state (ON/OFF)\end{tabular} & 
        \textcolor{indiagreen}\faCheck & 
        \textcolor{red}\faTimes & 
        \textcolor{red}\faTimes & 
        \textcolor{red}\faTimes & 
        \textcolor{red}\faTimes & 
        \textcolor{red}\faTimes & 
        \textcolor{indiagreen}\faCheck & 
        \textcolor{indiagreen}\faCheck & 
        \textcolor{red}\faTimes & 
        \textcolor{indiagreen}\faCheck & 
        \textcolor{indiagreen}\faCheck & 
        \textcolor{red}\faTimes & 
        \textcolor{red}\faTimes & 
        \textcolor{red}\faTimes & 
        \textcolor{red}\faTimes & 
        \textcolor{red}\faTimes & 
        \textcolor{red}\faTimes & 
        \textcolor{red}\faTimes 
        \\
        & &  \cellcolor[rgb]{.8,.8,.8}  \begin{tabular}{@{}c@{}}with tools\end{tabular} & 
         \cellcolor[rgb]{.8,.8,.8}  \textcolor{indiagreen}\faCheck & 
        \cellcolor[rgb]{.8,.8,.8}   \textcolor{red}\faTimes & 
        \cellcolor[rgb]{.8,.8,.8}   \textcolor{red}\faTimes & 
        \cellcolor[rgb]{.8,.8,.8}   \textcolor{red}\faTimes & 
        \cellcolor[rgb]{.8,.8,.8}   \textcolor{red}\faTimes & 
        \cellcolor[rgb]{.8,.8,.8}   \textcolor{red}\faTimes & 
        \cellcolor[rgb]{.8,.8,.8}   \textcolor{indiagreen}\faCheck & 
        \cellcolor[rgb]{.8,.8,.8}   \textcolor{indiagreen}\faCheck & 
        \cellcolor[rgb]{.8,.8,.8}   \textcolor{red}\faTimes & 
        \cellcolor[rgb]{.8,.8,.8}   \textcolor{indiagreen}\faCheck & 
        \cellcolor[rgb]{.8,.8,.8}   \textcolor{red}\faTimes & 
        \cellcolor[rgb]{.8,.8,.8}   \textcolor{red}\faTimes & 
        \cellcolor[rgb]{.8,.8,.8}   \textcolor{red}\faTimes & 
        \cellcolor[rgb]{.8,.8,.8}   \textcolor{red}\faTimes & 
        \cellcolor[rgb]{.8,.8,.8}   \textcolor{red}\faTimes & 
        \cellcolor[rgb]{.8,.8,.8}   \textcolor{red}\faTimes & 
        \cellcolor[rgb]{.8,.8,.8} \cellcolor[rgb]{.8,.8,.8}   \textcolor{red}\faTimes & 
        \cellcolor[rgb]{.8,.8,.8}   \textcolor{red}\faTimes 
        \\ 
        & & \begin{tabular}{@{}c@{}}object's surface\end{tabular} & 
        \textcolor{indiagreen}\faCheck & 
        \textcolor{red}\faTimes & 
        \textcolor{red}\faTimes & 
        \textcolor{red}\faTimes & 
        \textcolor{red}\faTimes & 
        \textcolor{red}\faTimes & 
        \textcolor{red}\faTimes & 
        \textcolor{indiagreen}\faCheck & 
        \textcolor{indiagreen}\faCheck & 
        \textcolor{red}\faTimes & 
        \textcolor{red}\faTimes & 
       \textcolor{red}\faTimes & 
       \textcolor{red}\faTimes & 
        \textcolor{red}\faTimes & 
        \textcolor{indiagreen}\faCheck & 
        \textcolor{red}\faTimes & 
        \textcolor{red}\faTimes & 
        \textcolor{red}\faTimes 
         \\
        & 
        & 
         \cellcolor[rgb]{.8,.8,.8} \begin{tabular}{@{}c@{}}\cellcolor[rgb]{.8,.8,.8} objects's temp.\end{tabular} & 
         \cellcolor[rgb]{.8,.8,.8} \textcolor{indiagreen}\faCheck & 
        \cellcolor[rgb]{.8,.8,.8}  \textcolor{red}\faTimes & 
         \cellcolor[rgb]{.8,.8,.8} \textcolor{red}\faTimes & 
         \cellcolor[rgb]{.8,.8,.8} \textcolor{red}\faTimes & 
        \cellcolor[rgb]{.8,.8,.8}  \textcolor{red}\faTimes & 
       \cellcolor[rgb]{.8,.8,.8}   \textcolor{red}\faTimes & 
        \cellcolor[rgb]{.8,.8,.8}  \textcolor{red}\faTimes & 
        \cellcolor[rgb]{.8,.8,.8}  \textcolor{indiagreen}\faCheck & 
       \cellcolor[rgb]{.8,.8,.8}   \textcolor{red}\faTimes & 
       \cellcolor[rgb]{.8,.8,.8}   \textcolor{red}\faTimes & 
       \cellcolor[rgb]{.8,.8,.8}   \textcolor{red}\faTimes & 
       \cellcolor[rgb]{.8,.8,.8}   \textcolor{red}\faTimes & 
       \cellcolor[rgb]{.8,.8,.8}   \textcolor{red}\faTimes & 
       \cellcolor[rgb]{.8,.8,.8}   \textcolor{red}\faTimes & 
       \cellcolor[rgb]{.8,.8,.8}   \textcolor{red}\faTimes & 
       \cellcolor[rgb]{.8,.8,.8}   \textcolor{red}\faTimes & 
       \cellcolor[rgb]{.8,.8,.8}   \textcolor{red}\faTimes & 
       \cellcolor[rgb]{.8,.8,.8}   \textcolor{red}\faTimes
        \\\cline{1-3}
         & 
         \multirow{5}*{} & 
         \begin{tabular}{@{}c@{}}Activity\\length (steps)\end{tabular} &
         \begin{tabular}{@{}c@{}}\textbf{300-}\\\textbf{20000}\end{tabular} & 
         <100 & 
         100-1000 &
         100-1000 &
         <100 & 
         100-1000 & 
         <100 &
         <100 & 
         100-1000 & 
         <1000 &
         <100 &
         <100 &
         <100 &
         <100 &
         <100 &          
         <100 &
         <100 & 
         100-1000  
        \\
        \multirow{5}*{\rotatebox{90}{~~Complexity}} & 
        & 
        \cellcolor[rgb]{.8,.8,.8}\begin{tabular}{@{}c@{}} Objs. per activity\end{tabular} & 
        \cellcolor[rgb]{.8,.8,.8}\textbf{3-34} & 
        \cellcolor[rgb]{.8,.8,.8}5 &
        \cellcolor[rgb]{.8,.8,.8}7-9 &
        \cellcolor[rgb]{.8,.8,.8}2-5 &
        \cellcolor[rgb]{.8,.8,.8}2-3 & 
        \cellcolor[rgb]{.8,.8,.8}~10 & 
        \cellcolor[rgb]{.8,.8,.8}1-24 &
        \cellcolor[rgb]{.8,.8,.8}2 &
        \cellcolor[rgb]{.8,.8,.8}5-10 &
        \cellcolor[rgb]{.8,.8,.8}1-2 &
        \cellcolor[rgb]{.8,.8,.8}1-2 &
        \cellcolor[rgb]{.8,.8,.8}1 &
        \cellcolor[rgb]{.8,.8,.8}1-3 &
        \cellcolor[rgb]{.8,.8,.8}1-3 &
        \cellcolor[rgb]{.8,.8,.8}1-3 &
        \cellcolor[rgb]{.8,.8,.8}1 &
        \cellcolor[rgb]{.8,.8,.8}0-1 &
        \cellcolor[rgb]{.8,.8,.8}N/A
        \\
        & 
        & 
         \begin{tabular}{@{}c@{}}\# Obj. cats. in act.\end{tabular} & 
        2-17 & 
        1-5 & 
        7-10 & 
        2-5 & 
        1 &
        1 & 
        \textbf{1-18} & 
        2 & 
       1-10 & 
        1-2 & 
        1-2 & 
        1 & 
        1-2 & 
        1-3 & 
        1-3 & 
        1-2 & 
        0-1 & 
        N/A
        \\
        & 
        & 
        \cellcolor[rgb]{.8,.8,.8}\begin{tabular}{@{}c@{}}Diff. state changes required\\per activity (see \ref{fig:activitiesstatistics})\end{tabular} & 
        \cellcolor[rgb]{.8,.8,.8}\textbf{2-8} & 
        \cellcolor[rgb]{.8,.8,.8}4 & 
        \cellcolor[rgb]{.8,.8,.8}4 &
        \cellcolor[rgb]{.8,.8,.8}4 &
        \cellcolor[rgb]{.8,.8,.8}2 &
        \cellcolor[rgb]{.8,.8,.8}1-3 & 
        \cellcolor[rgb]{.8,.8,.8}1-7 & 
        \cellcolor[rgb]{.8,.8,.8}2-3 & 
        \cellcolor[rgb]{.8,.8,.8}1 & 
        \cellcolor[rgb]{.8,.8,.8}1-3 & 
        \cellcolor[rgb]{.8,.8,.8}1-4 & 
        \cellcolor[rgb]{.8,.8,.8}4 & 
        \cellcolor[rgb]{.8,.8,.8}1 & 
        \cellcolor[rgb]{.8,.8,.8}1-3 & 
        \cellcolor[rgb]{.8,.8,.8}1-2 & 
        \cellcolor[rgb]{.8,.8,.8}1-2 & 
        \cellcolor[rgb]{.8,.8,.8}1 & 
        \cellcolor[rgb]{.8,.8,.8}1
        \\
        & & 
         \begin{tabular}{@{}c@{}}Benchmark focus: \\Task-Planning\\and/or Control\end{tabular} &
          TP+C & 
          TP & 
          TP+C & 
          TP+C & 
          TP+C & 
          C & 
          TP &         
          TP & 
          TP+C & 
          C & 
          TP+C & 
          C & 
          C & 
          C & 
          C & 
          C & 
          C & 
          C 
        \\ \cline{1-3}
        \rot{}&
        \rot{}& 
        \cellcolor[rgb]{.8,.8,.8}  \begin{tabular}{@{}c@{}}\# Human VR demos\end{tabular} & 
        \cellcolor[rgb]{.8,.8,.8}  \textbf{400} & 
        \cellcolor[rgb]{.8,.8,.8}  0 & 
        \cellcolor[rgb]{.8,.8,.8}  0 & 
        \cellcolor[rgb]{.8,.8,.8}  0 & 
        \cellcolor[rgb]{.8,.8,.8}  0 & 
        \cellcolor[rgb]{.8,.8,.8}  0 & 
        \cellcolor[rgb]{.8,.8,.8}  0 & 
        \cellcolor[rgb]{.8,.8,.8}  0 & 
        \cellcolor[rgb]{.8,.8,.8}  0 & 
        \cellcolor[rgb]{.8,.8,.8}  0 & 
        \cellcolor[rgb]{.8,.8,.8}  0 & 
        \cellcolor[rgb]{.8,.8,.8}  0 & 
        \cellcolor[rgb]{.8,.8,.8}  0 & 
        \cellcolor[rgb]{.8,.8,.8}  0 & 
       \cellcolor[rgb]{.8,.8,.8}   0 & 
        \cellcolor[rgb]{.8,.8,.8}  0 & 
        \cellcolor[rgb]{.8,.8,.8}  0 & 
        \cellcolor[rgb]{.8,.8,.8}  0 
        \\
        \rot{}&
        \rot{}& 
         \begin{tabular}{@{}c@{}}\# Human VR demos\end{tabular} & 
         \textcolor{red}\faTimes & 
         \textcolor{red}\faTimes & 
         \textcolor{red}\faTimes & 
         \textcolor{red}\faTimes & 
         \textcolor{red}\faTimes & 
         \textcolor{red}\faTimes & 
         \textcolor{indiagreen}\faCheck & 
         \textcolor{red}\faTimes & 
         \textcolor{red}\faTimes & 
         \textcolor{indiagreen}\faCheck & 
         \textcolor{red}\faTimes & 
         \textcolor{red}\faTimes & 
         \textcolor{red}\faTimes & 
         \textcolor{red}\faTimes & 
         \textcolor{red}\faTimes & 
         \textcolor{red}\faTimes & 
         \textcolor{red}\faTimes & 
         \textcolor{red}\faTimes 
        \\\cline{3-21}
    \end{tabular}
  }
  \vspace{0.5em}
  \caption{Comparison between \model and other existing benchmarks for embodied AI. Expanded version of Table~\ref{tab:comparingbenchmarks}.}
  \label{tab:comparingbenchmarksfull}
\end{table}

\subsection{Defining \model Activities}
\label{ss:activitydefs}

This section includes additional information on how we define the 100 activities of \model, including details on 1) the process to select them from the American Time Use Survey~\cite{atus} (ATUS), 2) \bddl, the predicate logic language to define them, 3) the crowdsourcing process to generate definitions (initial and goal conditions) for the activities, 4) and real \bddl examples of the generated definitions.

\subsubsection{Selection of \numActivities Activities for \model}
\label{sss:statisticsactivities}

\begin{figure}[!t]
\centering
    \begin{subfigure}[t]{0.5\linewidth}
    \vskip 0pt
        \centering
        \includegraphics[width=\linewidth]{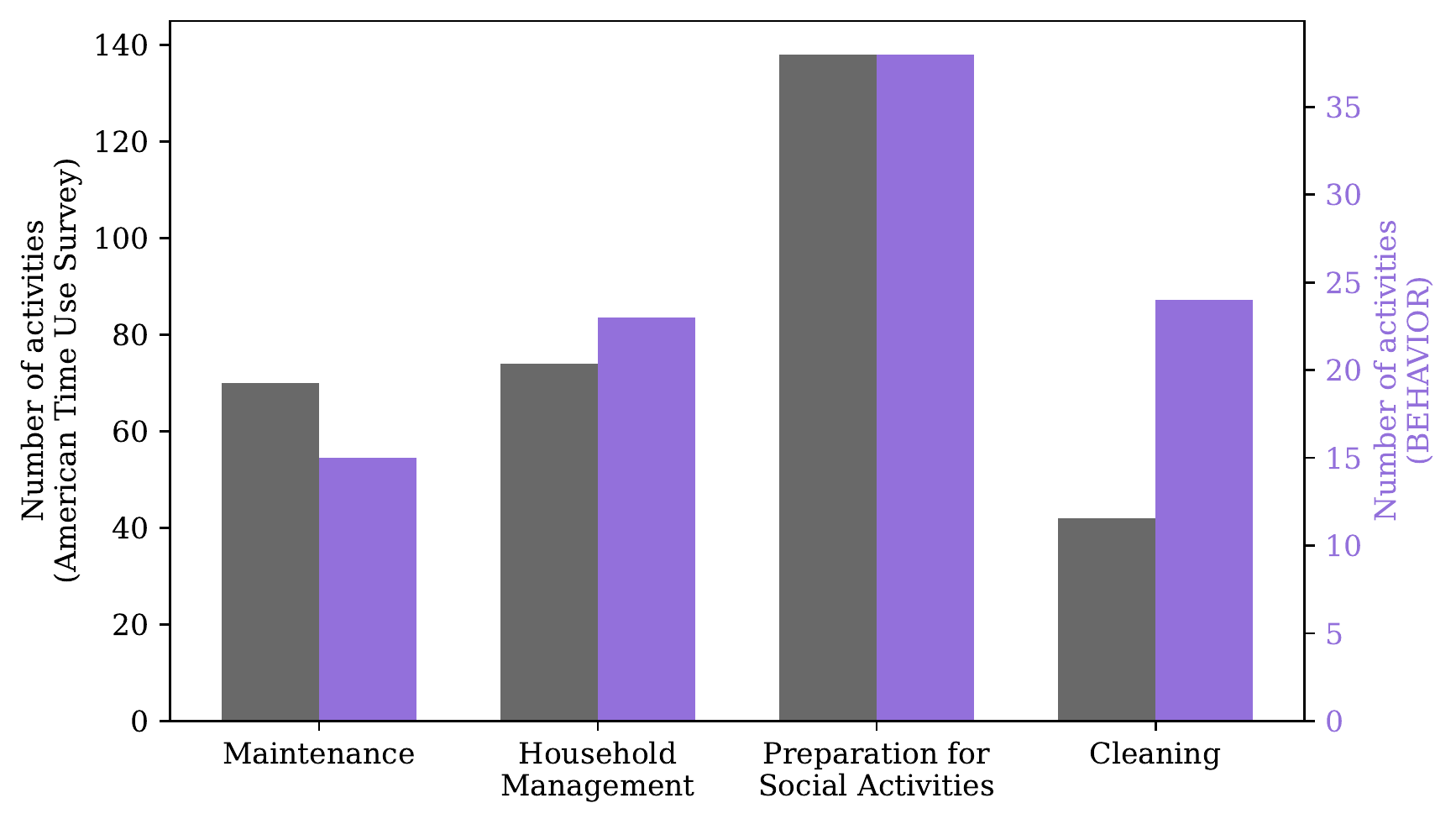} %
        \caption{Distribution Comparison of Activities in Categories from the American Time Use Survey~\cite{atus}}
        \label{fig:complexity_dist1}
    \end{subfigure}
    \hspace{0.5cm}
    \begin{subfigure}[t]{0.42\linewidth}
        \centering
        \vskip 0pt
        \includegraphics[width=\linewidth]{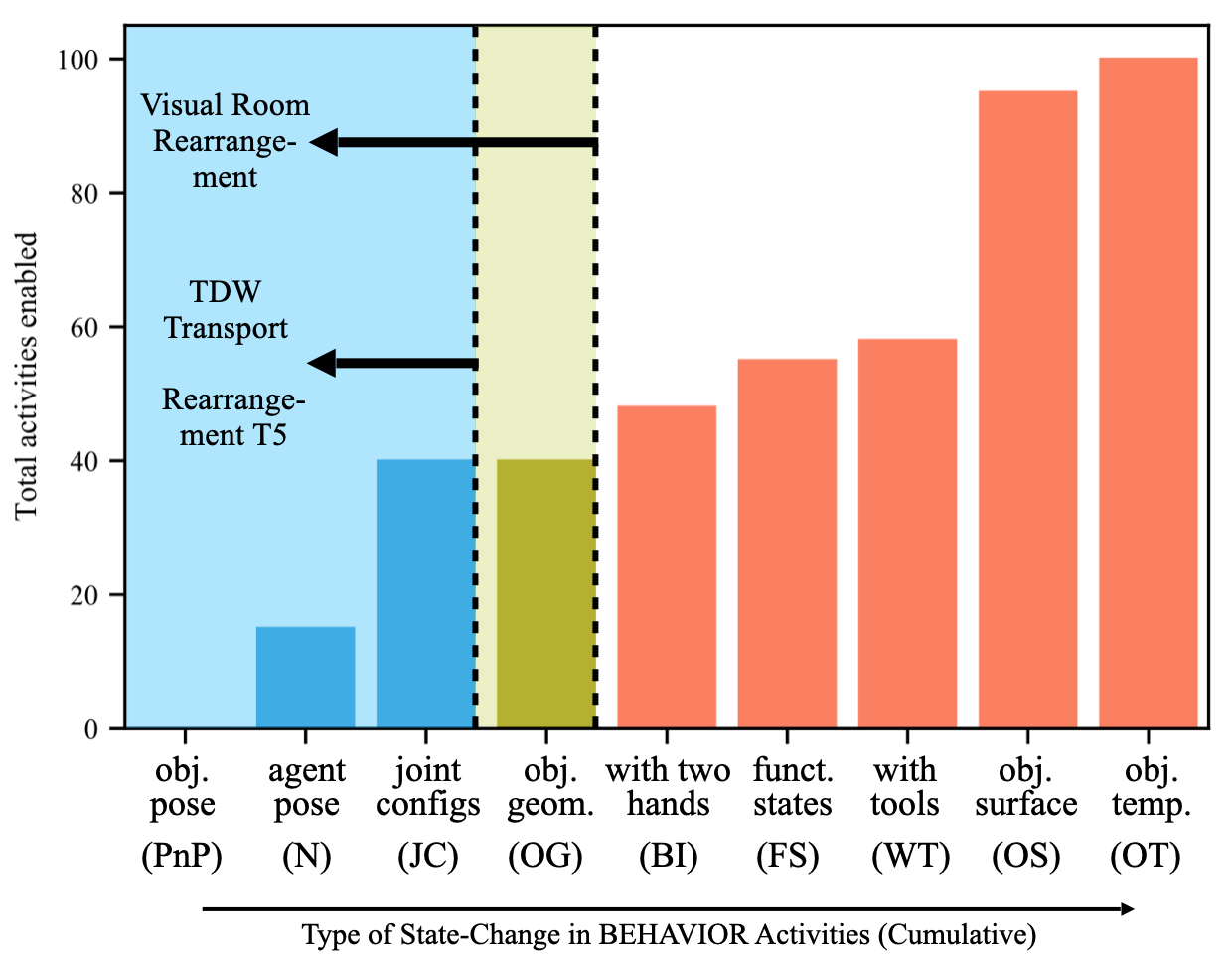}
        \caption{Number of Activities in \model enabled by Type of State-Change (cumulative)}
        \label{fig:complexity_dist2}
    \end{subfigure}
    \begin{subfigure}[b]{\linewidth}
        \centering
        \includegraphics[width=\linewidth]{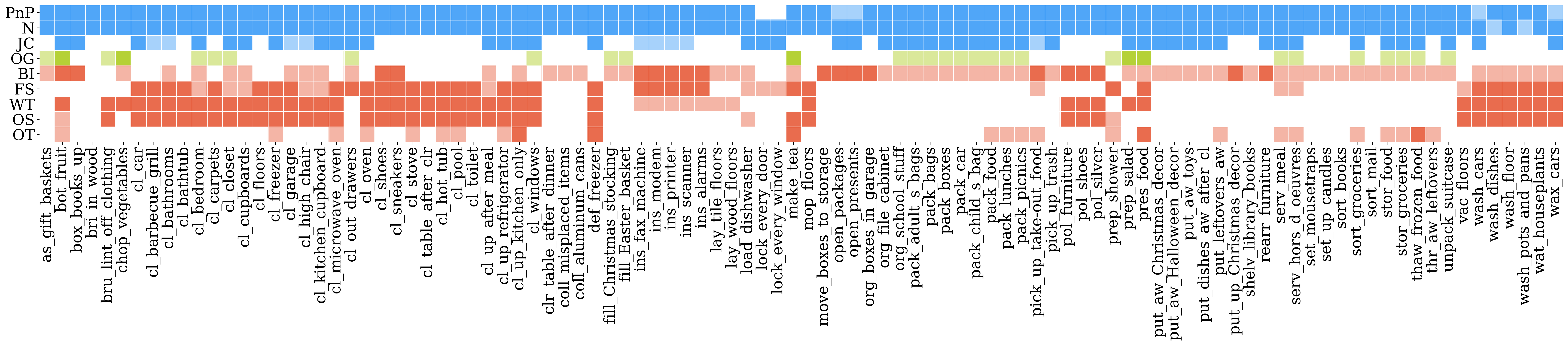} %
        \caption{Requirements of each Activity in \model}
        \label{fig:complexity_dist3}
    \end{subfigure}
    \caption{\textbf{Statistics of the \numActivities activities in \model:} a) Distribution of simulatable activities in the American Time Use Survey (left axis) and \model (right axis) based on categories from the survey -- \model covers a realistic distribution of activities; b) Cumulative visualization of activities enabled by different types of state changes in \model with comparison to recent prior work -- based on requirements, some activities could be considered transport/rearrangement (blue) or visual-room rearrangement (blue and green), while others are out of their scope (red); c) We visualize the specific requirements for each of the \model activities, with the same coloring scheme as in b). Activities in \model present significantly more diverse requirements than prior work focused on transport/rearrangement tasks~\cite{batra2020rearrangement,gan2021threedworld,weihs2021visual} enabling the evaluation of more general embodied AI solutions.
    }
    \label{fig:activitiesstatistics}
\end{figure}

Our activities are extracted from the American Time Use Survey~\cite{atus} that contains more than 2200 activities Americans spend their everyday time on. To select a subset for \model, we follow a set of criteria: \textbf{i) semantic diversity}: we select activities that span a wide range of semantic areas, from cleaning to food preparation, or repairing (see Fig.~\ref{fig:complexity_dist1}); \textbf{ii) diversity in the required state changes in the environment}: we select activities that requires manipulating different properties of the objects, their pose, temperature, cleanliness level, wetness level\ldots (see Fig.~\ref{fig:activitiesstatistics}, b and c); and \textbf{iii) simulation feasibility}: given the current state of simulation environments, we select for \model activities that can be realistically simulated entirely in an indoor environment, involving only objects, most of them rigid or articulated, excluding activities outdoors, interactions with other humans or animals, or heavy simulation of flexible materials and fluids. The resulting full list of \numActivities \model activities selected can be visualized in Fig.~\ref{fig:behavior_acts}. They cover a large variety of activities such as cleaning ({\small\texttt{CleaningBathtub}}, {\small\texttt{CleaningTheKitchenOnly}}, {\small\texttt{WashingPotsAndPans}}), installing ({\small\texttt{InstallingAScanner}}, {\small\texttt{InstallingAlarms}}), waxing/polishing ({\small\texttt{PolishingSilver}}, {\small\texttt{WaxingCarsOrOtherVehicles}}), tidying ({\small\texttt{PuttingAwayToys}}, {\small\texttt{PuttingDishesAwayAfterCleaning}}), packing/assembling ({\small\texttt{PackingPicnics}}, {\small\texttt{AssemblingGiftBaskets}}), and preparing food ({\small\texttt{PreservingFood}}, {\small\texttt{ChoppingVegetables}}). Fig.~\ref{fig:activitiesstatistics} depict statistics of the selected activities supporting that they approximate the semantic distribution of activities in the time use survey, and that they require a broad set changes in the environment. As comparison, Rearrangement tasks~\cite{batra2020rearrangement} and related benchmarks focus on activities that can be achieved by agent's pose (navigation), object's pose (pick-and-place), and joint configuration of articulated objects. VisualRoom Rearrangement~\cite{weihs2021visual} includes objects that can be broken (changing object geometry).

\begin{figure}[!t]
    \centering
    \includegraphics[width=\linewidth]{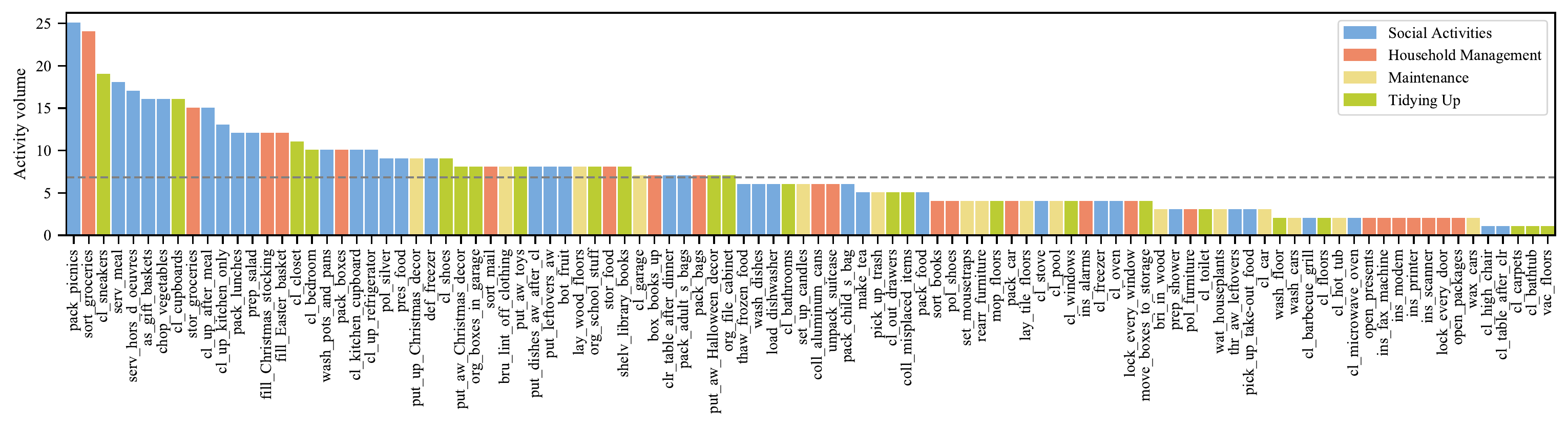}
    \caption{\textbf{Activity volume in \model:} The number of literals in flattened goal conditions (volume, see Sec.~\ref{sss:bddl}) provides a measure of the complexity of the activity and its length/horizon. The volume in \model activities span from one to 25 literals, very long horizon activities. Activities with one literal are often still long-horizon as they may require cleaning large surfaces (e.g. {\small\texttt{vacuumFloors}} or {\small\texttt{cleaningBathtub}})}
    \label{fig:act_volumes}
\end{figure}

\subsubsection{\bddl -- \model Domain Definition Language}
\label{sss:bddl}

In \model, activities are defined using a new predicate logic language, \bddl, \bddlfull. \bddl creates a logic-symbolic counterpart to the physical state simulated by \ig through a set of logic functions (predicates). In this way, \bddl defines a set of symbols grounded into simulated objects and their states. The goal of \bddl is to enable defining activities in a unique, unifying language that connects to natural language to facilitate interpretability. 
In this section, we provide additional information about the similarities and differences between \bddl and PDDL~\cite{mcdermott1998pddl}, a full description of the \bddl elements, syntax and grammar, and information about evaluation, grounding and ``flattening'' conditions, and the concept of ``activity volume''.

\paragraph{\bddl vs. PDDL:} While similar in name, \bddl is inspired by the Planning Domain Definition Language (PDDL)~\cite{mcdermott1998pddl} but strongly divergent. 
Both are derived from predicate logic and share a common logic-symbolic structure.
However, their goal and requirements are significantly different: while PDDL's main objective is to define a complete space for symbolic planning without any necessary connection to a physical world, \bddl's goal is to provide a diverse and fully-grounded symbolic representation of physical states to define activities as pairs of initial and goal logical conditions.
Therefore, PDDL requires to define additional symbols for agent's actions, while \bddl is only a representation of the state: agents act in the physical simulation to achieve the activities in \model.
To facilitate the adoption of \bddl as standard language to define activities in embodied AI, we assume the well-known syntax of PDDL for states.

\textbf{\bddl Syntax:} In \bddl, we consider the following syntactic elements, a subset of the syntax of predicate logic defined in~\citet{ullman1992focs}:
\begin{itemize}[
    topsep=0pt,
    noitemsep,
    leftmargin=6pt,
    itemindent=16pt]
    \item \textbf{Predicate:} logic function that takes as input one (unary) or two (binary) objects and returns a boolean value. Examples in \bddl: \texttt{ontop}, \texttt{stained}, \texttt{cooked}.
    \item \textbf{Variable:} element in a logical expression representing an object of the indicated category, always bound by a quantifier. Categories in \bddl are defined by WordNet~\cite{miller1995wordnet} synsets (semantic meaning), indicated by the label structure \texttt{categoryName.n.synsetEntry}. A variable is then indicated by a character \texttt{?} followed by the category. Examples in \bddl: \texttt{?apple.n.01}, \texttt{?table.n.02}.
    \item \textbf{Constant:} ground term, i.e., variable linked to a specific instance of an object. In \bddl, constants are identified by a numerical id suffix (\texttt{\_n}) appended to the variable name. Examples in \bddl: \texttt{apple.n.01\_1}, \texttt{table.n.02\_3}).
    
    \item \textbf{Category:} attribute of a constant or variable indicating the class of object it belongs to, and therefore which predicates it can be given as input to (e.g., \texttt{cooked}, \texttt{sliceable}). Examples in \bddl: \texttt{apple.n.01}, \texttt{table.n.02}.
    
    \item \textbf{Type:} synonymous with \textbf{category}, conventional for PDDL and therefore defined for \bddl.
    
    \item \textbf{Argument:} variable or constant used as input in a predicate.
    \item \textbf{Atomic formula:} single predicate with an appropriate number of arguments. Example in \bddl: \texttt{(onTop(apple.n.01\_1, table.n.02\_1)}
    \item \textbf{Logic operator:} Function mapping logical expressions to new logical expressions. In \bddl we include all four propositional logic operators: \texttt{and} ($\land$), \texttt{or} ($\lor$), \texttt{not} ($\lnot$), \texttt{if} ($\Rightarrow$), and \texttt{iff} ($\Leftrightarrow$).
    \item \textbf{Quantifier:} Function of a variable to map existing logical expressions to new logical expressions. In \bddl we include the standard universal quantification ($\forall$), and existential quantification ($\exists$), and additional operators: \texttt{for\_n}, \texttt{for\_pairs}, \texttt{for\_n\_pairs} (definitions below).
    \item \textbf{Logical expression:} expression obtained by composing atomic formulas with logical operators. Example in \bddl: \texttt{(and (onTop(apple.n.01\_1, table.n.02\_1)) (forall (?apple.n.01 - apple.n.01) cooked(apple.n.01))})
    \item \textbf{Initial condition:} set of atomic formulas that are guaranteed to be \texttt{True} at the beginning of all instances of the associated \model activity. See examples in Listings~\ref{lst:bddl1} and~\ref{lst:bddl2}.
    \item \textbf{Goal condition:} logical expression that must be \texttt{True} for the associated \model activity to be considered successfully executed. See examples in Listings~\ref{lst:bddl1} and~\ref{lst:bddl2}. 
    \item \textbf{Literal:} atomic formula or negated atomic formula. Example in \bddl: \texttt{not(onTop(apple.n.01\_1, table.n.02\_1))}
    \item \textbf{Fact:} ground atomic formula evaluated on the current state of the simulated world and returning a Boolean. Example in \bddl: \texttt{onTop(apple.n.01\_1, table.n.02\_1) = True}.
    \item \textbf{State:} set of facts about the current state of the simulated world providing a logical representation that can be evaluated wrt. the goal condition. 
\end{itemize}

Initial and final conditions for household activities could be expressed using the aforementioned first order logic syntax combined with \model's predicates. However, our activities are defined by non-technical annotators through a crowdsourcing procedure. The annotators are not required to have background knowledge in formal logic or computer science. To facilitate their work, we include the following additional non-standard quantifiers:
\begin{itemize}[
    topsep=0pt,
    noitemsep,
    leftmargin=6pt,
    itemindent=16pt]
    \item \texttt{for\_n}: for some non-negative integer $n$ and some object category $C$, the child condition must hold true for at least $n$ instances of category $C$
    \item \texttt{for\_pairs}: for two object categories $C1$ and $C2$, the child condition must hold true for some one-to-one mapping of object instances of $C1$ to object instances of $C2$ that covers all instances of at least one category 
    \item \texttt{for\_n\_pairs}: for some non-negative integer $n$ and two object categories $C1$ and $C2$, the child condition must hold true for at least $n$ pairs of instances of $C1$ and instances of $C2$ that follow a one-to-one mapping. 
\end{itemize}

Following the format of PDDL~\cite{mcdermott1998pddl}, in \bddl we consider two types of ``files'': a domain file shared for all activities, and problem files for each activity. The domain file defines all possible predicates, including object categories (corresponding in \model to categories from WordNet) and semantic symbolic states. Each activity in \model is defined by a different problem file that includes the object instances involved in the activity (categorized), the conditions for initial and final states.

\textbf{Evaluating Logical Expressions:} 
For a logical expression to be evaluated, we first decompose recursively it into subcomponents at the operators and quantifiers until we obtain a hierarchical structure of atomic formulae. 
Each atomic formula is composed of a predicate and arguments, i.e., a mathematical relationship on the simulated object(s) properties passed as arguments. 
For example, the atomic formula \texttt{(cooked apple.n.01\_1)} is evaluated by checking the relevant thermal information of the simulated object \texttt{apple.n.01\_1}. For details on the implementation of each predicate, see the attached cross-submission on the simulator \ig. Once the atomic formulae have been evaluated into facts with queries to the grounding simulated object states, we compose the facts through the logical operators to obtain the overall binary result of the whole expression. 
The BDDL symbolic definition of logical expressions creates flexibility: see Fig.~\ref{fig:multiple_init_goal} bottom row for examples of multiple correct solutions accepted by the same \bddl specification. 

\textbf{Instantiating and Grounding Initial Conditions:} The initial conditions of an activity in \model are defined at the beginning of each \bddl problem file. They include a list of object constants and a set of ground literals based on these constants. Instances of a \model activity are simulated physical states that fulfill all literals in the conditions. 
In our implementation of \bddl in \ig, the initial conditions are instantiated in the simulated state by assigning all object constants to physical objects of the appropriate category, either matching to physical objects already in the simulated scene or instantiating new ones in the locations specified by the binary atomic formulae (e.g., \texttt{ontop}, \texttt{inside}, etc.). 
The ground unary literals are satisfied by setting the physical states of the simulated objects according to the value their associated constants as given in the initial condition (e.g., \texttt{(not(cooked(chicken.n.01\_1)))} sets the temperature of the associated \texttt{chicken.n.01\_1} instance to a value that corresponds to \texttt{uncooked}).
Our instantiation of \bddl in \ig provides a sampling mechanism of unary and binary predicates that can generate potentially infinite variations of each set of initial conditions (more details in the \ig submission attached as supplementary) See Fig.~\ref{fig:multiple_init_goal} top row for examples of multiple instantiations from the same \bddl specification.

\textbf{``Flattening'' a Goal Condition in an Activity Instance:} BDDL provides a powerful mechanism to define the goal conditions in \model in their general form, e.g.,
\texttt{forall(?toy.n.01 - toy.n.01) inside(toy.n.01, box.n.01)}.
As logical expression, BDDL goal conditions are independent of the concrete objects and the scene, and thus valid to all instances and capturing all variants of the solution.
However, there are situations where grounding the goal conditions in the concrete instance of the activity at hand is helpful to understand the complexity (i.e., compute the activity volume), and the incremental progression towards the goal (i.e., compute the success score).
Following on the previous example, for a possible goal condition of \texttt{PickingUpToys}, the activity's complexity would be very different when the condition is applied on an activity instance (scene) with 100 toys or with only 1 toy.
We call goal condition ``flattening'' in an activity instance to the process of generating possible ground states of a specific simulated world fulfilling a condition. Flattening involves decomposing the nested structure of operators and quantifiers in the logical expression into a flat structure of disjointed conjunctions $C_i$ of ground literals $l_{j_i}$, {\small $\bigvee\limits_{C_i}\bigwedge\limits_{l_{j_i}}l_{j_i}$}, and grounding the literals in all possible ways in the given instance.
The final output of the flattening process is a list of \textit{options}, each of which is a list of ground literals that would satisfy the goal condition.
Because disjunctions, existential quantifiers, and \texttt{for\_n, for\_n\_pairs} are satisfied as soon as one/$n$ of their children is/are satisfied, our implementation of the flattening process for \bddl in \model acts lazily, generating only the minimal number of literals to fulfill each component of the goal condition. This prioritizes efficient solutions without losing any recall of possible solutions. 

\textbf{Activity Volume:} The result of flattening a goal condition in an activity instance is a list of possible options to accomplish the activity, each option being a list of ground atomic literals. We define the \textit{activity volume} as the length of the shortest flattened goal option for a given activity in a concrete instance. 
The activity volume provide a measure of the logical complexity of an activity, i.e., the number of atomic formulae that the agent needs to fulfill.
For our previous example for \texttt{PickingUpToys}, the activity would have a volume of $N$ for an activity instance with $N$ toys, indicating the different complexity for an instance with 100 or with 1 toys.

\subsubsection{Crowdsourcing the Annotation of Activities}
\label{ss:crowd}

Thanks to the connection in \bddl between the logical predicates and language semantics, \model activity definitions can be generated through crowdsourced annotation from non-experts workers, i.e.,  without background in computer science or logic. Through a visual interface, annotators can easily generate activity definitions in BDDL that reflect their idea of what the core of the activity is, and that are guaranteed to be simulatable in \ig.
We crowdsourced the generation of activity definitions to ensure that we do not introduce researcher biases in the design. The annotator pool was sourced from Upwork~\cite{upwork}, limiting to Upwork freelancers based in the United States of America to maintain consistency and familiary with ATUS activities. Each annotator was given a salary of \$15 per annotation, roughly \$20-30 per hour. Because the above process constitutes a complex annotation task, we developed a custom interface to guide and facilitate annotators' work, and guarantee simulatable output. 

\begin{figure}
    \centering
    \includegraphics[width=\linewidth]{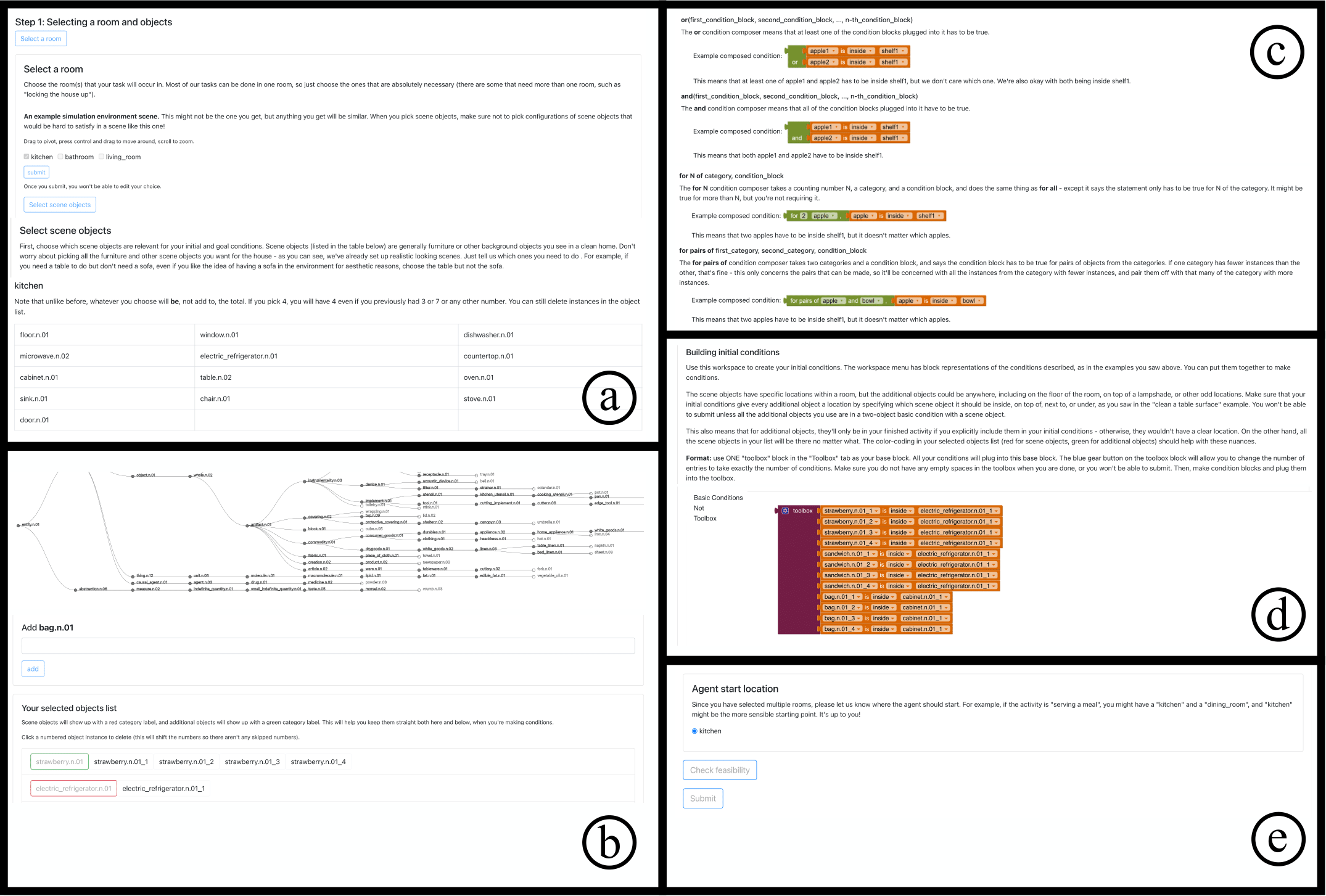}
    \caption{Sections of the interface given to activity definition annotators. \raisebox{.5pt}{\textcircled{\raisebox{-.9pt} {a}}} shows selection of relevant rooms and scene objects. For the purpose of creating definitions compatible with multiple \ig scenes and likely to fit with new scenes, annotators were allowed to pick scene objects from the intersection of object sets in three pre-selected scenes. \raisebox{.5pt}{\textcircled{\raisebox{-.9pt} {b}}} shows selection of additional objects that would be added to the scene during activity instantiation, sourced from wikiHow~\cite{wikihow} and taxonomized via WordNet~\cite{miller1995wordnet}. \raisebox{.5pt}{\textcircled{\raisebox{-.9pt} {c}}} shows examples of the Blockly~\cite{blockly} version of \bddl, and \raisebox{.5pt}{\textcircled{\raisebox{-.9pt} {d}}} shows the prompt for initial conditions and an example for a simple ``packing lunches'' definition. \raisebox{.5pt}{\textcircled{\raisebox{-.9pt} {e}}} shows the decision of the agent's start point and the interface for ``checking feasibility'', i.e. confirming that the \bddl is syntactically correct, the intial and goal conditions are satisfiable, and the set-up can be physically simulated in \ig by attempting a sampling in an \ig instance on a remote server. Not shown: introductory instructions, goal condition prompt and example (similar to initial condition), some \bddl blocks, remote server communication. Full interface available online: http://verified-states.herokuapp.com (server currently disabled).}
    \label{fig:annotation_interface}
\end{figure}
\begin{figure}[!t]
    \centering
    \includegraphics[width=\linewidth]{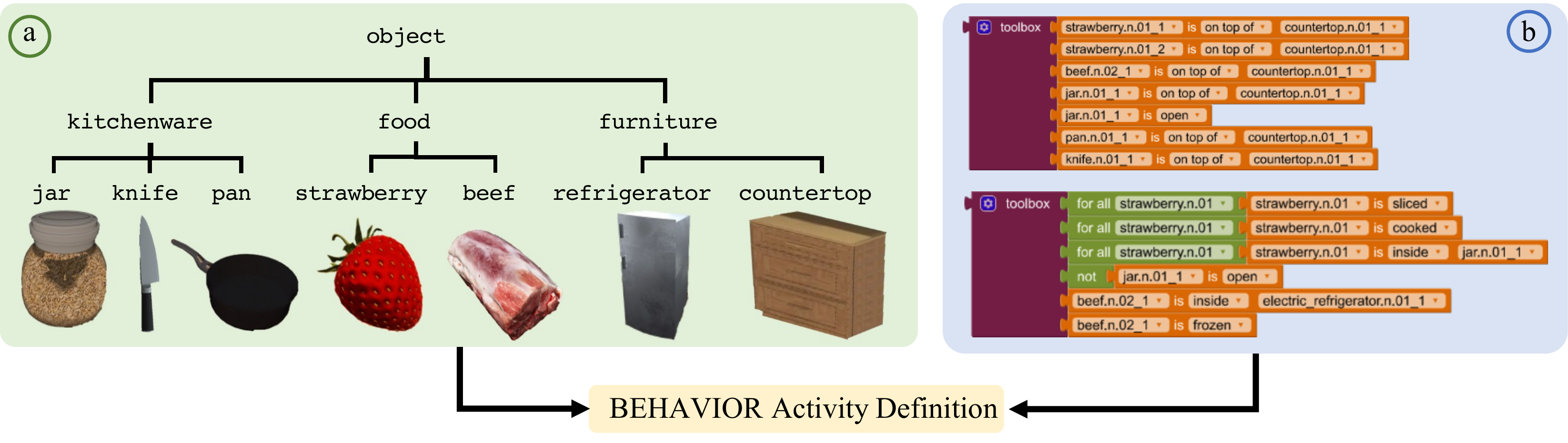}
    \caption{\textbf{Activity annotation process for \texttt{preserving\_food}:} a) annotators select objects from the WordNet organized \model Dataset of Objects; b) the selected objects are composed into logical predicates in \bddl for initial and final conditions using a visual interface derived from Blockly~\cite{blockly}; the result is a \bddl definition of the activity as logic predicates connected by logic operators and quantifiers, grounded in simulatable objects with physical properties}
    \label{fig:bddl_generation_diagram}
\end{figure}

\paragraph{Annotation Process and Interface:} The annotation procedure is as follows. First, the annotator is presented with a \model activity label. When necessary, we modify the original labels to add a numerical context, e.g., ``packing \textbf{four} lunches'' for the original \model activity ``packing lunches''. Then, the annotator reads the annotation instructions and enters the label into the interface. As response, the interface prompts the annotator to select one or more rooms that are relevant for the activity, and to choose objects already present in these rooms that are relevant to the activity (Fig.~\ref{fig:annotation_interface} (a)). The annotators then select small objects from the \model Dataset of Objects organized in the WordNet hierarchy (Fig.~\ref{fig:annotation_interface} (b)). 
To facilitate the annotation, instead of presenting the hierarchy for the entire \model Dataset, we preselect the most possible categories per activity based on a parsing procedure on how-to articles retrieved online, primarily from wikiHow~\cite{wikihow} (see Sec.~\ref{ss:datasetobj}). However, annotators can access the full hierarchy if the preselected items are not sufficient.

After this first phase to select activity-relevant objects, the annotator enters the second phase to annotate initial and goal conditions. First, they are introduced to a block-based, visual tool to generate \bddl (Fig.~\ref{fig:annotation_interface} (c)) built on Blockly~\cite{blockly}, which makes generating logical expressions intuitive and accessible to people without a programming background~\cite{puig2018virtualhome}. They use the tool to generate initial and goal conditions based on their concept of the activities (Fig.~\ref{fig:annotation_interface} (d)). The resulting definitions have several guarantees: 1) they only use objects from the \model Dataset of Objects, 2) they only apply logical predicates to objects in a semantically meaningful manner (e.g., \texttt{cooked} can only be applied to \texttt{cookable} objects such as food). This is because blocks' predicate fields are conditioned on entered categories that have been annotated with possible predicates in a separate manual WordNet annotation. 3) They will be in syntactically correct \bddl, through the implemented translation from Blockly. 4) They will not contain free variables or logically unsatisfiable conditions. 
5) It will be possible to simulate them physically in at least three simulated scenes from \ig. 
To guarantee feasibility, we assigned three possible home scenes from \ig to each activity and let the annotators evaluate the feasibility of their conditions at any point by clicking a button ``Check feasibility'' (Fig.~\ref{fig:annotation_interface} (e)). The request will send the \bddl definition to up to three \ig simulators on a remote server that will attempt to sample the initial conditions and check if the goal conditions are feasible, returning real-time feedback to the annotators to correct any unfeasible condition.
With the crowdsourcing procedure we obtain two alternative definitions per activity that are guarantee to be feasible in at least three simulated scenes.

\subsubsection{Example Definitions}

\definecolor{main-color}{rgb}{0,0,0}
\definecolor{back-color}{rgb}{0.942, 0.942, 0.93}
\definecolor{string-color}{rgb}{0.3333, 0.5254, 0.345}
\definecolor{key-color}{rgb}{0.847, 0.1059, 0.3765}
\definecolor{pred-color}{rgb}{0.1176, 0.5333, 0.898}
\definecolor{term-color}{rgb}{0., 0.31, 0.251}

\lstdefinestyle{mystyle}
{
    language = C,
    basicstyle = {\tiny \ttfamily \color{main-color}},
    backgroundcolor = {\color{back-color}},
    stringstyle = {\color{string-color}},
    keywordstyle = [2]{\color{pred-color}},
    keywordstyle = [3]{\color{key-color}},
    keywordstyle = {\color{term-color}},
    otherkeywords = {?},
    morekeywords = [2]{ontop, inside, inroom},
    morekeywords = [3]{define, problem, domain, objects, init, goal, and, for_n_pairs, forall, :, exists, and},
}

\noindent\begin{minipage}[b]{.48\textwidth}
\begin{lstlisting}[style = mystyle, caption={{\texttt{packing\_lunch}}},captionpos=b, label={lst:bddl1}]
(define 
  (problem packing_lunches_1)
  (:domain igibson)
    
  (:objects
    shelf.n.01_1 - shelf.n.01
    water.n.06_1 - water.n.06
    countertop.n.01_1 - countertop.n.01
    apple.n.01_1 - apple.n.01
    electric_refrigerator.n.01_1 - 
        electric_refrigerator.n.01
    hamburger.n.01_1 - hamburger.n.01
    basket.n.01_1 - basket.n.01
  )
    
  (:init 
    (ontop water.n.06_1 countertop.n.01_1)
    (inside apple.n.01_1 
        electric_refrigerator.n.01_1)
    (inside hamburger.n.01_1 
        electric_refrigerator.n.01_1)
    (ontop basket.n.01_1 countertop.n.01_1) 
    (inroom countertop.n.01_1 kitchen) 
    (inroom electric_refrigerator.n.01_1 
        kitchen) 
    (inroom shelf.n.01_1 kitchen)
  )
    
  (:goal 
    (and 
      (for_n_pairs
        (1)
        (?hamburger.n.01 - hamburger.n.01)
        (?basket.n.01 - basket.n.01)
        (inside ?hamburger.n.01 ?basket.n.01)
      )
      (for_n_pairs 
        (1) 
        (?basket.n.01 - basket.n.01) 
        (?water.n.06 - water.n.06) 
        (inside ?water.n.06 ?basket.n.01)
      ) 
      (for_n_pairs 
        (1) 
        (?basket.n.01 - basket.n.01) 
        (?apple.n.01 - apple.n.01) 
        (inside ?apple.n.01 ?basket.n.01)
      ) 
      (forall 
        (?basket.n.01 - basket.n.01) 
        (ontop ?basket.n.01 ?countertop.n.01_1) 
      )
    )
  )
)
\end{lstlisting}
\end{minipage}
\hfill
\noindent\begin{minipage}[b]{.48\textwidth}
\begin{lstlisting}[style=mystyle, caption={{\texttt{serving\_hors\_doeuvres}}},captionpos=b, label={lst:bddl2}]
(define 
  (problem serving_hors_d_oeuvres_1)
  (:domain igibson)

  (:objects
    tray.n.01_1 tray.n.01_2 - tray.n.01
    countertop.n.01_1 - countertop.n.01
    oven.n.01_1 - oven.n.01
    sausage.n.01_1 sausage.n.01_2 - sausage.n.01
    cherry.n.03_1 cherry.n.03_2 - cherry.n.03
    electric_refrigerator.n.01_1 - 
        electric_refrigerator.n.01
  )
    
  (:init 
    (ontop tray.n.01_1 countertop.n.01_1)
    (ontop tray.n.01_2 countertop.n.01_1)
    (inside sausage.n.01_1 oven.n.01_1)
    (inside sausage.n.01_2 oven.n.01_1)
    (inside cherry.n.03_1 
        electric_refrigerator.n.01_1)
    (inside cherry.n.03_2 
        electric_refrigerator.n.01_1)
    (inroom oven.n.01_1 kitchen)
    (inroom electric_refrigerator.n.01_1 
        kitchen)
    (inroom countertop.n.01_1 kitchen)
  )

  (:goal
    (and
      (exists
        (?tray.n.01 - tray.n.01)
        (and
          (forall
            (?sausage.n.01 - sausage.n.01)
            (ontop ?sausage.n.01 ?tray.n.01)
          )
          (forall
            (?cherry.n.03 - cherry.n.03)
            (not
              (ontop ?cherry.n.03 ?tray.n.01)
            )
          )
        )
      )
      (exists
        (?tray.n.01 - tray.n.01)
        (and
          (forall
            (?cherry.n.03 - cherry.n.03)
            (ontop ?cherry.n.03 ?tray.n.01)
          )
          (forall
            (?sausage.n.01 - sausage.n.01)
            (not
              (ontop ?sausage.n.01 ?tray.n.01)
            )
          )
        )
      )
    )
  )
)
\end{lstlisting}
\end{minipage}

In Listings~\ref{lst:bddl1} and \ref{lst:bddl2}, we include two examples of activity definitions (initial and goal conditions) in \bddl.
They are generating by mapping the input from crowdsourcing workers in our Blockly-like interface into \bddl language.
The activities include several objects and predicates in the initial and goal specifications.

\subsection{\ig}
\label{ss:igv2}

While \model is agnostic to the underlying simulator, we provide a fully functional instantation in \ig.
The details about \ig can be found in the cross-submission included in the supplementary material. 
Here we summarize its most important features in relation to \model.

\paragraph{Agents -- Realistic Sensing and Actuation:} We implement in \ig the two embodied agents mentioned in Sec.~\ref{s:instan} to perform \model activities: a bimanual humanoid and a Fetch robot.
Agents embodying the bimanual humanoid must control 24 degrees of freedom (DoF) to navigate, move and grasp (1 continuous DoF) with the hands, and move the pose of the head that controls the camera point of view. 
This is the embodiment used by humans in VR. Agents embodying the Fetch robot control 12 or 13 DoF: the navigating motion of the base, the pose of the end-effector (6 DoF), or alternatively, the joint configuration of the arm (7 DoF), one prismatic joint to grasp and release, and pan/tilt motion of the head that moves the cameras. 

\begin{figure}[!t]
    \centering
    \includegraphics[width=0.19\linewidth]{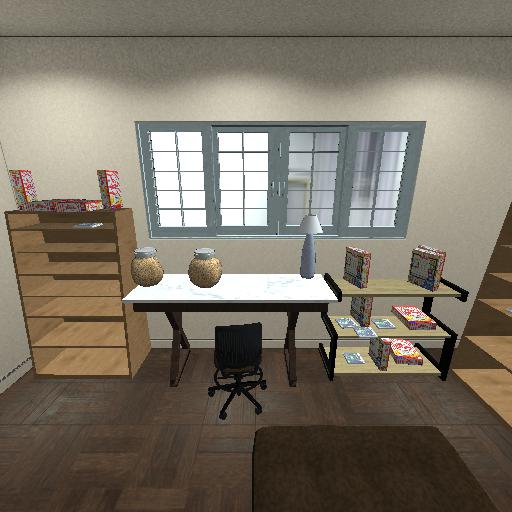}
    \hfill
    \includegraphics[width=0.19\linewidth]{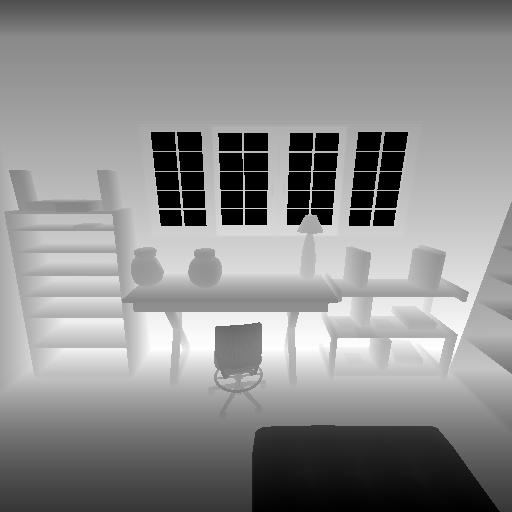}
    \hfill
    \includegraphics[width=0.19\linewidth]{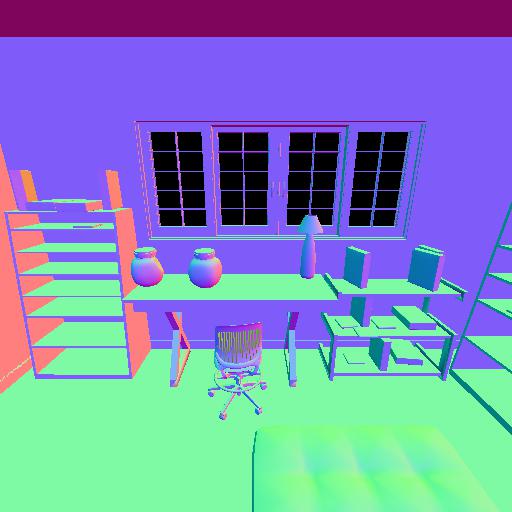}
    \hfill
    \includegraphics[width=0.19\linewidth]{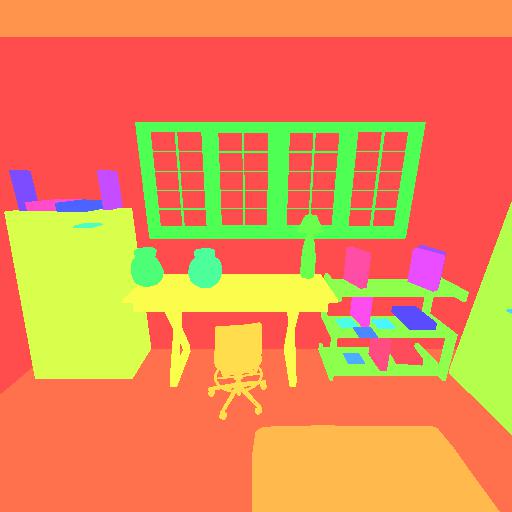}
    \hfill
    \includegraphics[width=0.19\linewidth]{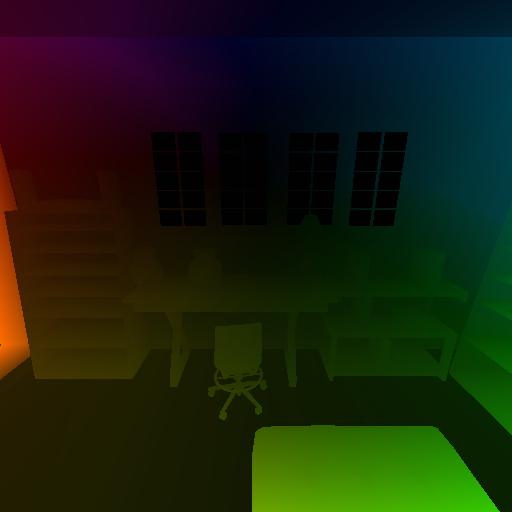}
    \caption{\textbf{Virtual visual sensor signals generated by \ig:} Color images are generated with a high-quality physics-based rendering procedure (PBR), exploiting the annotation of material (roughness, metallic) of all surfaces in our objects and scenes. \ig is able to generate RGB, depth, surface normals, semantic segmentation, instance segmentation, optical flow, scene flow and lidar (1-line and 16-line) sensors signals. Here we visualize a subset of those sensor signals, namely RGB, depth, surface normal, instance segmentation and optical flow.}
    \label{fig:igvisuals}
\end{figure}

The sensors used by humanoid agent and Fetch leverages the realistic sensor simulation from \ig. \ig features a physically-based renderer that can generate highly photorealistic RGB camera images, as well as other modalities, including depth, surface normal, semantic segmentation, instance segmentation, lidars, scene flows and optical flows. Fig.~\ref{fig:igvisuals} highlights a subset of the generated sensor signals. 

In terms of actuation, the actions are simulated accurately in pyBullet~\cite{coumans2016pybullet}, the physics engine used by \ig, with a very small physics simulation timestep of $\frac{1}{300}\text{s}$. The small physics timestep can reduce physics simulation artifacts, such as objects clipping into each other, increasing realism.

\begin{figure}[!t]
\centering
\includegraphics[width=0.32\linewidth]{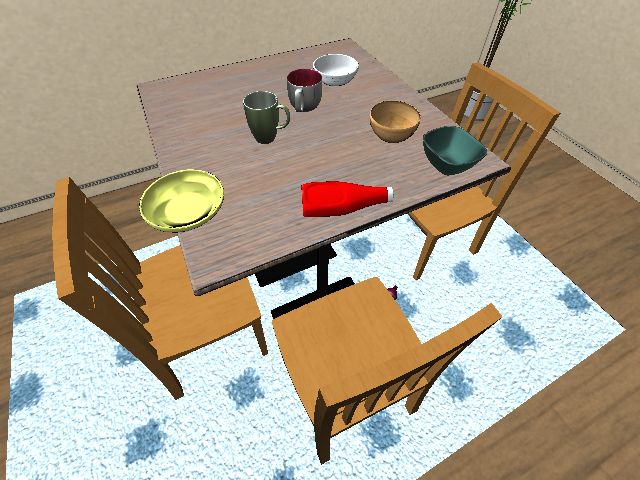}%
\hspace*{\hspacehere pt}%
\includegraphics[width=0.32\linewidth]{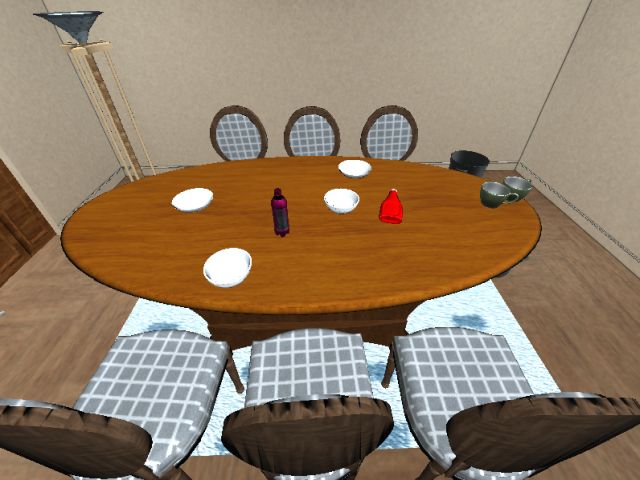}%
\hspace*{\hspacehere pt}%
\includegraphics[width=0.32\linewidth]{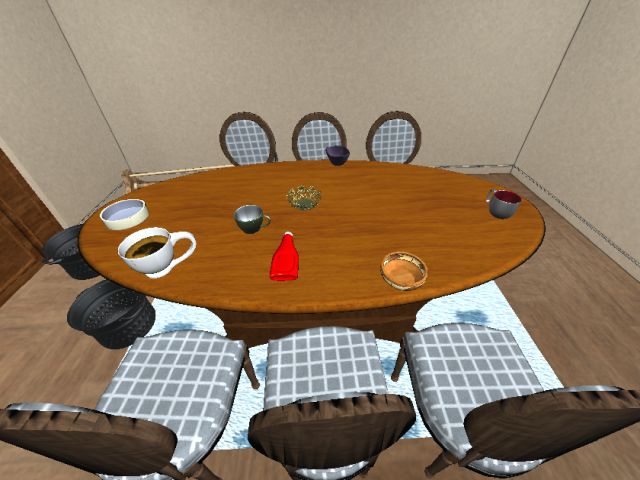}\vspace{\hspacehere pt}
\includegraphics[width=0.32\linewidth]{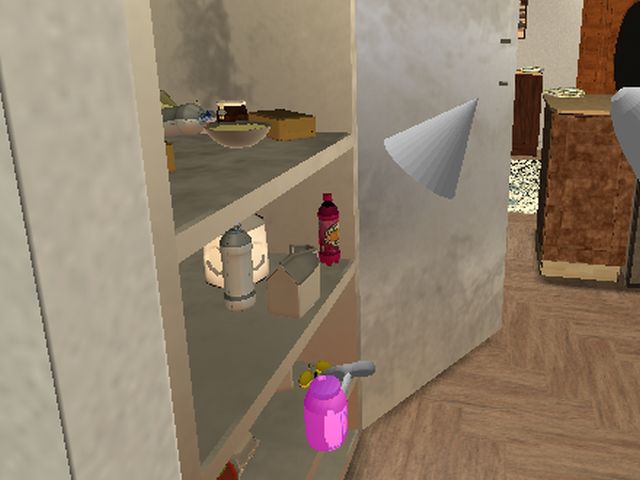}%
\hspace*{\hspacehere pt}%
\includegraphics[width=0.32\linewidth]{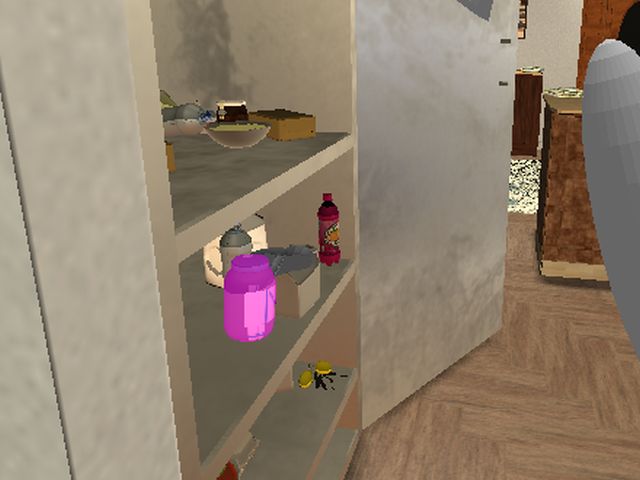}%
\hspace*{\hspacehere pt}%
\includegraphics[width=0.32\linewidth]{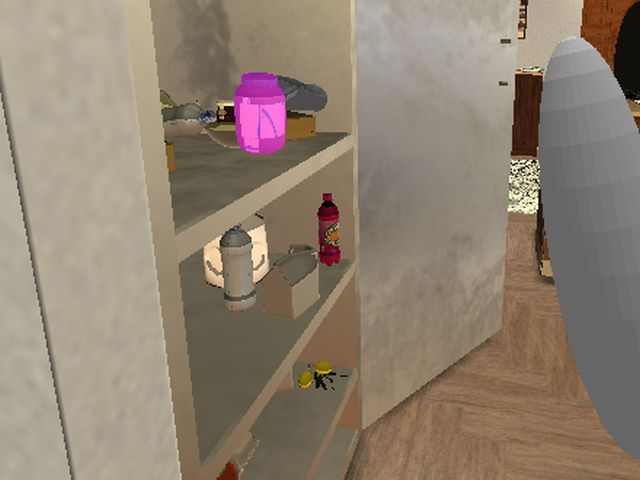}
\caption{\textbf{\bddl Initial and Goal Conditions:} Our implementation in \ig can generate diverse valid activity instances from each \bddl definition (top row), and detect all successful variations of the solution (bottom row), promoting diversity and semantically-meaningful activities}
\label{fig:multiple_init_goal}
\end{figure}

\paragraph{Condition Checking and Sampling:} The implementation of \model in \ig allows activities to be initialized, executed, and checked for completion. Given an activity definition in the BDDL, \model and \ig interface to generate a valid instance of the activity that satisfies the given object list and initial conditions. This mechanism can generate potentially infinite variation of scenes, objects and initial states to create different activity instances. In the generation of an activity instance, the goal conditions are checked for feasibility, avoiding the generation of activity instances that cannot lead to successful executions (see Fig.~\ref{fig:multiple_init_goal}, top). \ig implements all necessary checking functionalities for the logical states. These checking function execute in realtime together with the physical simulation and rendering, enabling live feedback to the agents for task completion and capturing all possible valid solutions (Fig.~\ref{fig:multiple_init_goal}, bottom). For more information about the condition checking and sampling, please refer to the concurrent submission \ig paper included as part of the supplementary material.

\paragraph{Implementation of the Action Primitives:} To facilitate the development of solutions and to study the effect of the activity complexity on the performance of embodied AI algorithms, we provide action primitives implemented in \ig and that can be used in \model. The action primitives are temporally extended actions. We implemented six action primitives, namely \texttt{navigate\_to(obj)}, \texttt{grasp(obj)}, \texttt{place\_onTop(obj)}, \texttt{place\_inside(obj)}, \texttt{open(obj)}, \texttt{close(obj)}. Each primitive can be applied relative to objects in the scene. For each action primitive, we implemented two variants. The first variant is ``fully-simulated motion primitive '', where we first check the feasibility of the target configuration, and then plan a full valid path between the initial and the target configurations with a sampling based motion planner~\cite{Jordan.Perez.ea:CSAIL13}. The second variant is ``partially-simulated motion primitive'', where we only check for feasibility of the desired final configuration, and directly set the state of the world (agent and objects) to this desired configuration. This can be highly unrealistic as we do not verify if there is a valid path between the initial and the final configurations. The purpose of partially simulated motion primitive is to reduce the computation during RL training and to measure the relative complexity of generating full interactions vs. just finding the sequence of states to achieve an activity. Note that for both partially and fully simulated motion primitives, privileged information is given to the agent and the motion planner. For example, the agent knows how many activity-relevant objects are in the scene, and the motion planner knows the full geometry of the environment.

For the implementation of partially-simulated motion primitives, we only perform feasibility check when attempting to perform an action. For example, when trying to \texttt{navigate\_to} an object, we will randomly sample points around the object and attempt to place the agent there: the goal is to find a collision-free location to place the agent. The second type of feasibility check is reachability: when attempting to \texttt{grasp}, \texttt{open} or \texttt{close} an object, we will check the distance from the hand to the closest point of the object is smaller than the arm length. When we \texttt{place} an object \texttt{inside} or \texttt{onTop} of an object, we use the sampling functionality available in \ig. 

For the implementation of fully-simulated motion primitives, in additional to the feasibility check, we attempt to plan and execute a collision free. We treat all objects as obstacles except for the objects given as argument for the primitive (e.g., objects that need to be picked up, receptacles that need to be opened), and plan a collision-free path from the start configuration to the target configuration. We use Bidirectional RRT~\cite{Jordan.Perez.ea:CSAIL13} for motion planning and execute the motion with position control. In our experiments, we found that fully simulated motion primitives have much lower success rate than partially simulated motion primitives (Table~\ref{tab:table_results1_am1}) indicating that the difficulty in \model strives from solving the entire interaction rather than deciding on the strategy at a task-level. Our partially-simulated primitives and other benchmarks that do not simulate the full interaction, bypass this critical challenge.

\paragraph{Runtime performance of \ig:} \ig improved performance when compared to iGibson 1.0~\cite{shen2020igibson}, with optimizations on both physics and rendering. To evaluate the performance of \ig in \model activities, we benchmarked the different phases of each simulation step. 
We benchmark the activities in ``idle'' setting, which means we initialize the activity, and runs the simulation and condition checking loop. The agent applies zero actions and stays still. We benchmarked in two conditions using the same action time step of $t_a$ but different physics time step of $t_s$, leading to slightly different reality in the physics simulation. The action step is the simulated-time between agent's actions, while the physics time step is the simulated-time interval that the kinematics simulator (pyBullet) uses to integrate forces and compute the new kinematic states. We execute $n_s$ queries to the simulator between agent actions, with $n_s = t_a/t_s$. The first condition we evaluate uses action time step $t_a = \frac{1}{30}\text{s}$ and physics time step of $t_s = \frac{1}{300}\text{s}$, which creates high-fidelity physics simulation.
The second condition uses action time step of $t_a = \frac{1}{30}\text{s}$ and physics time step of $t_s = \frac{1}{120}\text{s}$, which has slightly lower physics fidelity, but has better performance and is sufficient for RL training. 
Both settings are benchmarked on a computer with Intel 5930k CPU and Nvidia GTX 1080 Ti GPU, in a single process setting, rendering 128$\times$128 RGB-D images.

As shown in Table~\ref{tab:speed_benchmark}, for the highest-fidelity physics setup, we can achieve 36-59 steps per second, 47-71 steps per second with larger simulated timestep, even in a very large scene with 100-200 movable objects, and with all the physical and logical states evaluated at each step.
This frequencies provide pleasant experience in virtual reality.
However, it only provide a $~\times$2 acceleration over clock-time to train RL agents.
To increase the frequency in simulation and reduce the training time, we are exploring the parallelization of simulation and rendering and the more aggressive ``sleep'' of non-interacted objects.

\begin{table}[]
    \centering
    \resizebox{\linewidth}{!}{
    \begin{tabular}{|c|c|c|c|}
    \hline
         & \texttt{bringing\_in\_wood}	& \texttt{re-shelving\_library\_books} &  \texttt{laying\_tile\_floors} \\
         \hline
       Number of Objects & 134 & 144 & 216\\
        \hline 
            Simulation steps per second (@$t_s = \frac{1}{300}s$ / @$t_s = \frac{1}{120}s$) & 59 / 74 & 51 / 68 & 36 / 47\\
        \hline
        
        Kinematic State Update Time [ms] (@$t_s = \frac{1}{300}s$ / @$t_s = \frac{1}{120}s$) & 7.4 / 3.5 & 9.4 / 4.2 & 12.6 / 5.7\\
        \hline

        Non-kinematic State Update Time [ms] & 3.4 & 3.4 & 5.2\\
        \hline
        Rendering Time [ms] & 5.8 & 6.1 & 9.3\\
        \hline
        Logical Condition Checking Time [ms] & 0.4 & 0.4 & 0.6\\
        \hline
    \end{tabular}
    }
    \caption{Benchmarking Simulation Time for \model Activities in \ig }
    \label{tab:speed_benchmark}
\end{table}

\subsection{\model Dataset of Objects}
\label{ss:datasetobj}

\begin{figure}[!t]
    \centering
    \includegraphics[width=\linewidth]{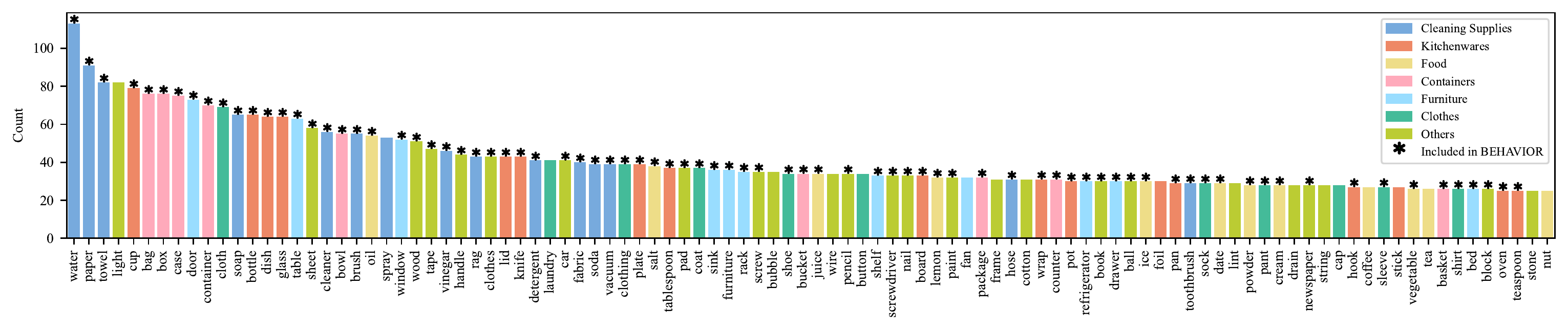}
    \caption{\textbf{Statistics of objects in descriptions for the \numActivities \model activities:} We parse descriptions from WikiHow~\cite{wikihow} and other online repositories of instructions for the activities in \model and obtain the frequencies of appearance for each noun. The nouns are mapped to corresponding synsets in WordNet~\cite{miller1995wordnet}. The categories shown in the figure are based on the WordNet taxonomy (best seen in color). For object categories included in the \model Dataset of Objects, we annotate the bar with an asterisk (the plot does not depict all categories in the dataset). We include the vast majority of most frequent objects involved in the activities as indicated by the natural language descriptions.}
    \label{fig:objects_from_wiki}
    \vspace{-1em}
\end{figure}

In order to instantiate \model activities in \ig, we created a new dataset of everyday objects, the \model Dataset of Objects. 
To guide the selection of object categories, we analyze how-to articles, primarily WikiHow~\cite{wikihow}, explaining how to perform the activities included in \model. Specifically, we extract nouns of tangible objects from these articles that are activity-relevant, map them to WordNet synsets, and then purchase 3D models of these object categories from online marketplaces such as TurboSquid. This procedure allowed us to provide activity annotators and VR demonstrators with the most frequent objects necessary for the activities (see Fig.~\ref{fig:objects_from_wiki}).

The diversity of \model activities naturally leads to the diversity of the object dataset. In total, we curate 1217 object models across 391 object categories, to support \numActivities \model activities. The categories range from food items to tableware, from home decorations to office supplies, and from apparel to cleaning tools. In Fig.~\ref{fig:objects_from_wiki}, we observe that the \model Dataset of Objects cover a wide range of object categories.

\begin{figure}[!t]
    \centering
    \includegraphics[width=0.9\linewidth]{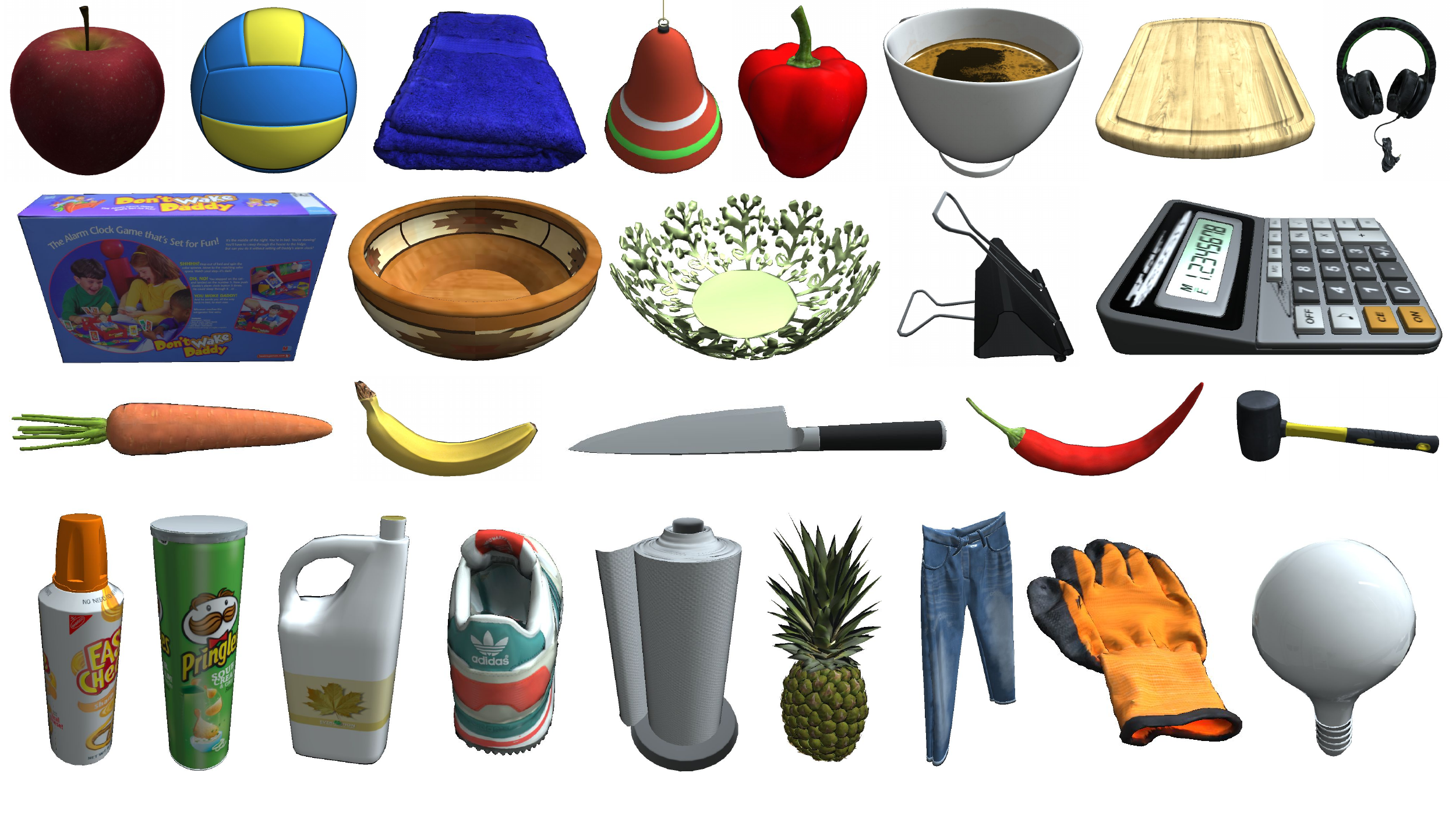}
    \caption{\textbf{Example Models in \model Dataset of Objects:} A selected subset of everyday objects in the dataset to support the \numActivities activities in \model. The models present high-quality geometry, material, and texture, and are annotated with realistic physical attributes such as size, mass, center of mass and moment of inertia, and semantic properties such as \texttt{cookable}, \texttt{sliceable}, or \texttt{toggleable}.}
    \label{fig:supp_object_collage}
    \vspace{-1em}
\end{figure}

To maintain high visual realism, all object models include material information (metallic, roughness) that can be rendered by \ig renderer. To maintain high physics realism, object models are annotated with size, mass, center of mass, moment of inertia, and also stable orientations. The collision mesh is a simplified version of the visual mesh, obtained with a convex decomposition using the VHACD algorithm. Object models with a shape close to a box are annotated with a primitive box collision mesh, much more efficient and robust for collision checking. For object categories that have the semantic property \texttt{openable} annotated, we make sure at least a subset of their object models have articulation, e.g. openable jars, backpacks, cars, etc. We either directly acquire them from the PartNet-Mobility Dataset~\cite{xiang2020sapien} or acquire non-articulated models from TurboSquid, manually segment the models into parts, and then create the articulation in the URDF files. A subset of the object models are visualized in Fig.~\ref{fig:supp_object_collage}.

We will publicly release the object dataset to be used for \model benchmarking. To preserve the rights of the model authors and the license agreement with TurboSquid, the 3D models are encrypted so that they can only be used within \ig and cannot be exported for other applications. 

\begin{figure}[!t]
    \centering
    \includegraphics[width=0.95\linewidth]{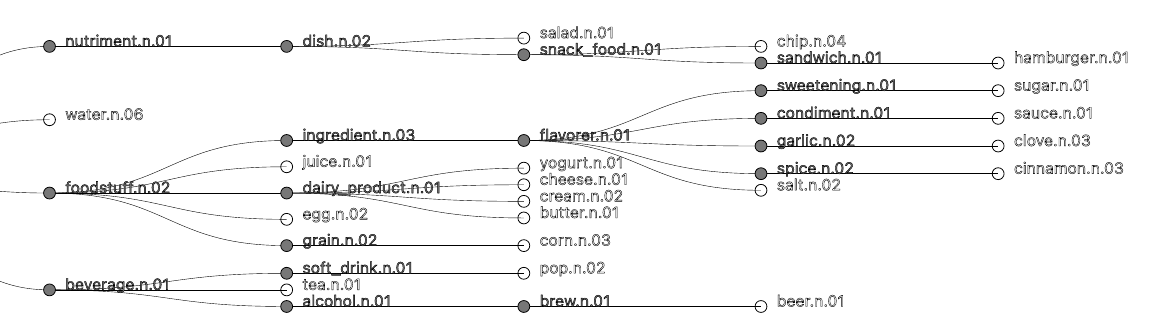}
    \caption{\textbf{Object Taxonomy in the \model Dataset of Objects:} Sample extract of the objects involved in an activity in \model, organized based on the taxonomy from WordNet~\cite{miller1995wordnet}; We map 3D models with annotation of physical and semantic properties to synsets in WordNet. \model activities in \bddl use any level entries of the WordNet taxonomy enabling the generation of more diverse instances with any object in the downstream task (e.g. any food item)}
    \label{fig:object_taxonomy}
\end{figure}

All models in the \model Dataset are organized following the WordNet~\cite{miller1995wordnet}, associating them to synsets. 
This structure allows us to define properties for all models of the same categories, but it also facilitates more general sampling of activity instances fulfilling initial conditions such as \texttt{onTop(fruit, table)} that can be achieved using any model within the branch \texttt{fruit} of WordNet. Fig.~\ref{fig:object_taxonomy} shows an example taxonomy of objects of the dataset organized in the WordNet taxonomy to perform a given household activity.

\subsection{\model Dataset of Human Demonstrations in Virtual Reality}
\label{ss:hd_vr_appendix}

The main role of the VR demonstrations in \model is to provide a mechanism to normalize metrics, allowing to compare different embodied AI solutions between activity instances and scenes. However, we believe that the generated dataset of VR demos has the potential to be applied to other purposes, e.g., to generate AI solutions through imitation learning, or to study the mechanisms used by humans to accomplish interactive activities. In the following, we provide additional details for users interested in the dataset of VR demonstrations. We include 1) additional information about the data collection procedure, 2) statistics of the data, and 3) information about collected human gaze data.

\subsubsection{Collecting Human Demonstrations in Virtual Reality}
\label{sss:vr_details}

To generate data, humans control a bimanual humanoid embodiment with a main body, two hands and a movable head based on stereo images displayed at 30 frames per second. The embodiment and the VR can be used with the most common VR hardware but for our dataset, we used a HTC Vive Pro Eye~\cite{htcviveproeye}. All recorded data can be deterministically replayed, achieving the same physical state transitions as reaction to the recorded physical interactions, which allows to generate any additional virtual sensor signal a posteriori. For more information about the VR interface, we provide the cross-submitted publication of \ig as part of the supplementary material. 

We collect {three} different demonstrations of the same activity instance (same scene, same objects, same initialization) for each of the \numActivities activities in \model, {100 additional demonstrations}, {one} for each activity for a different instance (different objects, different initialization) in the same scene, and {100 additional demonstrations}, {one} for each activity in a different scene. This 500 demonstrations cover both the diversity in human execution, and the dimensions of variability in activity instances of \model. The data has been collected by voluntary participants and our own team.

\subsubsection{Analysis and Statistics of Virtual Reality Demonstrations}
\label{sss:vr_stats2}

\begin{figure}[!t]
    \centering
    \includegraphics[width=\linewidth]{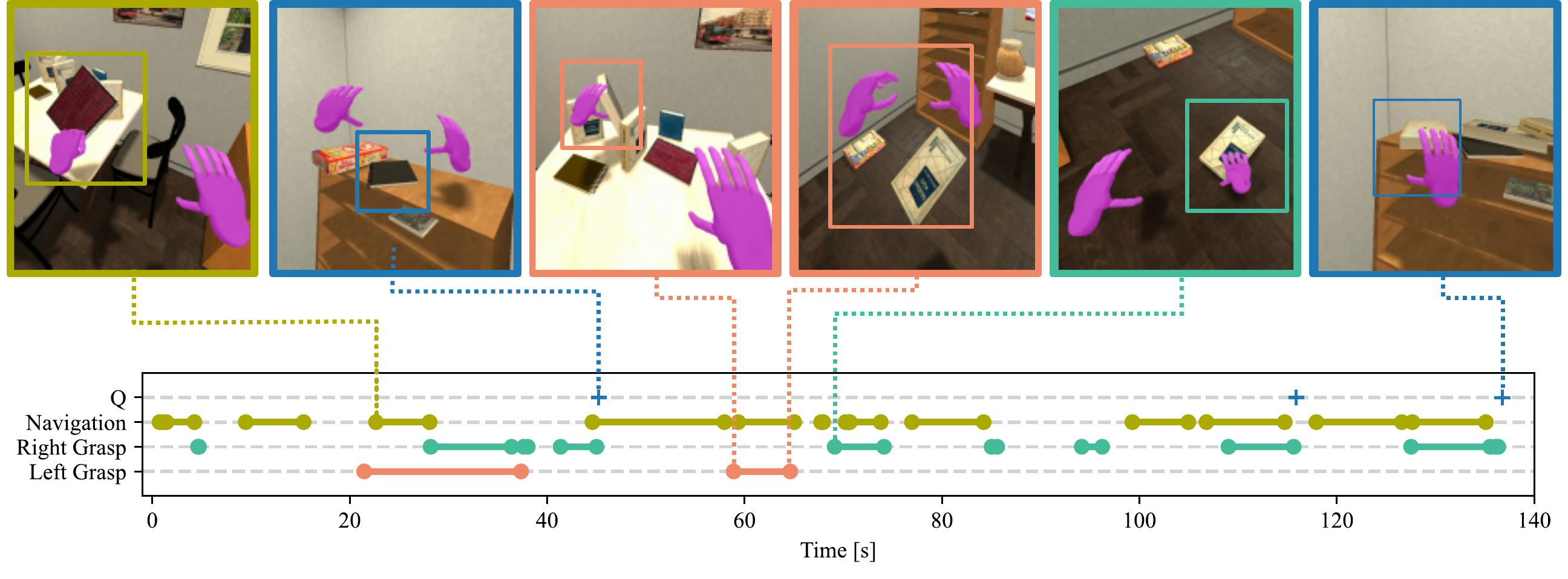}
    \caption{\textbf{Sub-activity segmentation across activity execution for {\small\texttt{re-shelvingLibraryBooks}}:} We observe multiple cycles of long-range pick-and-place operations that eventually lead to activity success. In this figure, we show a sequence of snapshots of first-person view along with key frames (i.e. target objects placed on shelf, items dropped and picked up with alternating hands).}
    \label{fig:task_segmentation}
\end{figure}

The \model Dataset of Human Demonstrations in VR provides rich data of navigation, manipulation, and problem-solving from humans for long time-horizon and multi-step activities. 
Analyzing the statistical characteristics of the data (duration, hand use, room visitation, etc.) provides insights on how humans achieve their level of performance combining interaction and locomotion in the large \model scenes. 
Fig.~\ref{fig:task_segmentation} depicts the segmentation of a VR demonstration into navigation and grasping phases while performing a pick-and-place rearrangement activity. This segmentation reveals multiple initial phases of navigation as the demonstrator observes the scene and locates activity relevant objects. For example, once the demonstrator reaches the table supporting the target objects (approx. at 18s), they pick up the target object with the non-dominant hand (approx. at 20s) and navigate to the goal location, before transferring the object to their dominant hand while positioning it (approx. at 30s). The demonstrator shows a preference for moving objects one-at-a-time, instead of stacking or carrying objects with each hand; this strategy will perform more poorly on the efficiency metrics $T_{sim}$, $L_{body}$, $L_{right}$, and $L_{left}$.

\begin{figure}[!t]
\centering
\begin{subfigure}[t]{0.99\linewidth}
\centering
\includegraphics[width=\linewidth]{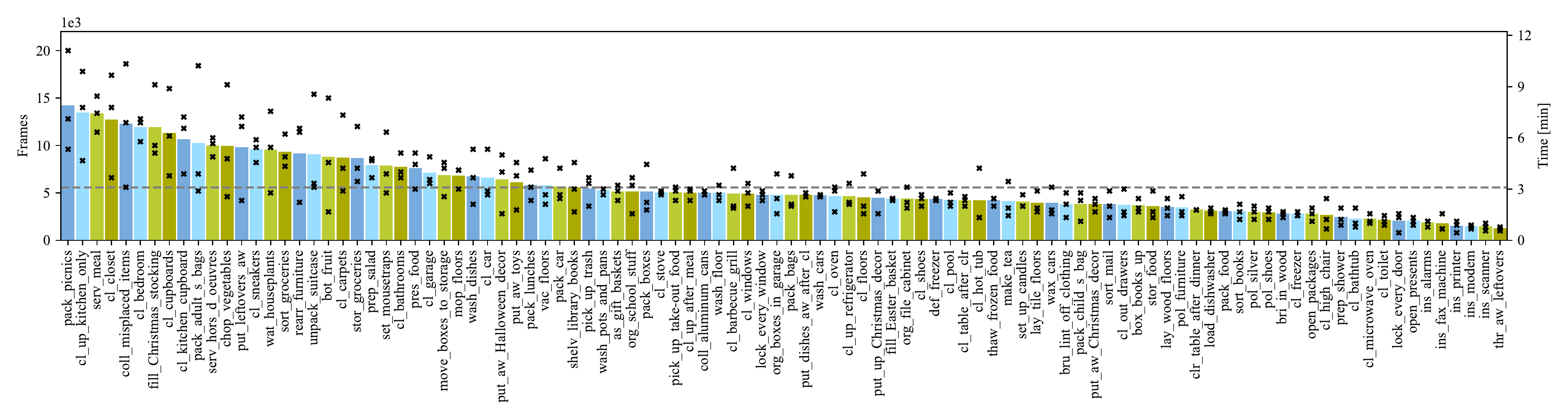}
\caption{}
\label{fig:timeact}
\end{subfigure}\\
\begin{subfigure}[t]{0.3\linewidth}
\centering
\includegraphics[width=\linewidth]{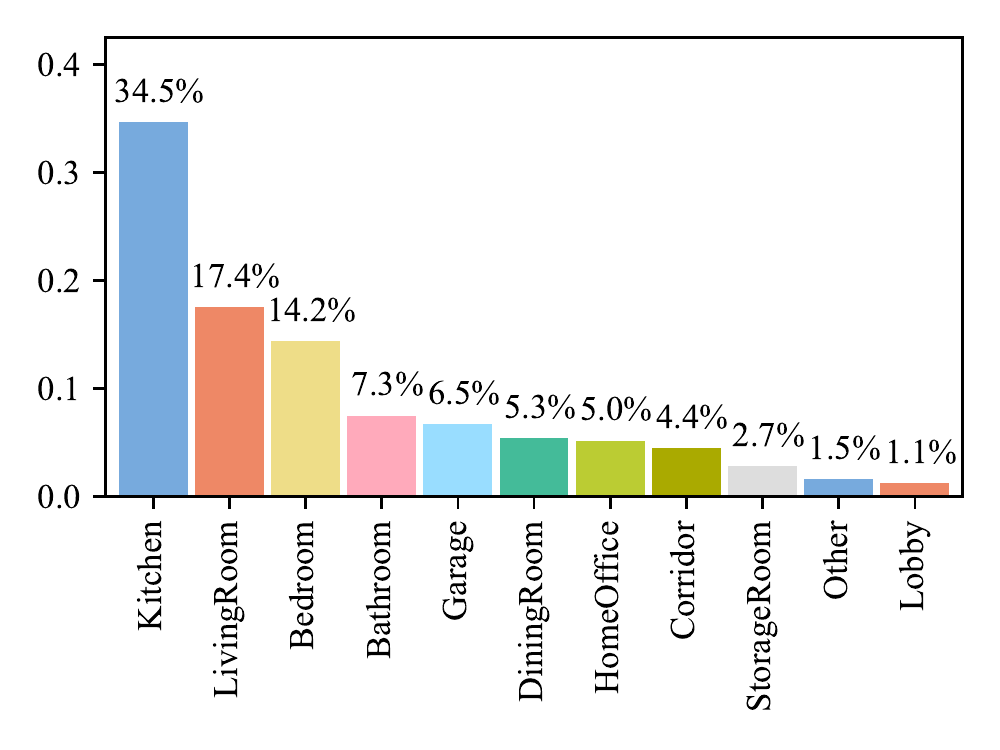}
\caption{}
\label{fig:roomtotal}
\end{subfigure}
\hfill
\begin{subfigure}[b]{0.3\linewidth}
\centering
\includegraphics[width=\linewidth]{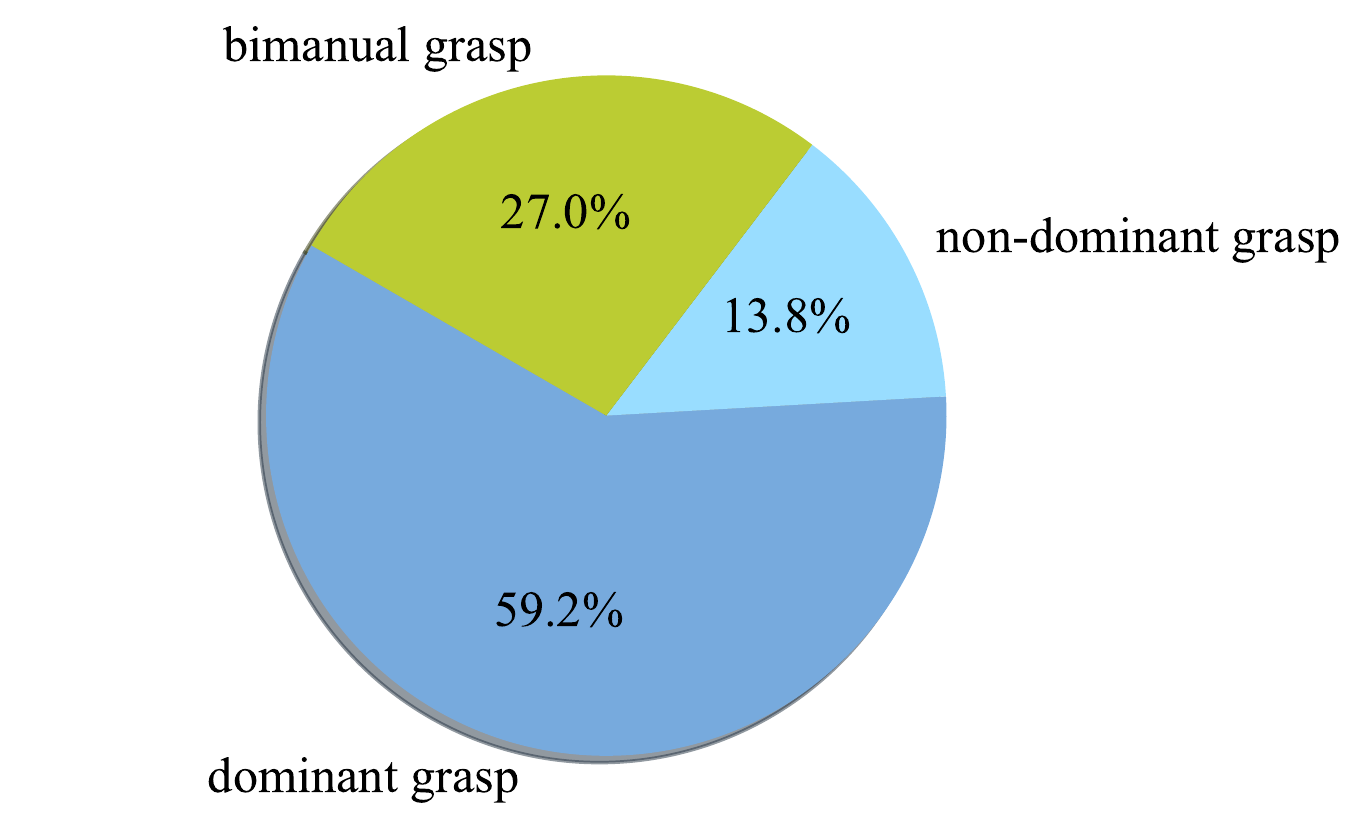}
\caption{}
\label{fig:leftrightbitotal}
\end{subfigure}
\hfill
\begin{subfigure}[t]{0.3\linewidth}
\centering
\includegraphics[width=\linewidth]{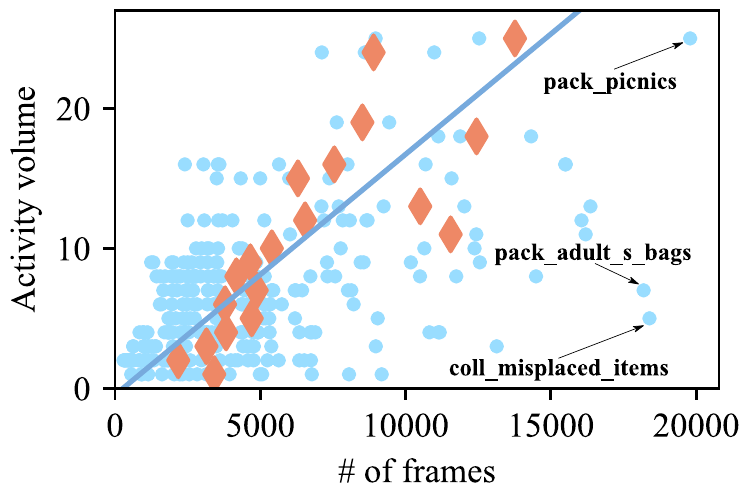}
\caption{}
\label{fig:actvol}
\end{subfigure}
    \caption{\textbf{Analysis of human demonstrations of \model activities in virtual reality:} a) Duration of each successful demonstration (mean and individual trials, decreasing order); b) Fraction of total VR time spent in each type of room; c) Fraction of total VR time spent manipulating with the dominant, non-dominant, or both hands; d) Duration of each VR demonstration wrt. activity volume; blue dots denote individual demos and red diamonds denote the mean time for each number of ground literals (activity volume). Larger volume correlates with larger duration ($R^2$ = 0.826).}
    \label{fig:vrstatistics}
\end{figure}

Fig.~\ref{fig:vrstatistics} includes a) the duration of the VR demonstrations, b) the time spent in different room types, c) the hand used to interact and manipulate, and d) the complexity of the activities in logical representation vs. time. We observe that \model activities cover a wide range of time-horizons, from less than 2 minutes to more than 11 minutes.
The activities show a bias towards living spaces (kitchen, living-room, bedroom), with the most prevalent room being the kitchen. A large portion of \model activities involve preparing food or cleaning appliances that are only supported in kitchens. 
Furthermore, as expected, the data reflect a bias towards dominant hand manipulation, followed by bimanual grasping, which is required for lifting and manipulating large objects. 
The high use of two hands to manipulate correlate to the use in real-world; we hope that our dataset helps exploring this type of interaction that has been traditionally less studied in embodied AI.
The total number of ground predicates (activity volume) is strongly correlated with the total activity time indicating that the volume is a good measure of the complexity of an activity. Outliers include activities with a high ratio of time to goal condition such as the ones that require cleaning a large area ({\small\texttt{cleaningCarpets}}, {\small\texttt{vacuumingFloors}}) or searching ({\small\texttt{collectMisplacedItems}}).

\begin{figure}[!t]
\centering
    \begin{subfigure}[b]{0.99\linewidth}
        \centering
        \includegraphics[width=\linewidth]{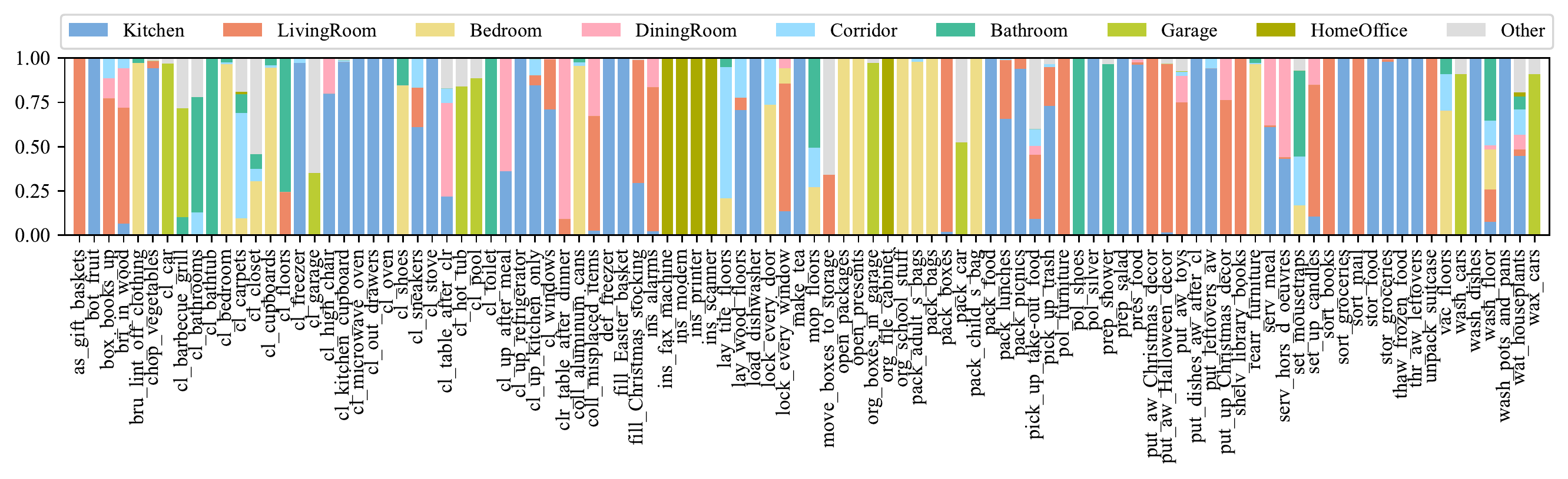}
        \caption{}
        \label{fig:roomact}
    \end{subfigure}\\
    \begin{subfigure}[b]{0.99\linewidth}
        \centering
        \includegraphics[width=\linewidth]{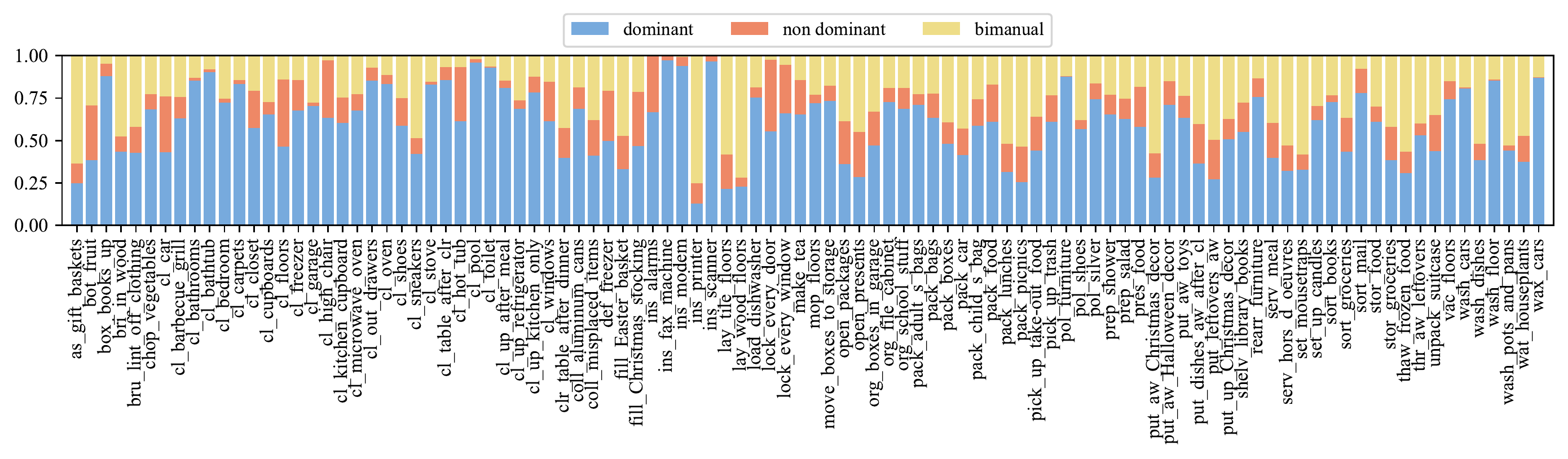}
        \caption{}
        \label{fig:rightleftbiact}
    \end{subfigure}
    \caption{\textbf{Further analysis of human demonstrations of \model activities in virtual reality:} a) Fraction of the duration of each activity spent in different types of room; b) Fraction of each activity spent manipulating with the right, left, or both hands; \model activities present a large diversity of room types: while some activities are mostly performed on a single type of room, others requires visiting different types; while the majority the activities in \model are performed with the dominant hand, a significant number of them require using both hands, e.g., {\small\texttt{installingPrinter}} or {\small\texttt{assemblingGiftBasket}}. }
    \label{fig:vrstatistics2}
\end{figure}

Analyzed individually, Fig.~\ref{fig:vrstatistics2} shows that room occupancy depends heavily on the type of activity. Room occupancy reflects common intuition about household activities; the ones associated with living-space decorations ({\small\texttt{puttingAwayChristmasDecorations}}, {\small\texttt{puttingAwayHalloweenDecorations}}) take place primarily in the living room, whereas cooking activities ({\small\texttt{preparingSalad}}, {\small\texttt{preservingFood}}) occur primarily in the kitchen. Similar activity preferences are observed in the grasping data; activities requiring installing unwieldy objects ({\small\texttt{layingWoodFloors}}, {\small\texttt{layingTileFloors}}) require the use of both hands, whereas simple cleaning activities ({\small\texttt{cleaningThePool}}) that require using a cleaning tool are performed with the dominant hand.

\begin{figure}[!t]
    \centering
    \includegraphics[width=0.34\linewidth]{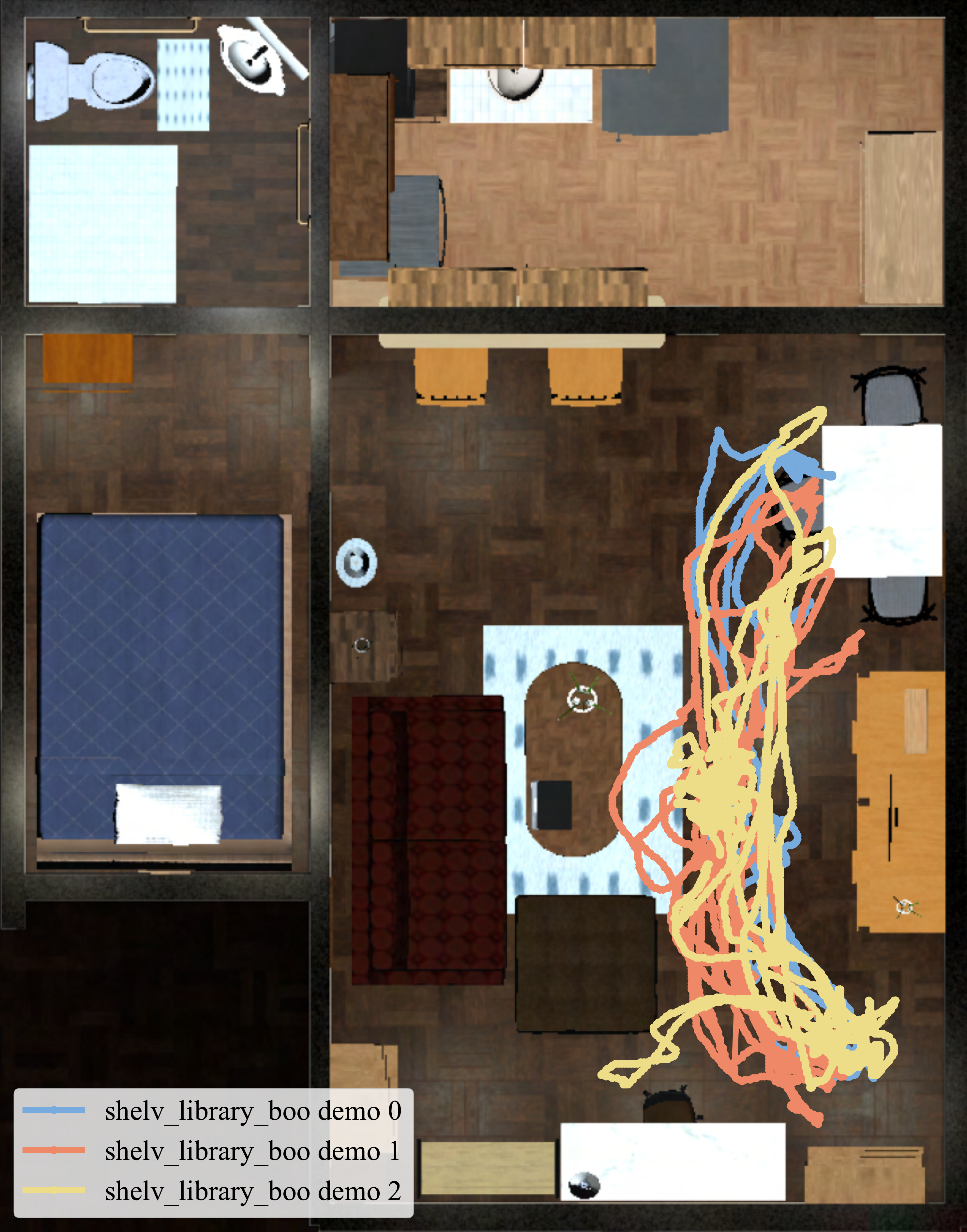}
    \hfill
    \includegraphics[width=0.34\linewidth]{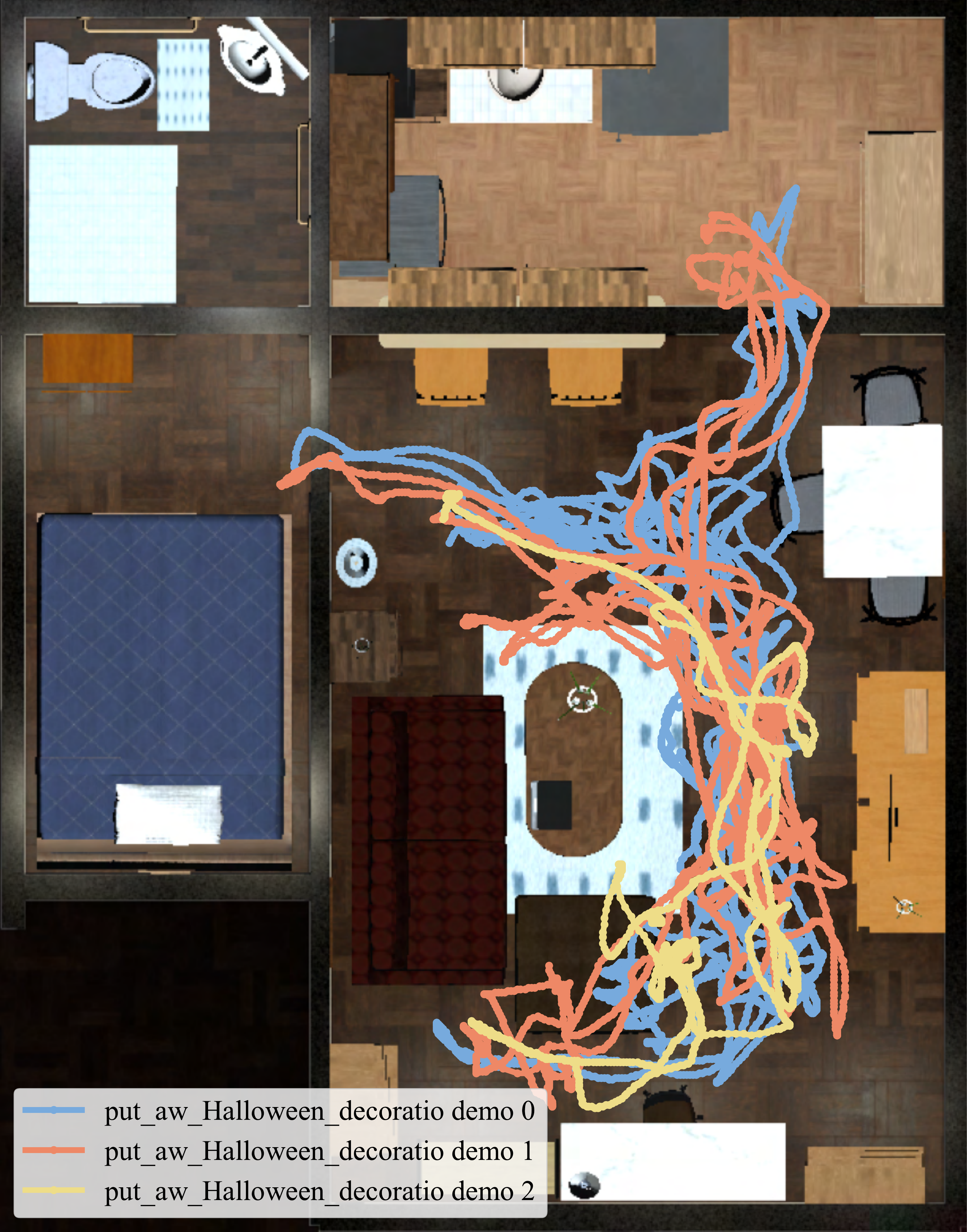}
    \hfill
    \includegraphics[width=0.285\linewidth]{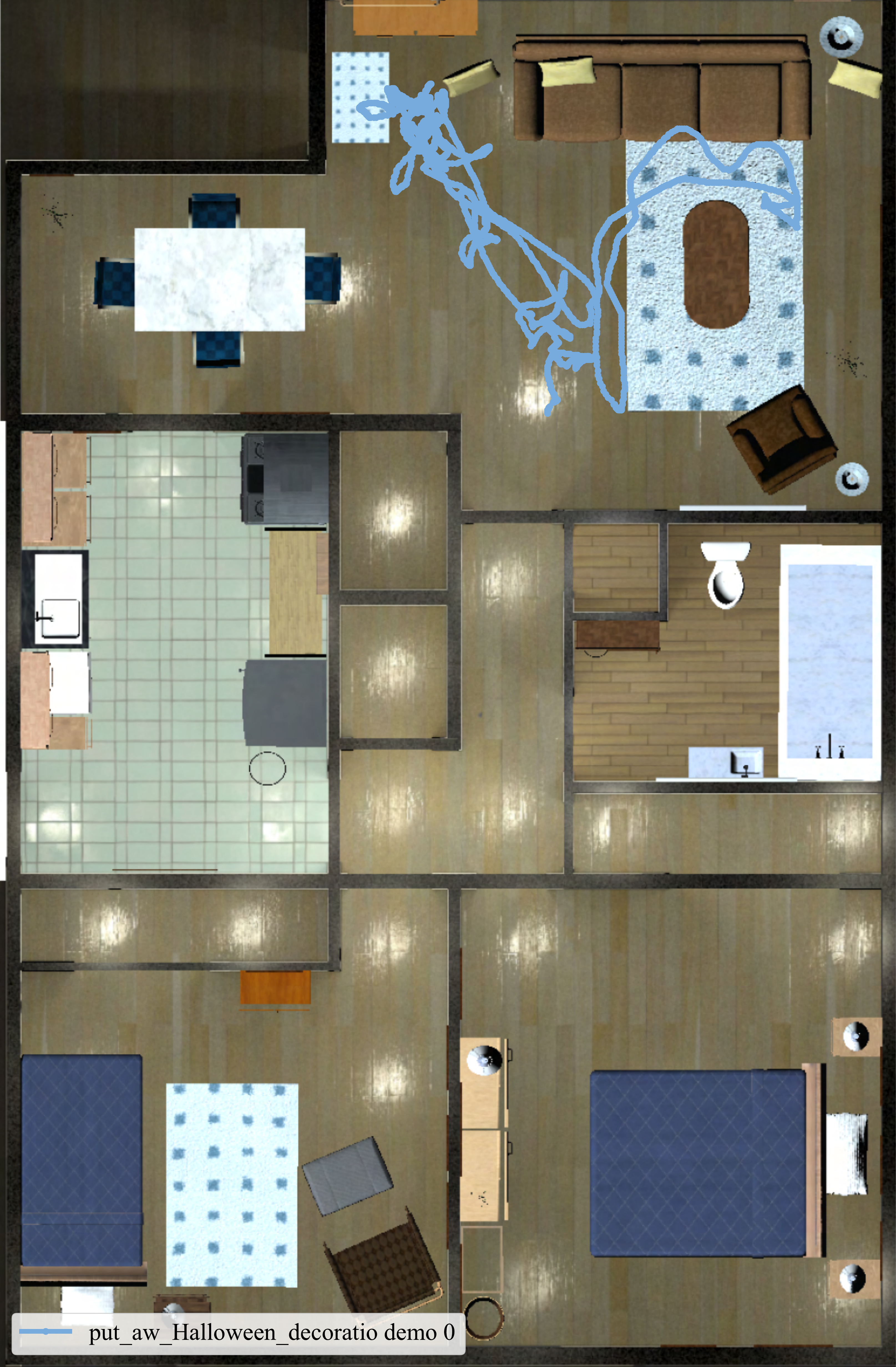}
    \caption{\textbf{Navigation trajectories of humans demonstrating activities in virtual reality:} Trajectories of different demonstrators in two scenes, \texttt{Rs\_int} (left, center) and \texttt{Merom\_1\_int} (right), for two activities, {\small\texttt{re-shelvingLibraryBooks}} (left) and {\small\texttt{puttingAwayHalloweenDecorations}} (center, right); Demonstrator trajectories present variation within activity instance and scene (each figure); Different activities in the same scene (left, center) require different rooms and areas to be explored; Trajectories differ between  scenes (center, right) due to the placement of target objects and goal locations}
    \label{fig:trajectory_map}
\end{figure}

\subsubsection{Gaze Tracking in Virtual Reality}

\begin{figure}[!t]
    \centering
    \begin{subfigure}{.16\textwidth}
      \centering
      \includegraphics[width=1.0\linewidth]{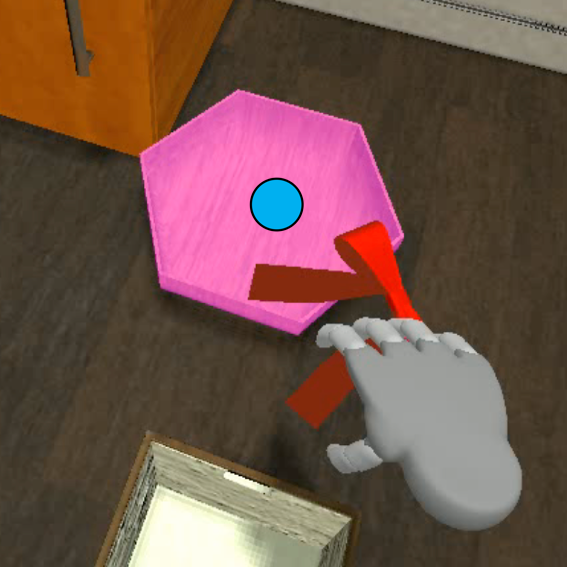}  
      \vspace{-0.5cm}
      \caption*{\tiny {assembling gift baskets}}
    \end{subfigure}
    \hfill
    \begin{subfigure}{.16\textwidth}
      \centering
      \includegraphics[width=1.0\linewidth]{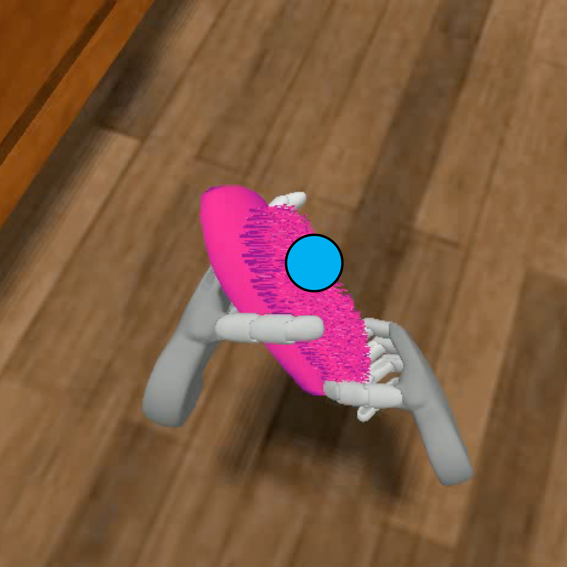} 
      \vspace{-0.5cm}
      \caption*{\tiny brushing lint off clothing}
    \end{subfigure}
    \hfill
    \begin{subfigure}{.16\textwidth}
      \centering
      \includegraphics[width=1.0\linewidth]{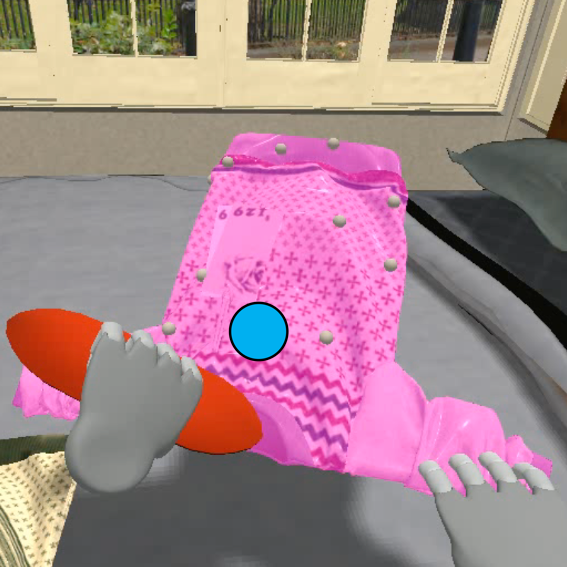}  
      \vspace{-0.5cm}       
      \caption*{\tiny brushing lint off clothing}
    \end{subfigure}
    \hfill
    \begin{subfigure}{.16\textwidth}
      \centering
      \includegraphics[width=1.0\linewidth]{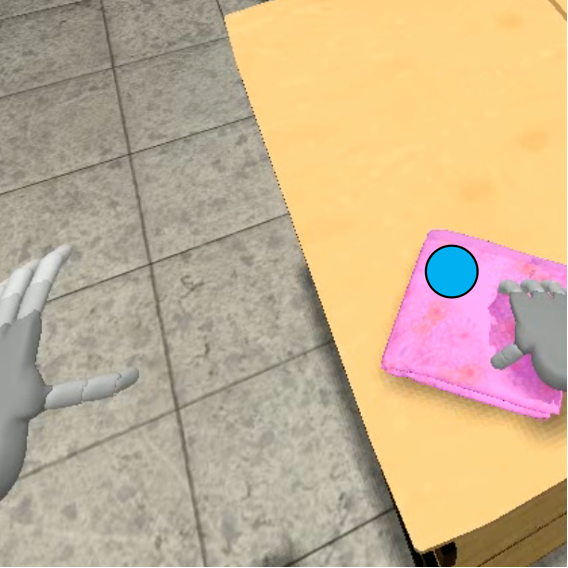}    
      \vspace{-0.5cm}       
      \caption*{\tiny cleaning high chair}
    \end{subfigure}
    \hfill
    \begin{subfigure}{.16\textwidth}
      \centering
      \includegraphics[width=1.0\linewidth]{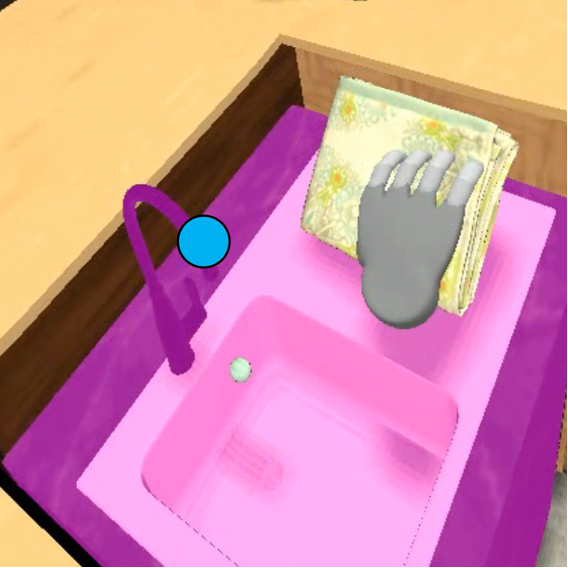}   
      \vspace{-0.5cm}       
      \caption*{\tiny cleaning high chair}
    \end{subfigure}
    \hfill
    \begin{subfigure}{.16\textwidth}
      \centering
      \includegraphics[width=1.0\linewidth]{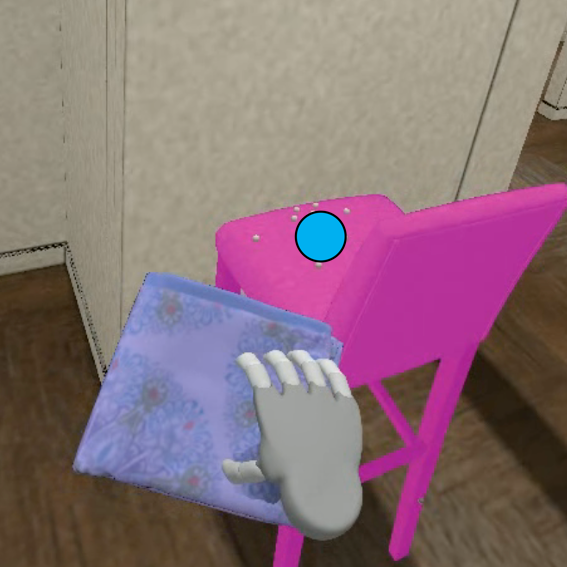}  
      \vspace{-0.5cm}       
      \caption*{\tiny cleaning high chair}
    \end{subfigure}\\
    \begin{subfigure}{.16\textwidth}
      \centering
      \includegraphics[width=1.0\linewidth]{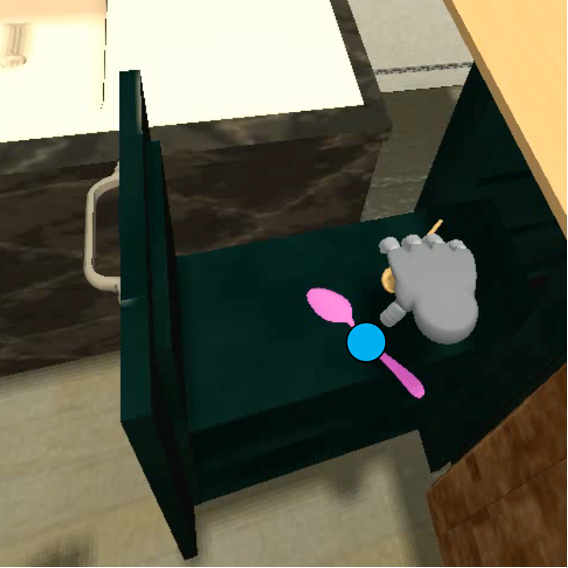}  
      \vspace{-0.5cm}       
      \caption*{\tiny cleaning out drawers}
    \end{subfigure}
    \hfill
    \begin{subfigure}{.16\textwidth}
      \centering
      \includegraphics[width=1.0\linewidth]{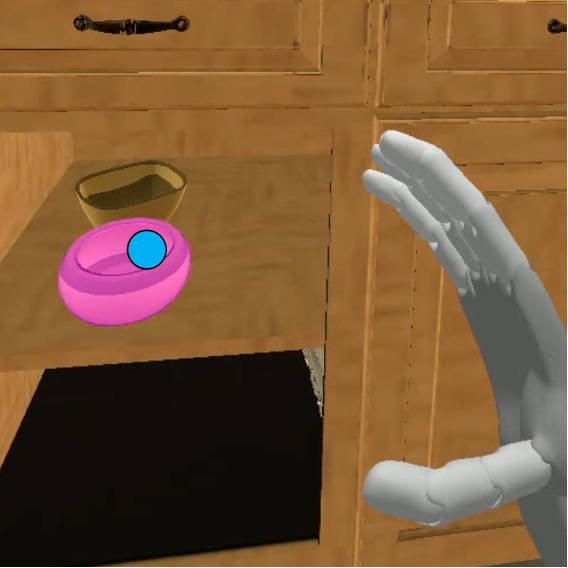}    
      \vspace{-0.5cm}       
      \caption*{\tiny cleaning out drawers}
    \end{subfigure}
    \hfill
    \begin{subfigure}{.16\textwidth}
      \centering
      \includegraphics[width=1.0\linewidth]{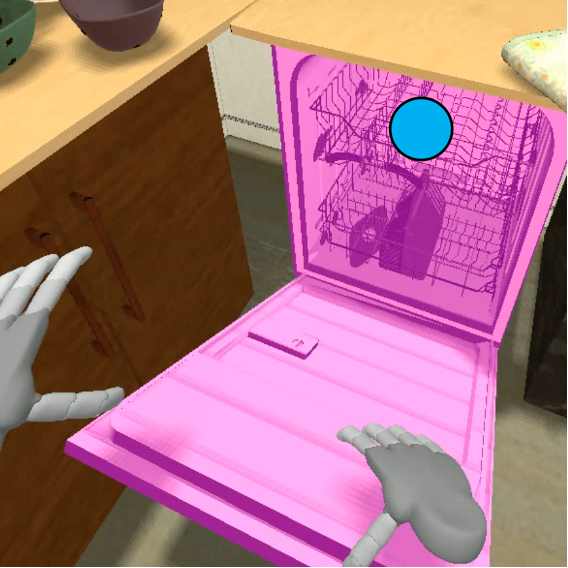}  
      \vspace{-0.5cm}       
      \caption*{\tiny loading the dishwasher}
    \end{subfigure}
    \hfill
    \begin{subfigure}{.16\textwidth}
      \centering
      \includegraphics[width=1.0\linewidth]{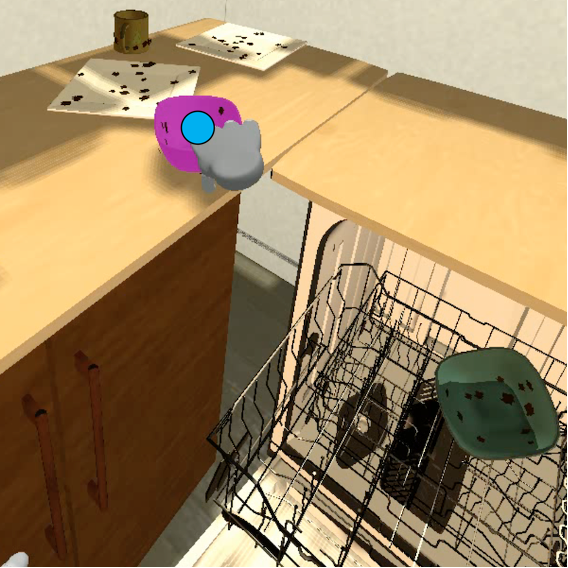}    
      \vspace{-0.5cm}       
      \caption*{\tiny loading the dishwasher}
    \end{subfigure}
    \hfill
    \begin{subfigure}{.16\textwidth}
      \centering
      \includegraphics[width=1.0\linewidth]{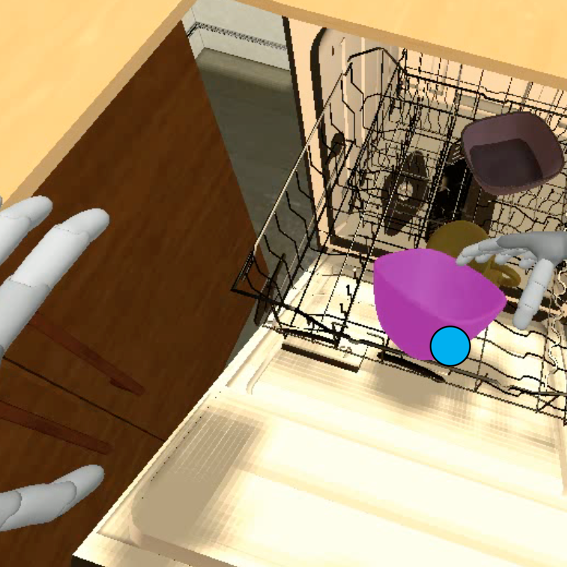}   
      \vspace{-0.5cm}       
      \caption*{\tiny loading the dishwasher}
    \end{subfigure}
    \hfill
    \begin{subfigure}{.16\textwidth}
      \centering
      \includegraphics[width=1.0\linewidth]{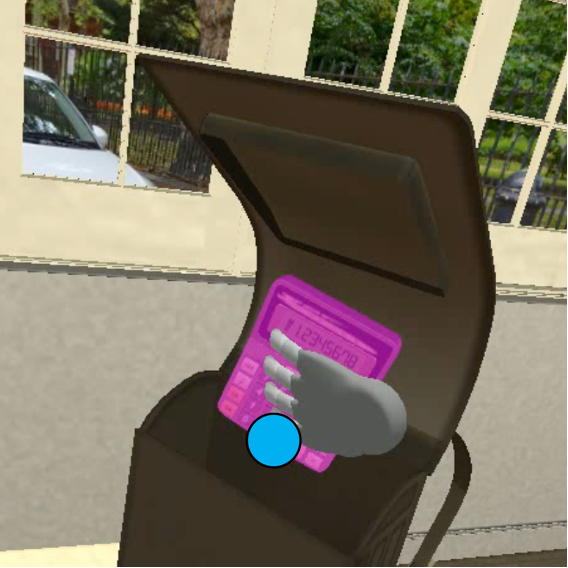}  
      \vspace{-0.5cm}       
      \caption*{\tiny organizing school stuff}
    \end{subfigure}
    \caption{Human gaze during activity execution; Red dot: human gaze point, Magenta: object gazed; The \model Dataset of Human Demonstrations in Virtual Reality includes \numDemos demonstrations (\totalTime min) with gaze information while humans navigate and interact (accuracy: $\pm$4$^\circ$~\cite{sipatchin2020accuracy}); The gaze information correlates strongly with activity; We hope that this data can support new research in visual attention and active vision to control agent's camera}
    \label{fig:human_gaze_examples}
\end{figure}

\begin{figure}[!t]
\centering
    \begin{subfigure}[t]{0.99\textwidth}
      \centering
      \includegraphics[width=1.0\linewidth]{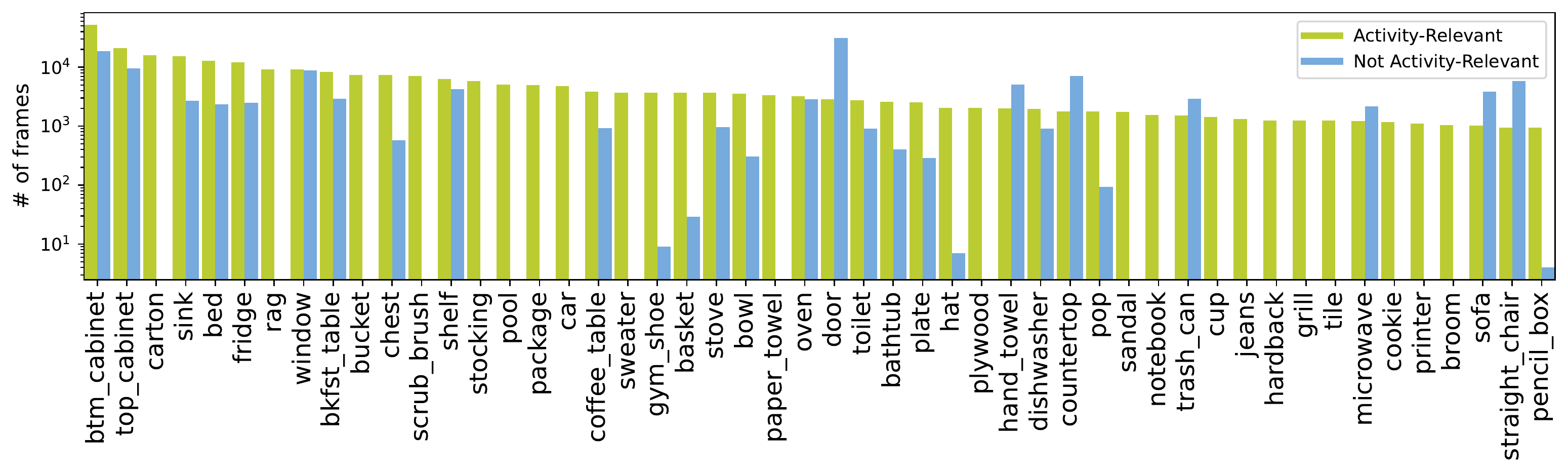}
    \end{subfigure}\\
    \begin{subfigure}[t]{.285\textwidth}
      \centering
      \caption*{\tiny assembling gift baskets}
      \vspace{-0.25cm}
      \includegraphics[width=1.0\linewidth]{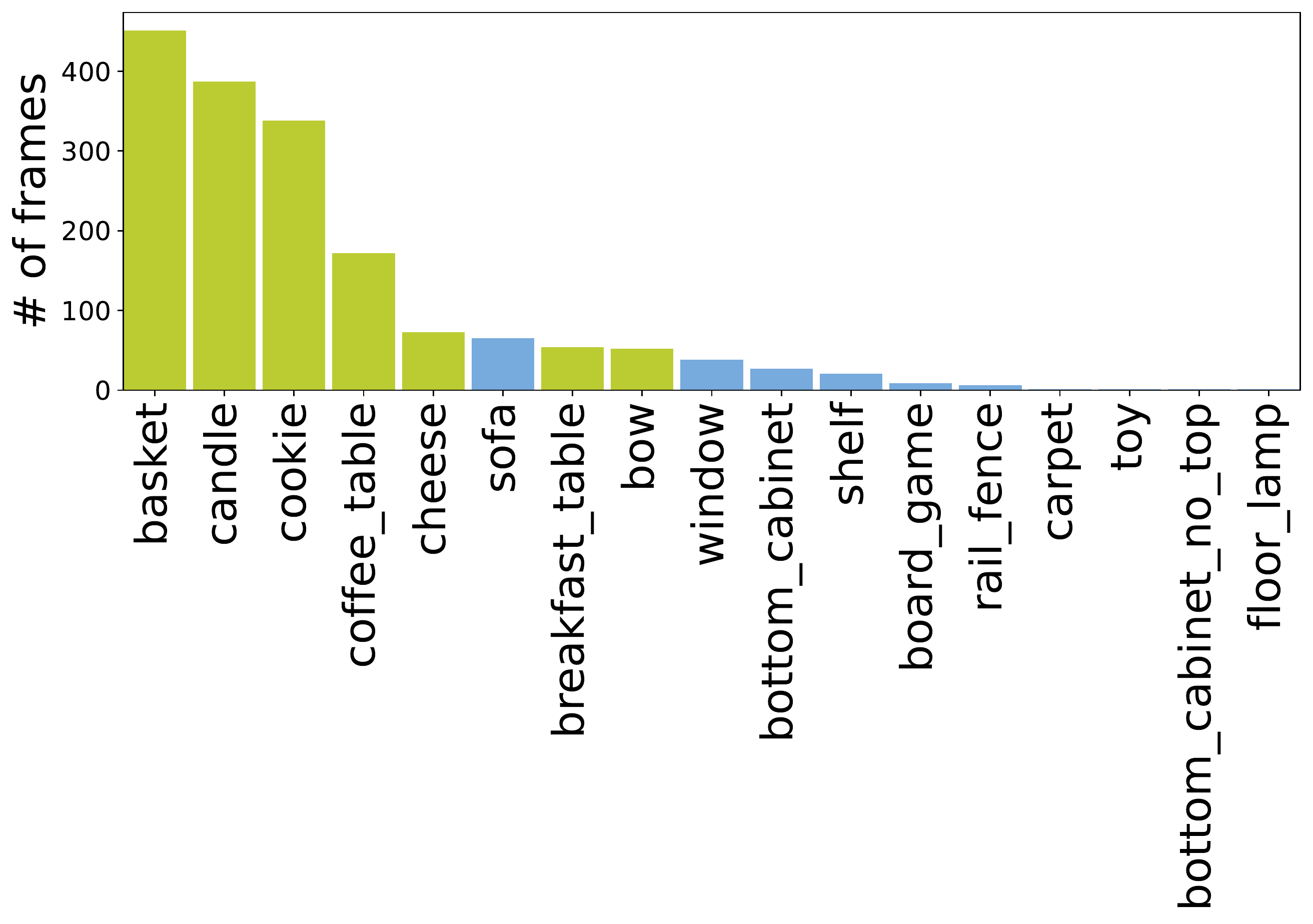} 
    \end{subfigure}
    \hfill
    \begin{subfigure}[t]{.26\textwidth}
      \centering
      \caption*{\tiny boxing books up for storage}
      \vspace{-0.25cm}
      \includegraphics[width=1.0\linewidth]{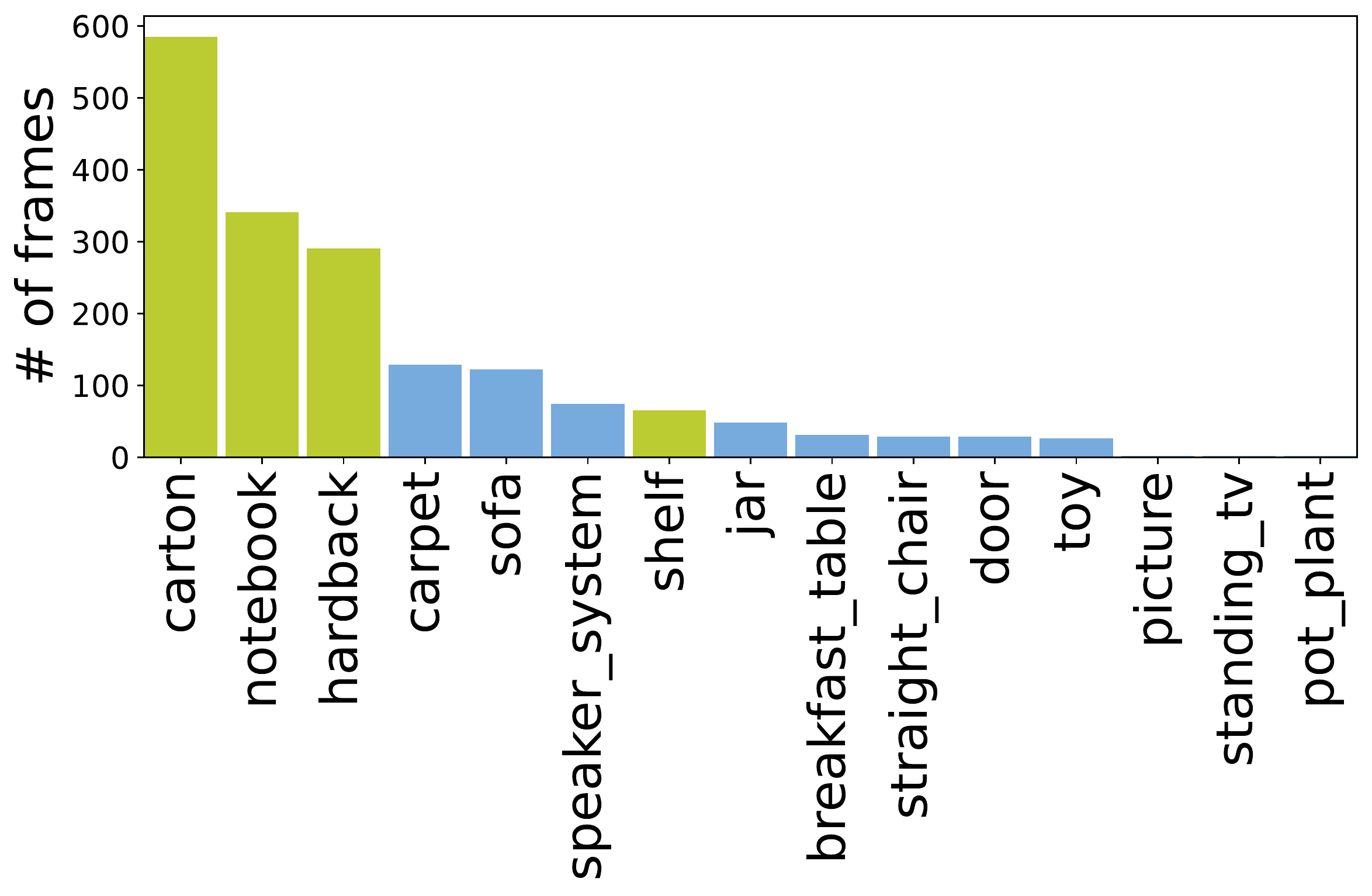}  
    \end{subfigure}
    \hfill
    \begin{subfigure}[t]{.24\textwidth}
      \centering
      \caption*{\tiny bringing in wood}
      \vspace{-0.25cm}
      \includegraphics[width=1.0\linewidth]{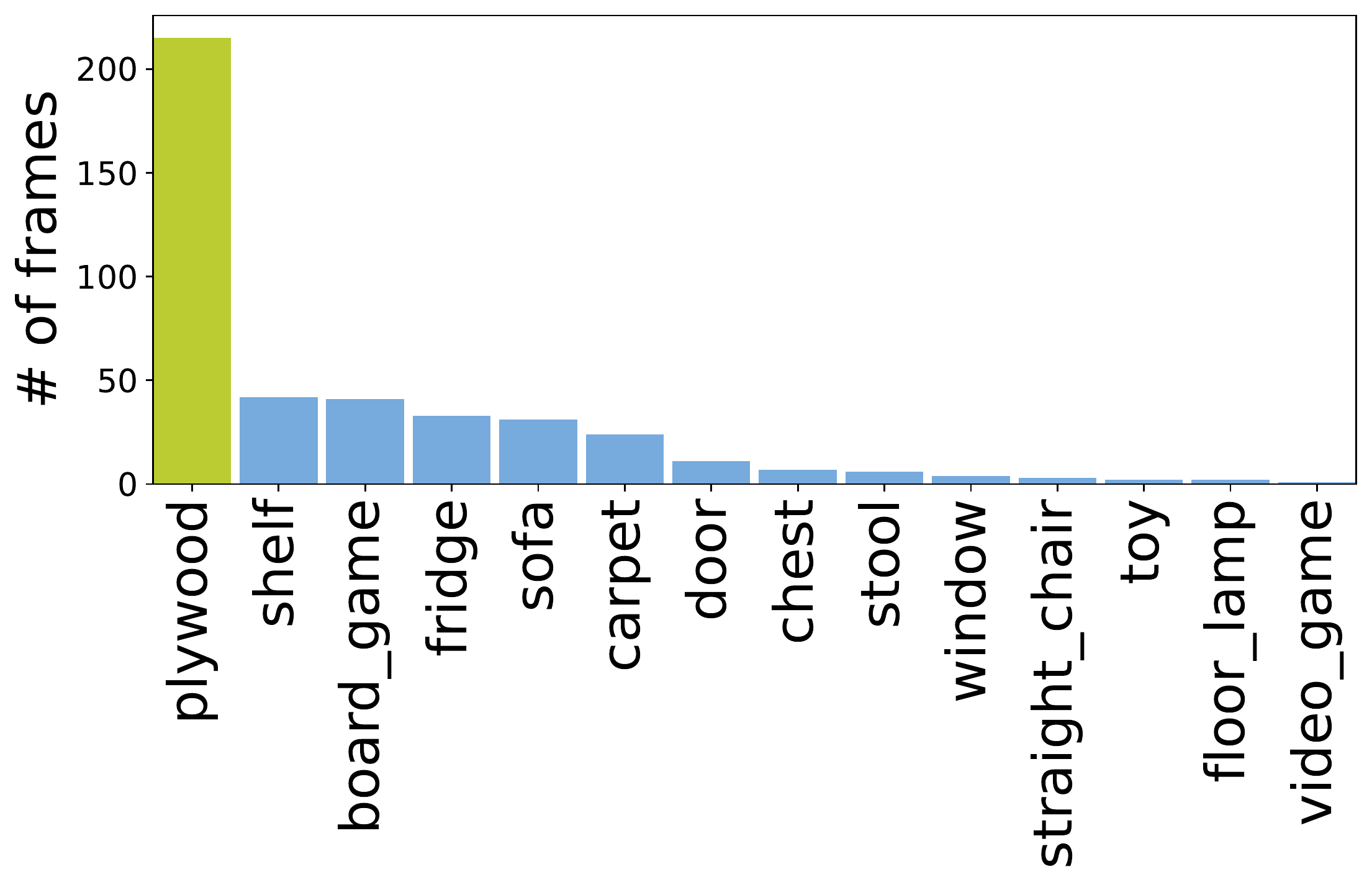}   
    \end{subfigure}
    \hfill
    \begin{subfigure}[t]{.195\textwidth}
      \centering
       \caption*{\tiny brushing lint off clothing}
       \vspace{-0.25cm}
      \includegraphics[width=1.0\linewidth]{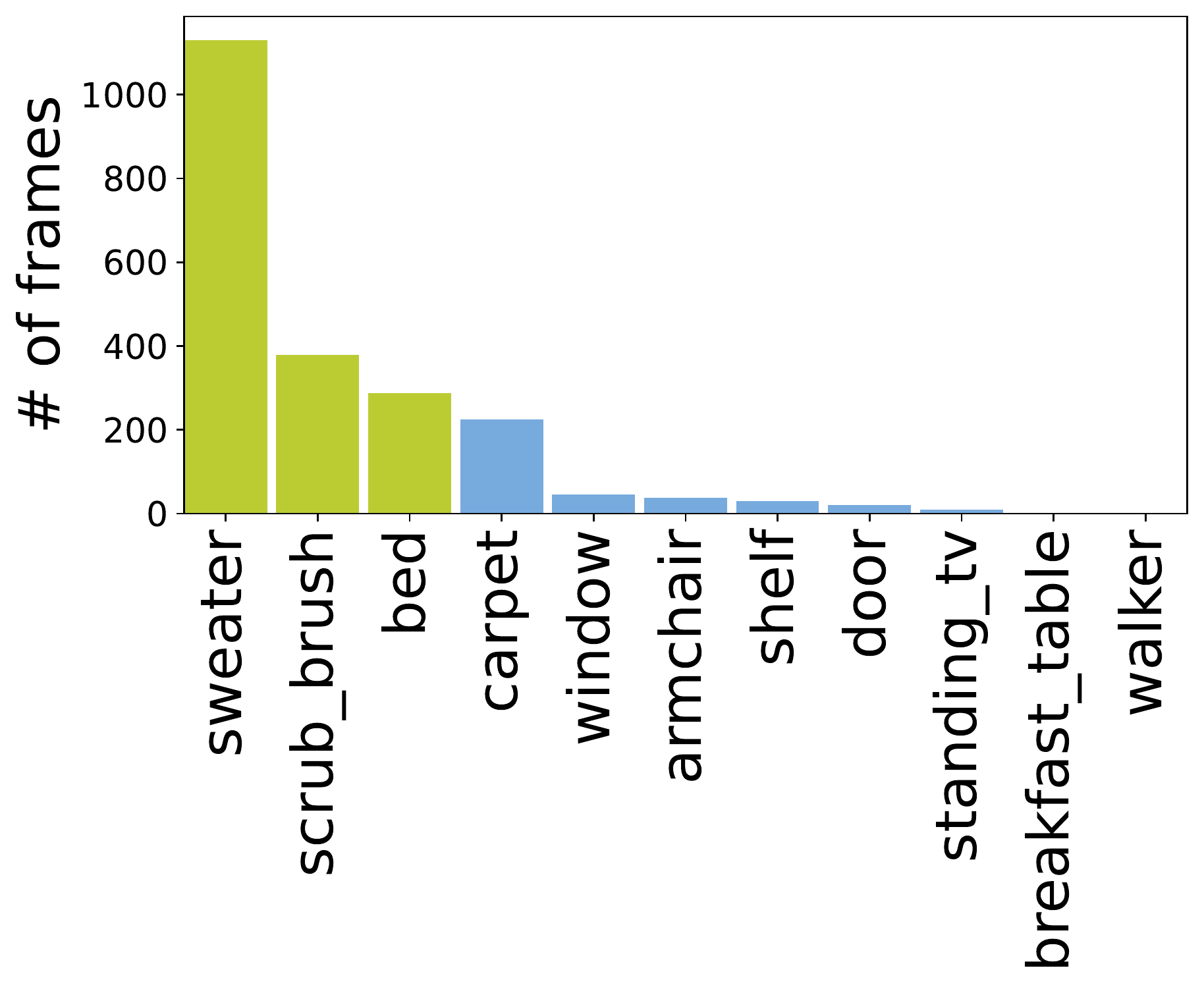}   
    \end{subfigure}
    \caption{Statistics of the attention over object instances of WordNet categories in the \model Dataset of Human Demonstrations in virtual reality aggregated for all demonstrations (top, logarithmic scale), and segregated for four activities (bottom row, linear scale), for activity-relevant (green) and not activity-relevant (blue) objects. In households, the aggregated of the visual attention goes to containers of objects (cabinets) and doors separating rooms; For individual activities, visual attention concentrate on specific activity-relevant objects}
    \label{fig:gaze_stats}
\end{figure}

Our preference on HTC Vive Pro to collect the \model Dataset of Human Demonstrations is motivated by its ability to track the gaze (pupil movement) of the demonstrator.
We consider gaze information to be a valuable source to understand human performance in the activities.
While other datasets of gaze are available~\cite{kothari2020gaze}, this is the largest dataset of active gaze attention during manipulation in simulation, providing synchronized ground-truth information of the object being observed, its state and full shape.
Fig.~\ref{fig:human_gaze_examples} depict examples of the tracked human gaze during activity execution, with the object attracting the gaze indicated in magenta.
Fig.~\ref{fig:gaze_stats} includes several statistics of the gaze attention over object categories in the entire dataset and for some example activities.
Both figures indicate a clear correlation between the gaze data and the goal of the activities: we expect the dataset to be useful to study and predict human gaze attention, and to develop new embodied AI algorithms for active~\cite{aloimonos1988active,ballard1991animate} and interactive perception~\cite{bohg2017interactive}.


\subsection{Additional Details on the Experimental Setup}
\label{ss:expdetails}

In the following, we share more details about the experimental setup and training procedure. We primarily use two different training setups: reinforcement learning (RL) with continuous action space and RL with motion primitives. We will first elaborate the shared setup between the two, then go into their differences.

\paragraph{Shared Setup:} In the normal ``partial observability'' setup, the observations include $128 \times 128$ RGB-D images from the onboard sensor on the agent's head and proprioceptive information (Head pose in local frame, hand poses in local frame, and a fraction indicating how much each hand is closed). The proprioceptive information is 20 dimentional.
For the experiments with ``full observability'', the observations include the ground truth object poses for all the activity-relevant objects, the agent's pose, and the proprioceptive information.

The agent receives a reward of 1 for every ground goal condition (literal) that it satisfies during the episode. The episode terminates if the agent achieves a success score $Q$ of 1 (achieved all literals in the goal condition) or it times out. 

The policy network architecture is largely shared in the following setups. With RGB-D images as input, we use a 3-layer convolutional neural network to encode the image into a 256 dimensional vector. Proprioceptive information and/or poses for all activity-relevant objects are also encoded into a 256 dimensional vector with an MLP, respectively. The features are concatenated and pass through another MLP to generate action representation, which could be a box action space or discrete action space depending on the setup (continuous actions or action primitives). 

\paragraph{RL with Continuous Action Space:} For this agent variant, we use Soft Actor-Critic (SAC)~\cite{haarnoja2018soft} implemented by TF-Agents~\cite{TFAgents}. The action space is continuous and has a dimensionality of 18. The first three dimensions represent the locomotion actions: desired x-y translation of the robot body and the desired rotation around the vertical axis. The next seven dimensions represent the linear and angular velocities of the left hand (in Cartesian space, 6 DoF) and 1 DoF closing/opening of the hand. The last seven dimensions is the same action but for the right hand. The maximum episode length depends on the experimental setup. For instance, if the initial state corresponds to \SI{1}{\second} away from a goal state, we will give the agent three times the amount of time (i.e. \SI{3}{\second}) to accomplish the activity. We train for 20K episodes, evaluate the final policy checkpoint and report the results in Table~\ref{tab:table_results1_am1}. 

\paragraph{RL with Motion Primitives:} For this agent variant, we use Proximal Policy Optimization (PPO)~\cite{schulman2017proximal} implemented with TF-Agents~\cite{TFAgents}. The action space is discrete, with $n_r \times m$ choices, where $n_r$ is the number of activity-relevant objects and $m$ is the number of action primitives. Here we didn't allow the agent to operate on all objects in the scene, but focus on activity-relevant objects to facilitate learning. Following our implementation of motion primitives, $m = 6$. Laying out the choices on a $n_r \times m$ grid, and $i$-th column $j$-th row means to apply $j$-th action primitive on $i$-th activity-relevant object. Not all combinations of action primitive and object are compatible and action that is not feasible is converted into no-ops. The maximum episode length is set to $100$ for all activities. We experiment with partially simulated motion primitives and fully simulated motion primitives, as described in Sec.~\ref{ss:igv2}). We train with partially simulated motion primitives until convergence, and evaluate and report the results on partially simulated motion primitives and fully simulated motion primitives, since training with motion planning in a complex scene is very time-consuming. In the experimental results shown in Table~\ref{tab:table_results1_am1}, generally fully simulated motion primitives results are much worse than partially simulated motion primitives, this is intuitive because motion planning performs more rigorous checks and complies with the physical model, highlighting the complexity of \model. 

\paragraph{Experimental Setup for the Effect of Diversity:} To evaluate diversity, we train for individual skills instead of full \model activities. Here, we adopt an easier experimental setup that allows us to study the effect of diversity; the results are reported in Table~\ref{tab:diversity_exp}. Specifically, we use RL with continuous action space but with a more constrained action space: 6-dimensional representing the desired linear and angular velocities of the right hand (assuming the rest of the agent is stationary). For grasping, we adopt the ``sticky mitten'' simplification from other works~\cite{batra2020rearrangement}: we create a fixed constraint between the hand and the object as soon as they get in contact. We also use distance-based reward shaping to encourage the hand to approach activity-relevant objects. To evaluate the effect of diversity in object poses, we use the same object models and randomize their initial poses during training. To evaluate the effect of diversity in object instances, we randomize the object models during training. For example, for the \texttt{sliced} single-predicate activity, the agent will encounter different types of fruit (e.g. peach, strawberry, pineapple, etc) during training. We train for 10K episodes, evaluate the final policy checkpoint and report the results in Table~\ref{tab:diversity_exp}.

\subsection{Potential to Transfer to Real-World}
\label{ss:sim2real}

\model is a benchmark in simulation. 
This facilitates a continuous evaluation of solutions, fair and equal conditions, and increased accessibility without expensive robot hardware. 
It is also instrumental for modern robot learning procedures that require generating large amount of experiences.
However, the use of simulation introduces a gap between the activities in our benchmark and the equivalent activities in real world.
We argue that, while not negligible, we have taken measures to close this gap with the goal of providing a benchmark where the performance of embodied AI solutions is close to the performance they would have in a real world system.

Our instantation of \model includes realistic scenes and object models, with high-quality visuals and close-to-real physical properties (mass, center of mass, friction) annotated in manual process assisted by the information obtained in the Internet. 
The underlying physics engine, pyBullet~\cite{coumans2016pybullet}, is acknowledged as one of the standards in robotics and a high quality approximation of the underlying mechanical processes.
The physics-based rendering from \ig generates high quality images to use as input in our evaluation.
While our bimanual agent is not realistic, our second provided embodiment is a realistic robot model, a Fetch, with similar kinematics, actuation and sensing, facilitating the evaluation of solutions in \model that could act similarly on a real robot.
Previous works have demonstrated good results developing solutions in \ig that could transfer to real world~\cite{hirose2019deep,kang2019generalization}, and evaluated the similarities between simulated and real-world sensor signals~\cite{xia2020relmogen}.
This indicates a high potential for the solutions evaluated in simulation in \model to perform similarly in the real world, a claim that we plan to evaluate experimentally after the pandemic.

\subsection{Ethical Considerations}
BEHAVIOR includes data (activity demonstrations) generated by humans. After evaluation, our institution's Institutional Review Board (IRB) considered this project exempt from review: the data does not reveal any private information about the participants. Human demonstrations are collected in line with standard ethics practices, among lab members and volunteers. In terms of broader societal impacts, \model is aimed to facilitate research of autonomous robots performing activities of daily living. The potential impacts of this line of work, particularly on labor impacts of automation and the physical safety of humans interacting with autonomous robots, are far-reaching.

\end{document}